\documentclass[10pt,twocolumn,letterpaper]{article}

\usepackage{iccv}
\usepackage{times}
\usepackage{epsfig}
\usepackage{appendix}
\usepackage{chngcntr}
\usepackage{graphicx}
\usepackage{comment}
\usepackage{booktabs} 
\usepackage{microtype} 
\usepackage{amsmath,amssymb} %
\usepackage{color}
\usepackage{array}
\usepackage[export]{adjustbox}
\usepackage[accsupp]{axessibility}

\DeclareMathOperator*{\argmax}{arg\max}

\usepackage{color}

\newcommand{\leaveout}[1]{}

\newcommand{\zsl}{ZSL }

\newcommand{\zslinter}{ZSL-Interactive }
\newcommand{\zslinternospace}{ZSL-Interactive}

\newcommand{\sibvar}{Sibling-variance}
\newcommand{\sibvareq}{Q_{sv}}
\newcommand{\sibvarpi}{\pi_{sv}}
\newcommand{\encchange}{Representation-change}
\newcommand{\encchangeeq}{Q_{rc}}
\newcommand{\encchangepi}{\pi_{rc}}

\newcommand{\imagebased}{Image-based}
\newcommand{\imagebasedeq}{Q_{I}}
\newcommand{\imagebasedpi}{\pi_{I}}

\usepackage[pagebackref=true,breaklinks=true,letterpaper=true,colorlinks,bookmarks=false]{hyperref}

\iccvfinalcopy %
\ificcvfinal\fi

\begin{document}

\title{Field-Guide-Inspired Zero-Shot Learning}

\author{Utkarsh Mall, Bharath Hariharan, Kavita Bala\\
Cornell University\\
{\tt\small \{utkarshm, bharathh, kb\}@cs.cornell.edu}
}

\maketitle

\begin{abstract}
Modern recognition systems require large amounts of supervision to
achieve accuracy.  Adapting to new domains requires significant data
from experts, which is onerous and can become too expensive.
Zero-shot learning requires an annotated set of attributes for a
novel category. Annotating the full set of attributes for a novel
category proves to be a tedious and expensive task in
deployment.  This is especially the case when the recognition domain
is an expert domain.  We introduce a new {\bf field-guide-inspired} approach
to zero-shot annotation where the learner model interactively asks for
the most useful attributes that define a class.  We evaluate our
method on classification benchmarks with attribute annotations like
CUB, SUN, and AWA2 and show that our model achieves the performance of
a model with full annotations at the cost of significantly fewer
number of annotations.  Since the time of experts is precious,
decreasing annotation cost can be very valuable for real-world
deployment. 
\end{abstract}

\section{Introduction} \label{sec:intro}
\begin{figure}
\centering
\includegraphics[width=0.4\linewidth]{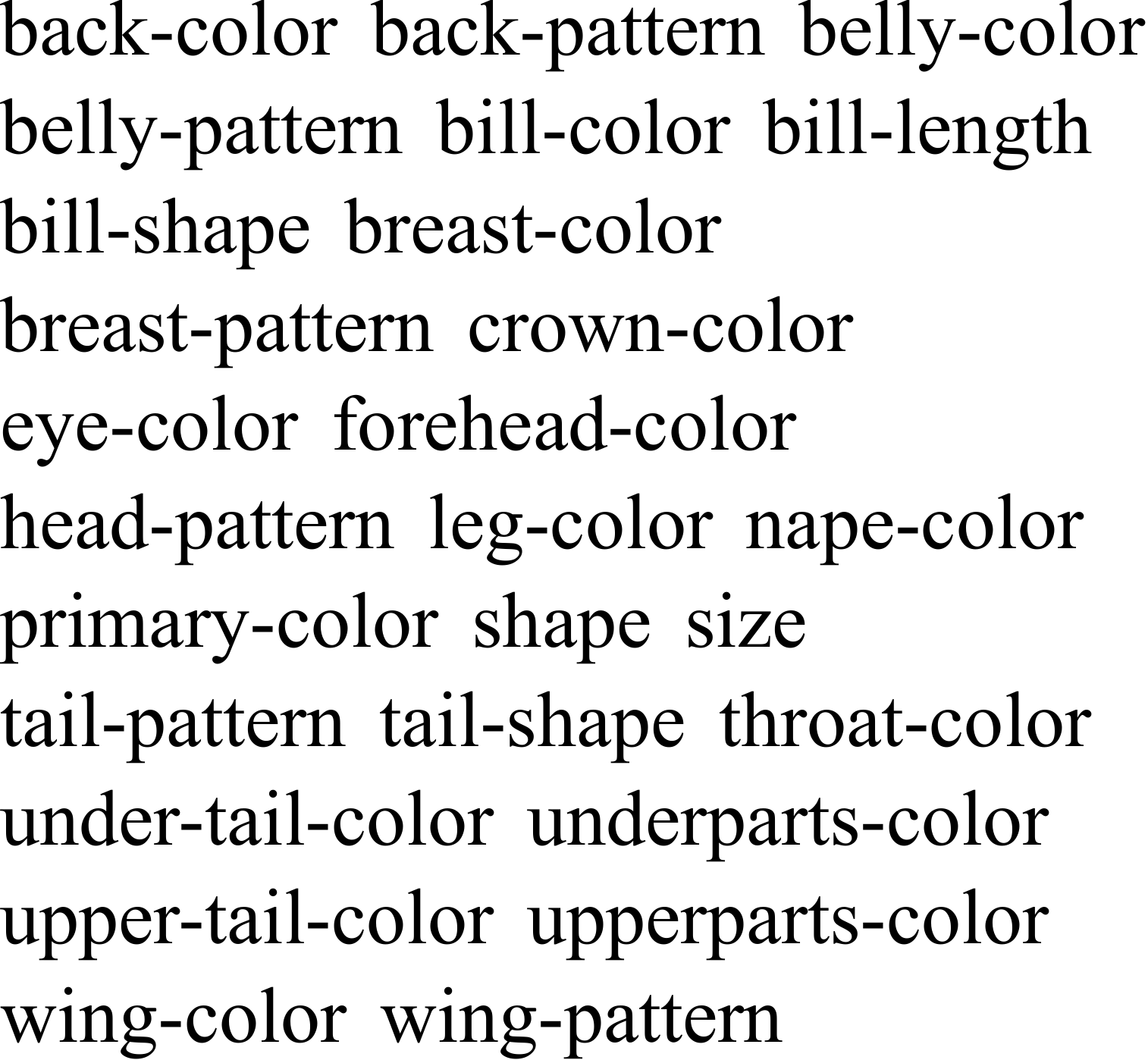}
\includegraphics[width=0.57\linewidth]{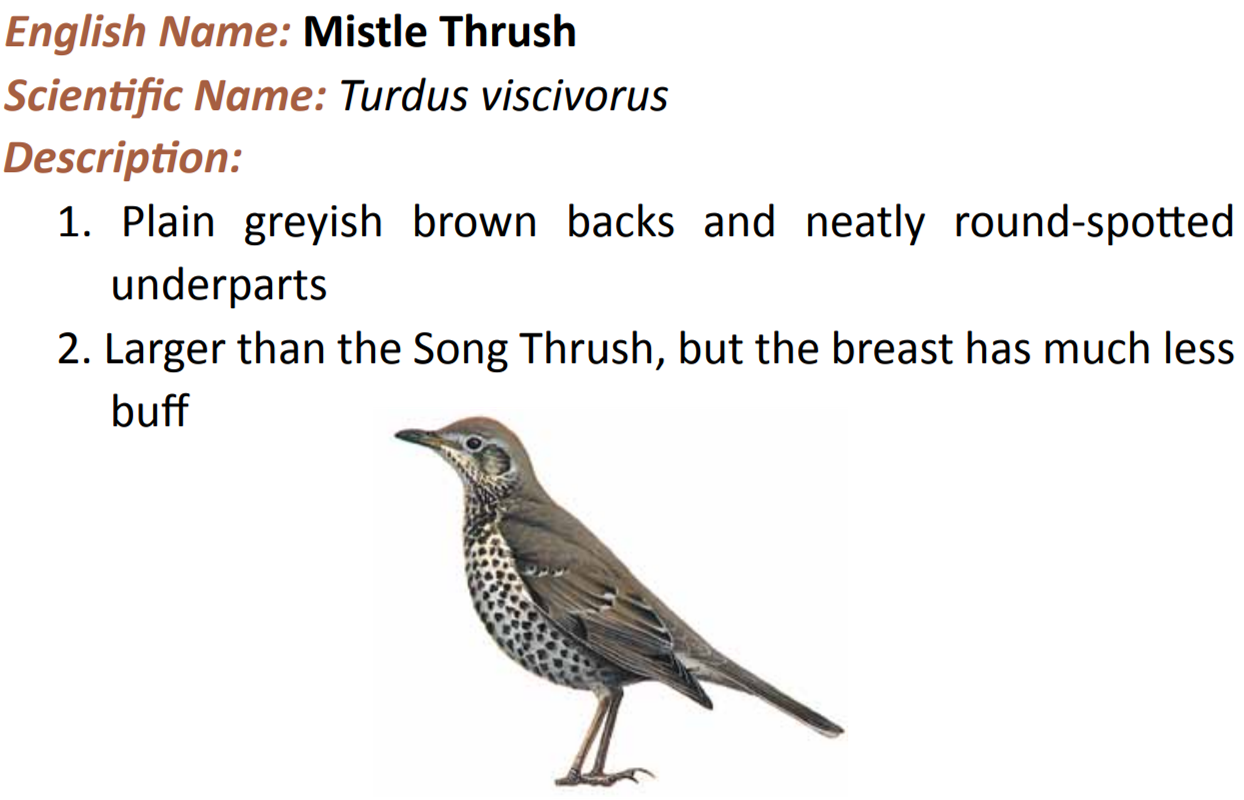}
        \caption{\textbf{Left:} For each new class in the CUB dataset,
an annotator must label 28 attributes (and 312 different
quantities).
        The attributes are multidimensional and continuous-valued,
        \textbf{Right:} A {\em field guide} example of a bird class. 
        There is no full attribute description; instead the bird is described by a few key attributes that distinguish it from close cousins.
		}
\label{fig:fig1} 
\vspace{-1.5em} 
\end{figure}

Modern recognition systems require vast amounts of labeled data.  
This is infeasible to acquire in many domains, especially when the classes in question involve subtle distinctions that require an expert: experts have limited time and availability and cannot annotate thousands of images. 
This has motivated research into \emph{zero-shot learning} (ZSL) where the goal is to build effective recognition models from \emph{class descriptions} alone. 
The name ``zero-shot'' suggests that the labeling effort for the annotator is zero.
However, this is not true: in current ZSL systems, the annotator must specify for each class \emph{hundreds} of attributes for \emph{thousands} of classes (Figure~\ref{fig:fig1} (left)).
For example, in one of our preliminary experiments, it took an
ornithologist more than \textbf{15 minutes} to fully describe the 312 different
attributes in CUB; annotating all 200 classes would have
taken the expert more than \textbf{50 hours}! 
For ZSL to truly decrease the annotator's burden,  the cost of
attribute description has to be significantly cheaper.

Past work on addressing this concern has looked at using freely available text from the internet, e.g., Wikipedia\cite{elhoseiny-13, lei-15, alhalah2-17, qiao-16, zhu-18}, or relying on word embeddings of class names\cite{alhalah-16, frome-13, rohrbach-10, akata-15}.
While such ``unsupervised'' ZSL is intriguing, neither class names nor Wikipedia descriptions are intended to be used for \emph{identifying} classes.
Class names are terse and may often be based on location rather than appearance (e.g.,``Northern Royal Albatross'').
Wikipedia articles might likewise include irrelevant information (e.g., about etymology) and miss vital visual details.
As such, we find that these approaches result in a substantial
reduction in accuracy of almost 20 points(see Figure \ref{fig:quantitative} (middle row)).
This suggests that expert-provided visual characteristics or attributes are indeed crucial for performance.

How can we record expert knowledge of class distinctions as completely as possible while still reducing their burden?
One could simply reduce the \emph{vocabulary} of attributes that the expert has to specify.
But doing so might preclude important class distinctions, resulting in a dramatic reduction in accuracy.
A better approach is provided by \emph{field guides} that experts write to train \emph{human} novices.
In these guides, for each class, the expert first identifies a very similar previously defined class, and then specifies \emph{only the most important attributes} that distinguish the two classes (Figure~\ref{fig:fig1} (right)).
This allows for concise descriptions that are both easy to write for the expert and also complete enough for the novice.
What if one used these kinds of descriptions for ZSL?

While intuitive, in our experiments we find that this field guide-based approach is even worse than simply choosing a \emph{random} smaller, fixed vocabulary of attributes to begin with.
This is because it assumes that what the expert
\emph{thinks} are important attributes, are in fact what the machine
finds relevant to make class distinctions. 
But machines are \emph{not people}; the attributes that humans find most salient may not in fact be salient to machines.
More broadly, what people find to be obvious class distinctions might appear extremely subtle to the machine and vice versa.
Thus, we need a new approach that similarly reduces annotator effort by focusing only on important attributes, but that crucially relies on the machine to define attribute importance.

With these considerations in mind, we propose a new \emph{learning interface} for ZSL learners to learn from experts.
For each novel class, the expert first identifies a close cousin that the learner already knows about.
The learner then chooses the attributes that it thinks will be most informative for it to learn and actively queries the expert for just these attributes.
This learning space brings to the fore the question of query strategies: how must the learner choose attributes to query that maximize performance (achieve good accuracy) but minimize expert effort (choose informative attributes)?
Similar problems are explored in active learning, where learners must choose unlabeled data points to label.
However, where active learning techniques have a priori access to unlabeled examples to make a range of measurements (e.g., prediction uncertainty), in our case the learner must choose attributes to query for a \emph{completely unseen class}.

We address this challenge by proposing multiple novel kinds of query strategies to learn from this new active ZSL
interface.
We design strategies based on a new measure of attribute uncertainty based on class taxonomies and a notion of expected change in model predictions.
We also design strategies for when we have access to an \emph{image} of the novel class, thus generalizing to the regime of combined zero- and few-shot learning.
Our proposed query strategies can work out-of-the-box with existing zero-shot learners.

We experiment with three datasets, namely CUB, SUN and AWA2, and show
significant reduction in annotation costs while keeping the
performance on par with the full annotation models.  With only $35\%$
of annotations, we get close to full model performance on SUN and CUB.
Our approach also significantly outperforms prior unsupervised ZSL
work on CUB and SUN \emph{without a single attribute annotation}.
Our contributions are:

\noindent $\bullet$ A new \emph{field-guide-inspired active ZSL} approach to collect
expert annotations that is more \emph{accurate} than using class
names/textual descriptions and more \emph{time-efficient} than
annotating a full attribute description.

\noindent $\bullet$ New query strategies (e.g., based on
uncertainty and expected model change), to actively query expert attributes
to rapidly train the learner.  Our results suggest that
thinking about what information to acquire from the annotator, and
what interface the expert uses, is a promising direction for building
accurate recognition models that are easy to train.

\section{Related Work} \label{sec:relwork}

\noindent{\bf Zero-shot learning.} In zero-shot learning
\cite{lampert-13,larochelle-08}, the model learns to classify unseen classes without any training images by leveraging side information like attribute descriptions.
Initial
work by Lampert ~\etal \cite{lampert-13} proposed first predicting
attributes from images and then classifying images based on the predicted
attributes. 
Recent work has looked at projecting image features into attribute space, and measuring
similarity with class descriptions~\cite{frome-13,akata-15,romera-15,kodirov-17,socher-13}.
More recent work has used the attribute description to produce improvements by embedding these descriptions as well as images into a shared feature space~\cite{xian-19,changpinyo-16,verma-17,norouzi-13}.
The auxilliary losses such as a reconstruction loss can be used
to regularize the problem~\cite{socher-13,xian-19}.  
Our proposed framework builds on these ideas, but leverages sparser but richer information about each class from the expert.

Because collecting attributes from experts is expensive, researchers have looked at other sources of class information.
These methods have been colloquially referred to as ``Unsupervised
ZSL'' as they do not require attribute supervision for test classes. 
They range from using word embeddings of the class names \cite{frome-13, rohrbach-10, akata-15}, class and attribute embedding \cite{alhalah-16}, or textual descriptions (usually from Wikipedia) \cite{elhoseiny-13, lei-15, alhalah2-17, qiao-16, zhu-18} of classes instead of attribute descriptions. 
Note that while these methods are called unsupervised, they still require information such as a large training corpus for word embedding or text articles (written by domain experts).
In addition, these sources of information are not typically designed for \emph{identifying} classes visually.
While we share the same goal of reducing annotation cost, we propose an alternative active-learning framework for attribute annotation.

Our paper also considers the possibility of going beyond zero-shot, and assuming that the annotator can provide us with one image of the novel class.
This notion of combining images with zero-shot attribute information was first introduced by Tsai \etal\cite{tsai-17}.
Schonfeld \etal\cite{schonfeld-19} introduce a VAE model to align the image 
features and attributes in a latent space to perform combined few-shot and zero-shot learning.
Since a field-guide has images as well as distinctive attributes,
models combining these two are worth exploring.

\noindent{\bf Active learning.} 
Active learning has been extensively explored in the the area of machine learning and computer vision.
These methods aim to make judicial use of an annotation budget by selecting useful unlabeled data to be labelled first.
Several methods have been proposed that pick data to be labelled based on objectives such as ``bringing larger expected model change'' \cite{settles-08, goodman-04}, ``increasing the diversity of labelled set''\cite{sener-17, yang-17}, ``reducing the uncertainty in predictions'' \cite{lewis-94, scheffer-01, gal-16}.
Recently many new techniques utilize adversarial learning to acquire labels for most informative data \cite{yu-19, sinha-19, zhang-20, wang-20}.
Our work is inspired by some of these techniques, but applies to a completely different and new problem: that of choosing \emph{attributes} to label. 
Since our methods do not have access to unlabelled data, the traditional active learning query strategies no longer apply.
Nevertheless, we take inspiration from this prior work to define a family of new techniques more suited to the zero-shot setup.

\noindent{\bf Learning from experts.}
There are other pipelines that are motivated by the question of learning from experts.
Misra et al.~\cite{misra-19} propose a visual question answering system that learns by interactions. 
Unfortunately, it is not clear if this system can generalize to the general recognition setting.
Earlier work leveraged experts to automatically correct part locations~\cite{branson-11} or define patches for fine-grained classification~\cite{patterson-13}.
However, these models rely explicitly on the part-based design of the recognition models.
There has also been work on recognition techniques that query humans during \emph{inference} for similarity comparisons~\cite{wah-14} and attribute descriptions~\cite{wah-13,branson-14}.
In contrast, we query the human annotator for particular attributes when we are learning a novel class.

\begin{figure}
\centering
\includegraphics[width=\linewidth]{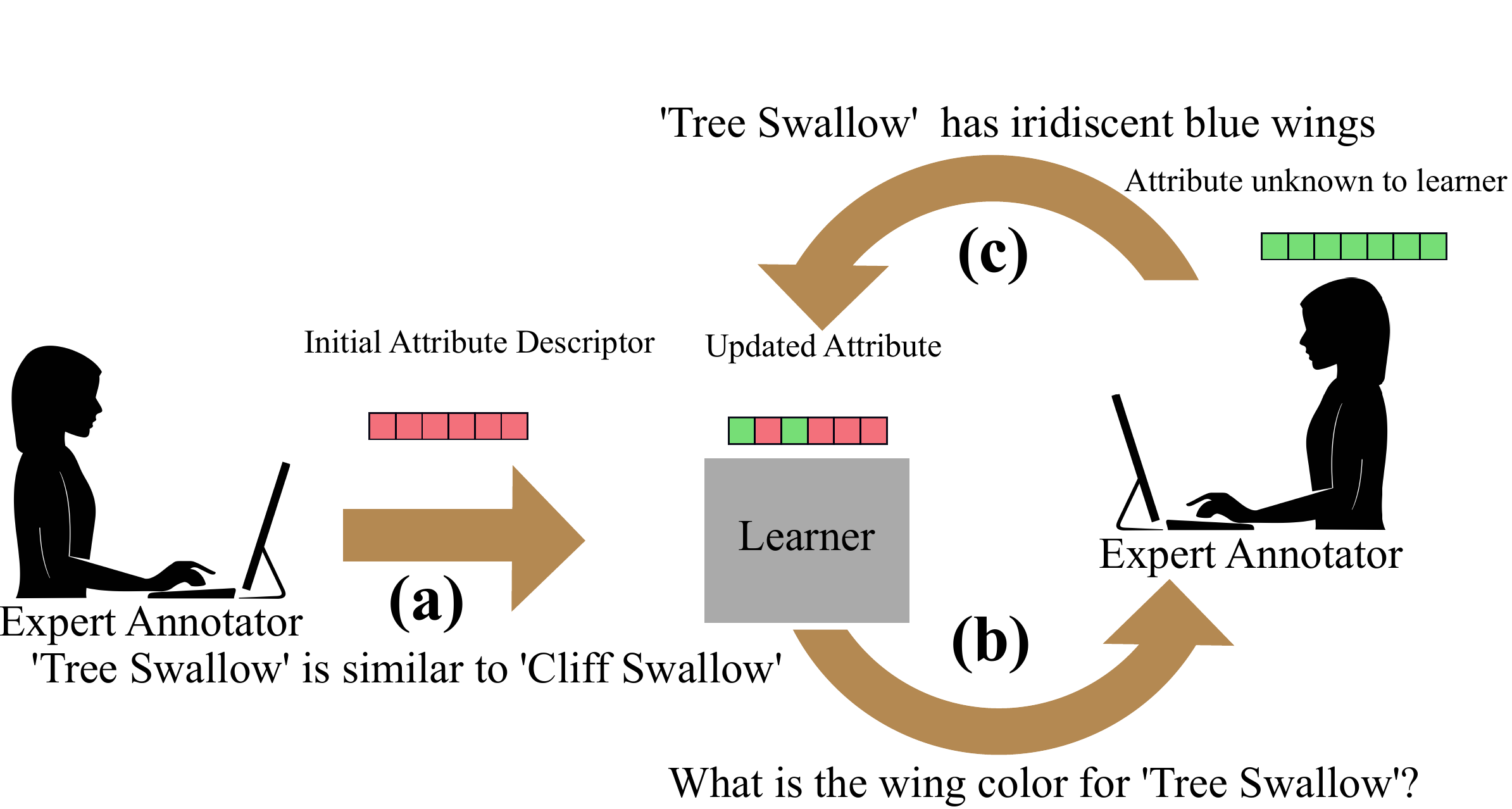}
	\caption{Overview of our \emph{field-guide-inspired} workflow. 
	(a) An expert annotator introduces the learner to a novel class by first providing a similar base class. 
	(b) The Learner then interactively asks for values of different attributes using an Acquistion function.	
	(c) The expert annotator provides the values of the
attributes and the learner updates
its state and class description. 
}
\label{fig:pipeline}
\vspace{-1.5em} 
\end{figure}
\section{Method} \label{sec:method}

\subsection{Problem Setup}
Our goal is to produce a learner that can learn new classes from very few interactions with an expert annotator.
As in traditional zero-shot learning, we assume that the learner is first trained on some set of ``base'' or ``seen'' classes $\mathcal{B}$.
For each base class $y \in \mathcal{B}$, the learner knows a vector of attributes $A(y) \in \mathbb{R}^d$, where $d$ is the number of attributes (we assume real-valued attributes in this work, in line with previous work on ZSL).
The learner is also provided with a large labeled \emph{training set} for the base classes consisting of images $\mathbf{x}_i, i=1, \ldots, n$ and corresponding labels $y_i$.

Once trained and deployed, the learner gets a set of hitherto unseen classes, $\mathcal{N}$ to recognize.
It is here that our problem setup diverges from traditional zero-shot learning.
A traditional zero-shot learning system needs the full attribute description of all novel classes: $\{A(y) \forall y \in \mathcal{N}\}$. 
In contrast, in our proposed setup, the learner must use \emph{very few}  \emph{field-guide}-like interactions with the expert annotator.

In particular, for each new class $y \in \mathcal{N}$, the annotator first gives the learner the most similar class $S(y) \in \mathcal{B}$ from the set of base classes.
Next, the learner incrementally asks the annotator for the values of attributes, one attribute at a time.
The goal of the learner is then to learn to recognize the novel class from as few attribute queries as possible.
We call this new kind of learning interface and the associated technique \zslinternospace (Figure~\ref{fig:pipeline}).

To design a learner in this interactive setup, two questions must be answered: (a) how should we learn with the incomplete attribute description?, and (b) how should we choose which attributes to query? 
We address the first question in the next section and then discuss our strategy for the more challenging question of choosing attributes to annotate.

\subsection{Learning from Sparse Attribute Annotation}
For every novel class $y$, the learner is told the most similar class $S(y)$, and it queries values for a subset of attribute indices $I(y) \subseteq \{0 \cdots d\}$.
For attributes where novel class information is missing, we impute the attribute descriptor using the values from $S(y)$.
\vspace*{-1em}
\begin{equation}
A'(y)[i] = \begin{cases}
A(y)[i] & i \in I(y) \\
A(S(y))[i] & i \notin I(y) \\ 
\end{cases}
\label{eq:zsldiff}
\end{equation}
\vspace*{-1em}

The imputed vector $A'(y)$ is used as the attribute descriptor for the novel class $y$ in the zero-shot model.
Note that this approach is model-agnostic and can be used with any zero-shot model out-of-the-box.

\subsection{Strategies for Querying Attributes}
We now describe how we pick the attributes iteratively to collect the
sparse attribute annotation.
Suppose the learner is learning about novel class $y$, with $S(y)$ as its most similar base class.
Suppose it has already queried for a subset $I(y)$ of the attributes 
resulting in an imputed attribute vector $A'(y)$.
Given this information, it must now choose a new attribute to query for.
The learner will make this choice using a \emph{query strategy} or \emph{acquisition function} $\pi$; the attribute chosen is $\pi(S(y), I(y), A'(y))$.

The notion of a query strategy is reminiscent of active learning, where the learner must choose unlabeled data points that it wants labeled.
However, in active learning, the learner has access to the unlabeled data, and so it can use its own belief about the unlabeled data points to make the call.
For example, it can choose to label data points for which it is most uncertain, or for which an annotation is most likely to change its state significantly.
In contrast, in our case, the learner is faced with a \emph{completely unseen class}.
How can the learner identify informative attributes when it has never seen this class before?

Below, we present two solutions to this challenge.
The first solution uses a taxonomy over the base classes to find informative attributes.   
The second uses the imputed attribute vector and looks at the change in representation space with respect to the change in a attribute.

\subsubsection{Taxonomy-Based Querying}
Class taxonomies are common in many domains (e.g., birds) and can be defined easily by a domain expert.
\emph{Siblings} in this taxonomy are likely to be similar to each
other, sharing several common attributes, and probably differing from each other in only a few attributes.
For example, in the domain of birds, all classes of the ``Oriole'' supercategory have black wings, a yellow body, and conical bills.
Therefore, it is not prudent to query for wing color, body color or bill shape for an Oriole class.
Instead we might want to query for the nape color, which varies among Orioles: ``Hooded Orioles'' have a yellower nape than ``Baltimore Orioles''. 
Thus, in general, we want to query attributes that vary a lot among classses in the relevant subtree, and ignore those that don't.

\begin{figure}
\centering
\includegraphics[width=\linewidth]{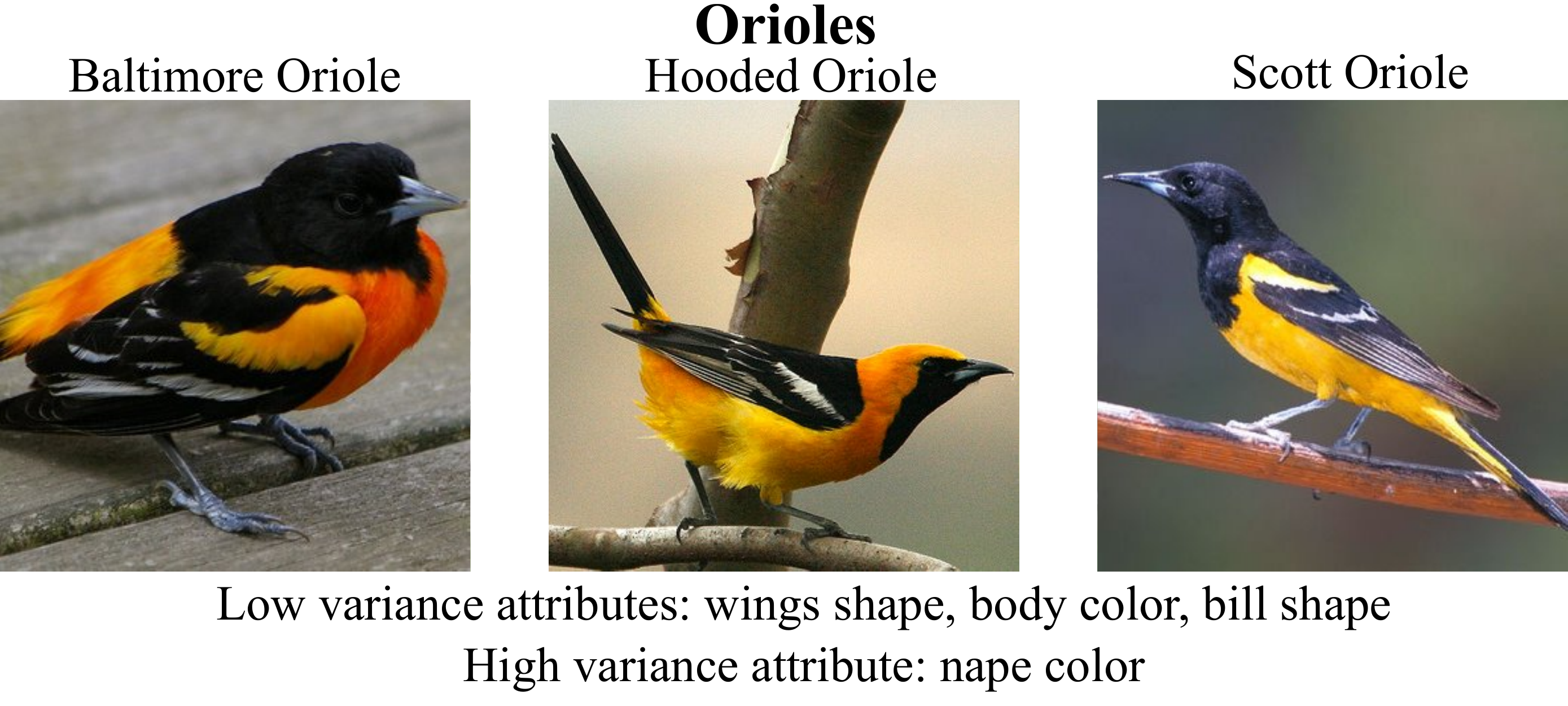}
	\caption{While attributes like ``wing shape'', ``bill shape'' are common to all orioles, color of the nape has high variance.}
\label{fig:illus}
\vspace{-1.5em}
\end{figure}

We can use this intuition to intelligently pick attributes to query by leveraging the class taxonomy.
Let $R_z$ be the set of sibling classes for a base class $z$ including the class $z$ itself. 
When the annotator provides the similar base class $S(y)$ for the novel class $y$, we 
look at the \emph{variance} of each attribute in sibling classes $R_{S(y)}$.
Attributes with lower variance are common to all the sibling classes
and will be less informative for a similar novel class $y$. 
In contrast, attributes with higher variance vary a lot in this subtree, so querying for their value is prudent.
Thus, the \emph{variance} for each attribute among the sibling categories $R_{S(y)}$ is a
measure of which attributes are more informative for this set of classes.

 We define this measure of variance amongst siblings as \emph{\sibvar} for novel class $y$ (denoted by $\sibvareq(y)$):

\vspace*{-0.5em}
\begin{equation}
\sibvareq(y)[j] = \textrm{Var}(\{A(s_i)[j]; s_i \in R_{S(y)}\}) 
\label{eq:sibvar}
\vspace{-0.5em}
\end{equation}
where $j$ indexes the attribute, Var denotes the variance of a set of values, and $s_i$ ranges over the siblings of $S(y)$

The most informative attributes are the ones with maximum \emph{\sibvar}.
We therefore choose attributes to label in \emph{decreasing order of \sibvar}:

\vspace*{-1em}
\begin{equation}
\sibvarpi(S(y), I(y), A'(y)) = \argmax_{j \notin I(y)} \sibvareq(y)[j]
\label{eq:zsldiff}
\end{equation}
\vspace*{-1em}

Attributes sometimes belong to groups.
For example, in CUB, many different attributes all correspond to body color: each attribute captures a different color, and together the body color attributes define a multinomial distribution over the different colors.
For such groups of attributes, we only look at the maximum varying attribute to find \emph{\sibvar}.
Thus, if $g$ is such a group of attributes, \emph{\sibvar} for this group is be defined as, 

\vspace*{-1em}
\begin{equation}
\sibvareq(y)[g] = \max_{j\in g}\textrm{Var}(\{A(s_i)[j]; s_i \in R_{S(y)}\})
\label{eq:sibvar}
\end{equation}
\vspace*{-1em}

We select the whole group of attributes with maximum \emph{\sibvar}
for annotation.  

Note that this method resembles uncertainty measure-based
active learning methods.  The learner first picks attributes for
annotation where the learner is uncertain.  
But the way we measure this uncertainty is different from these methods, as we have no information about the novel class.

\subsubsection{Querying Based on Representation Change}
Many zero-shot learners encode both the attributes and images in a
common latent space, and train classifiers 
in this space.  
Attribute encoders in these approaches can learn to perform a variety of useful functions, such as weighing attributes more if they are more identifiable or more discriminative.

Because the classifier operates in this latent space, annotations that significantly change this latent representation are more likely to influence classification decisions.
This suggests that we should query attributes which when changed cause the largest change in latent space.
We call this method of looking at changes in representation ``\emph{\encchange}'' denoted by $\encchangeeq$.

Concretely, let $\mathbf{E_a}: \mathbb{R}^d \rightarrow \mathbb{R}^l$, be the attribute encoder function that maps an attribute into a latent representation space of dimension $l$.
Suppose that $A'(y)$ is the current imputed attribute representation (note that $A'(y)$ starts as $A(S(y))$ and is gradually filled in with the true attribute values $A(y)$ as the learner queries the annotator).
This attribute vector is represented in latent space as $ \mathbf{E_a}(A'(y))$.
Now we wish to know which attributes when perturbed will lead to the largest change in this representation.
Since the encoder architectures are non-linear, we make use of the local partial derivatives as represented by the Jacobian of the encoder to measure these changes.
\vspace*{-1em}
\begin{equation}
\mathbf{J_{E_a}} (\mathbf{x}) = \left[
			\frac{\partial \mathbf{E_a}(\mathbf{x}) }{\partial \mathbf{x}_1}
			\cdots 
			\frac{\partial \mathbf{E_a}(\mathbf{x}) }{\partial \mathbf{x}_d} 
			\right]
\label{eq:jacobi}
\end{equation}
\vspace*{-1em}

We define the {\emph{\encchange}} \ of attribute $i$ as:
\vspace*{-0.5em}
\begin{equation}
\mathit{\encchangeeq}(y | A'(y))[i] =\left\| \left. \frac{\partial \mathbf{E_a}(\mathbf{x})}{\partial \mathbf{x}_i} \right|_{\mathbf{x}=A'(y)}\right\|_2
\label{eq:jacobi}
\end{equation}
\vspace*{-1em}

The l2-norm of the partial derivative measures how much a small
perturbation to that particular attribute changes the encoding in the
latent space.  The learner then queries the attribute with the maximum \emph{\encchange}.

\vspace*{-1.5em}
\begin{equation}
\encchangepi(S(y), I(y), A'(y)) = \argmax_{j\notin I(y)} \mathit{\encchangeeq}(y | A'(y))[j]
\label{eq:zsldiff}
\end{equation}
\vspace*{-1em}

For grouped attributes, similar to
\emph{\sibvar}, we measure \emph{\encchange} to be 
the maximum \emph{\encchange} within the attributes of that group.

\emph{\encchange} is similar to the ``expected model change''-based active
learning methods\cite{settles-08, goodman-04}.
These methods pick entities based on the most change to the model in expectation.
Similarly, \emph{\encchange} picks attributes that change the encoding the most in latent space.

\subsubsection{Querying Using Image Data}
In practice, an expert can easily provide a single image of the novel class. 
Only recently have ZSL techniques begun to leverage this information\cite{tsai-17, schonfeld-19}.
Here, we show how we can use this single image to better choose attributes to label, using the following \emph{\imagebased} strategy.subsection

The key intuition here is that the image provides a representation for the class that must ultimately match the final attribute description.
As such, the learner can \emph{recognize} attributes in the image, and attempt to reconcile differences between the recognized attributes and the imputed attribute description based on the expert annotation.

Concretely, suppose the image available for the novel class $y$ is $x_y$.
Suppose also that we have a trained \emph{image encoder}  $\mathbf{E_i}$ that maps the image to the latent space, and an \emph{attribute decoder} $\mathbf{D_a}$ that decodes this latent representation into an attribute vector; many recent methods train these~\cite{schonfeld-19}.
Using these modules, the learner can get an attribute vector from the image: $\tilde{A}(y) = \mathbf{D_a}(\mathbf{E_i}(x))$.

In general, $\tilde{A}(y)$ will not match the imputed attributes $A'(y)$.
Hypothetically replacing the $i$-th imputed attributes in $A'(y)$ with the image-derived counterpart in $\tilde{A}(y)$ would produce a new attribute description $A_i(y)$:
\vspace*{-0.5em}
\begin{equation}
A_i(y) = \begin{cases}
A'(y)[j] & j \ne i\ \\
\mathbf{D_a}(\mathbf{E_i}(x_y))[j] & j=i \\
\end{cases}
\end{equation}
\vspace*{-1em}

\emph{\imagebased} strategy then picks the attribute $j$ which maximally pushes the embedding of the hypothetical attribute vector $A_j(y)$ closest to the novel class image encoding:

\vspace*{-0.5em}
\begin{align}
\mathit{\imagebasedeq}(y | A'(y), x_y)[i] = ||\mathbf{E_a}(A_i(y))-\mathbf{E_i}((x_y))||_2^{-1}\\
\imagebasedpi(I(y), A'(y), x_y) = \argmax_{j\notin I(y)} \mathit{\imagebasedeq}(y | A'(y), x_y)[j]
\label{eq:imagebased}
\end{align}
\vspace*{-1.5em}

\begin{figure}[h!]
\centering
\includegraphics[width=0.45\linewidth]{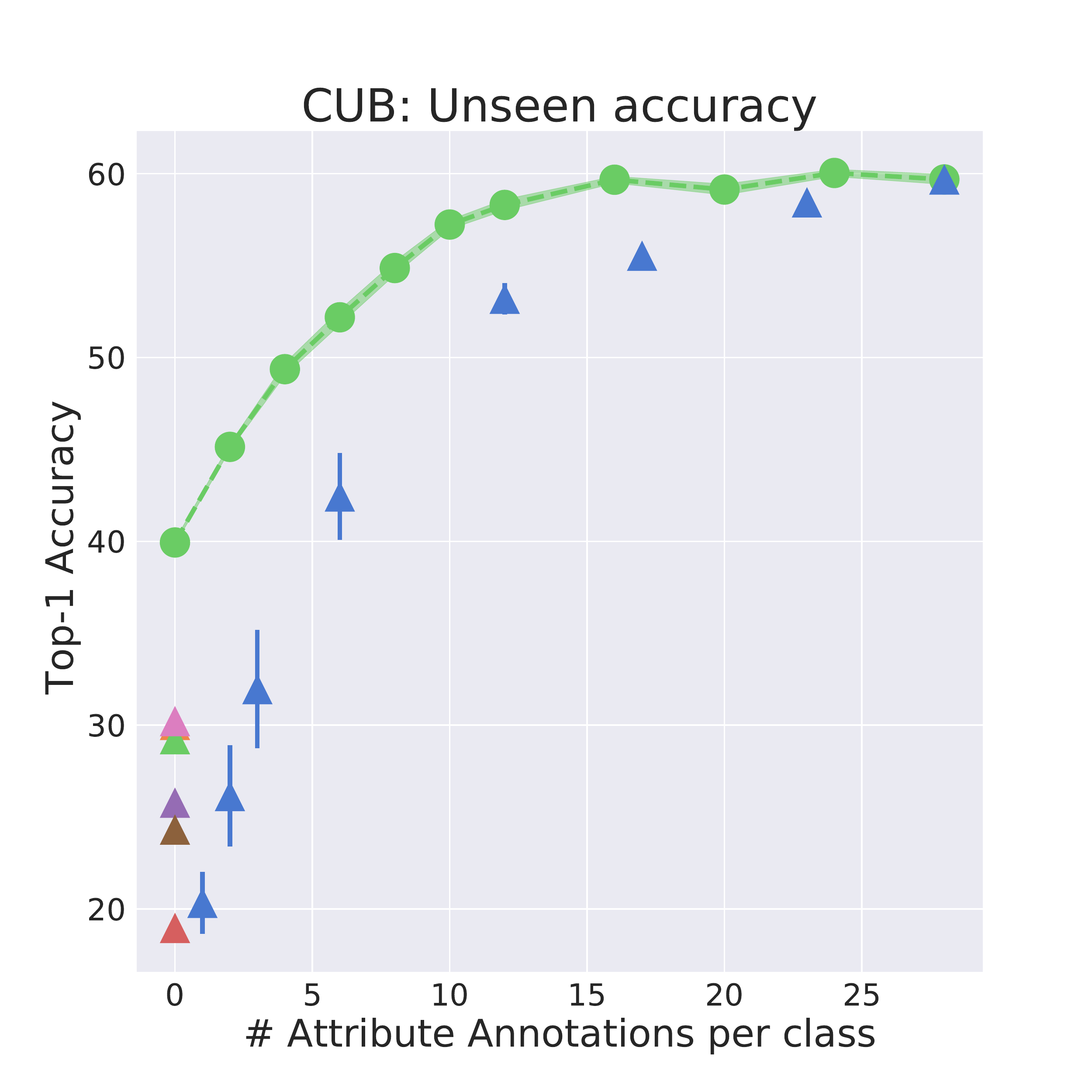}
\includegraphics[width=0.45\linewidth]{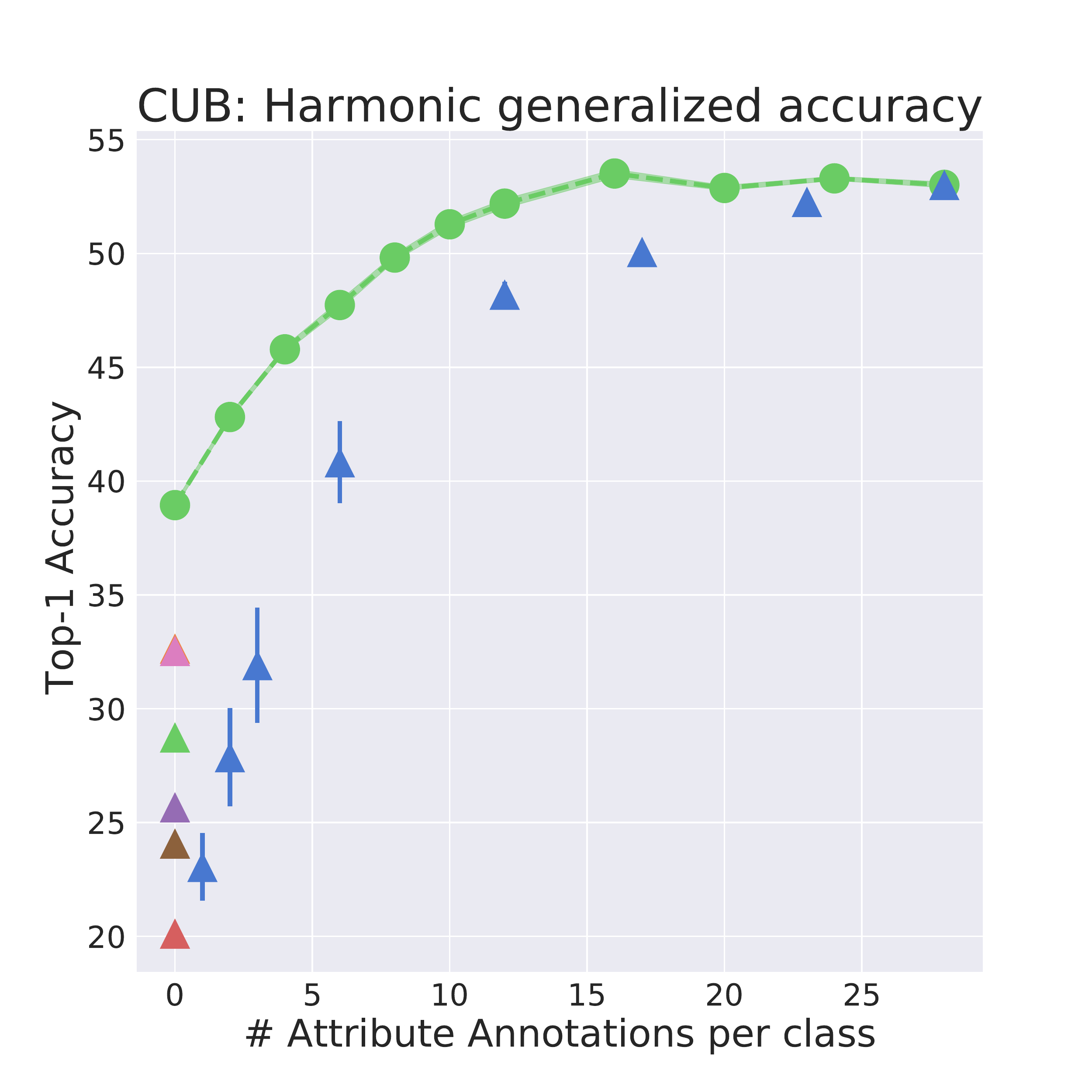}
\includegraphics[width=0.45\linewidth]{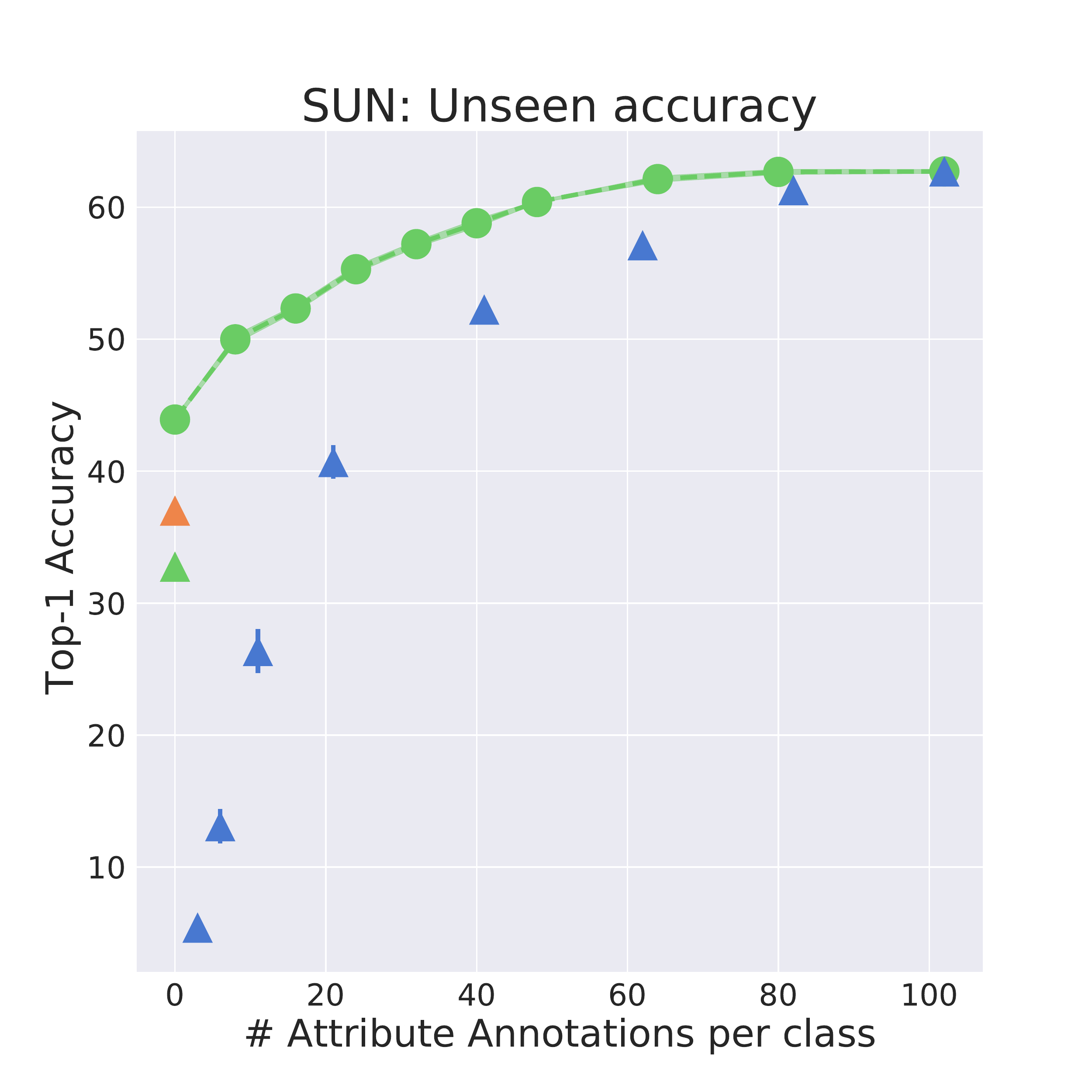}
\includegraphics[width=0.45\linewidth]{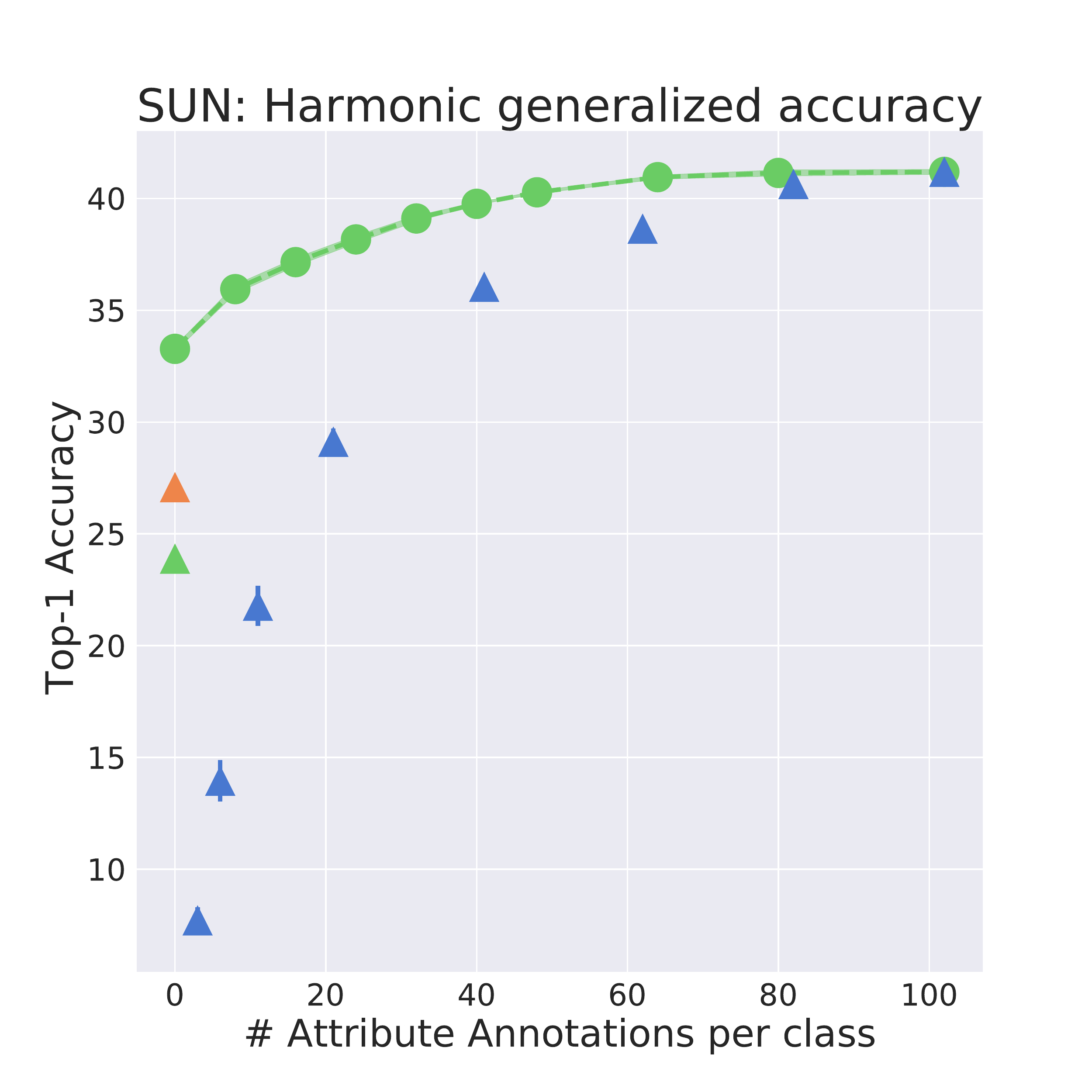}
\includegraphics[width=0.45\linewidth]{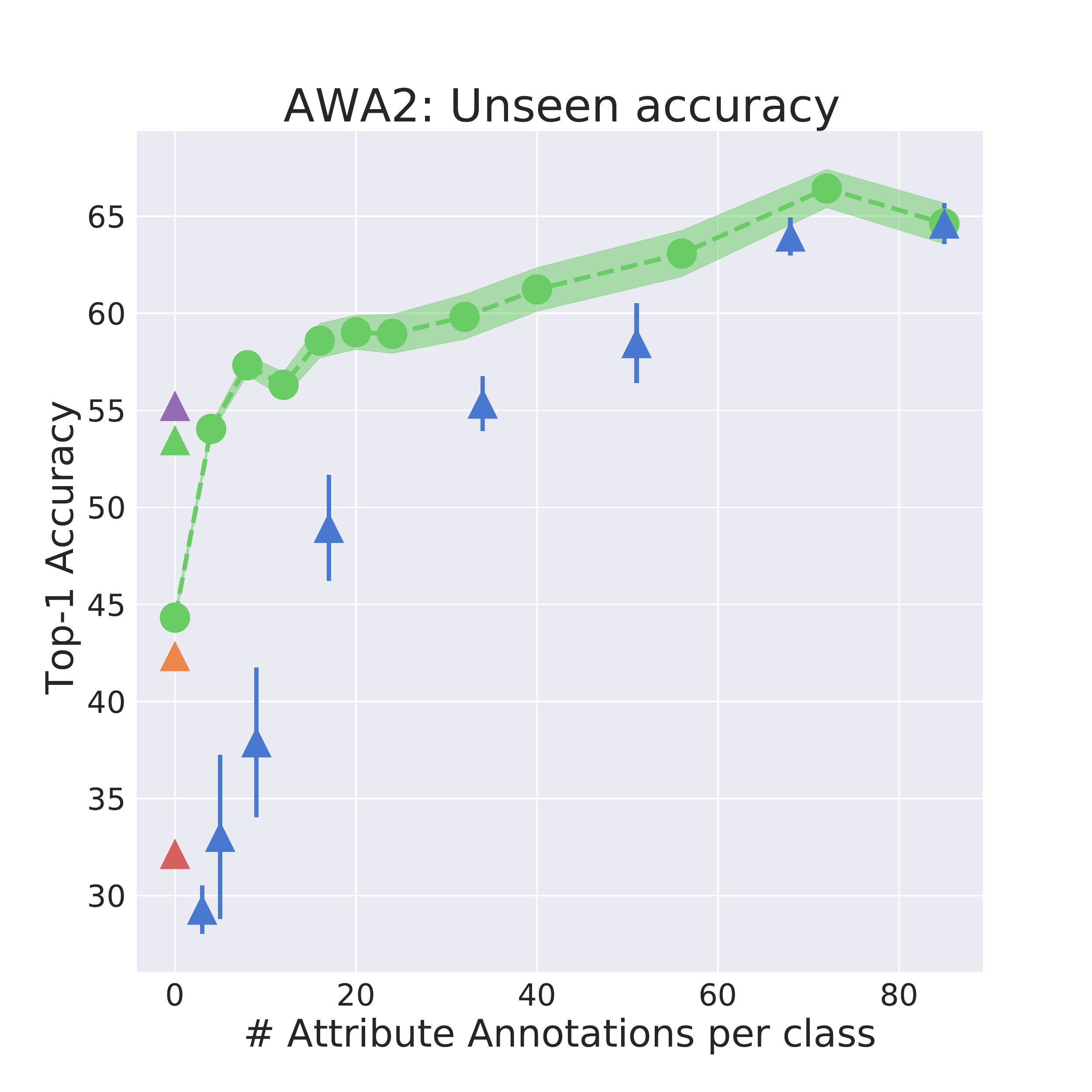}
\includegraphics[width=0.45\linewidth]{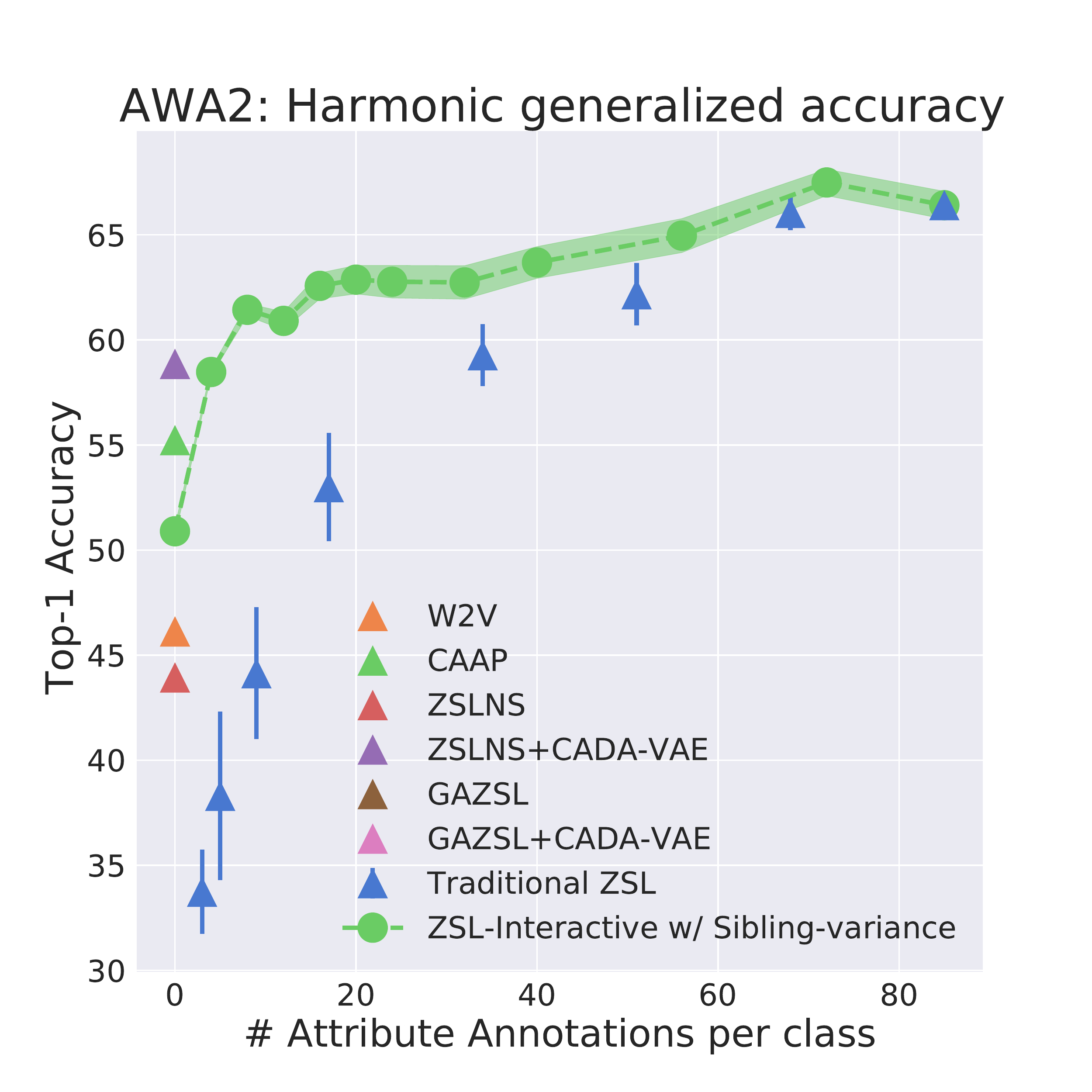}
	\caption{Performance of our method in terms of top-1 per-class
classification  accuracy vs. number of annotations provided by the
annotators during deployment with CADA-VAE as base model. \zslinter performs better than the traditional ZSL on all benchmarks and better than unsupervised baselines on CUB and SUN, proving its effectiveness.}
\label{fig:quantitative}
\vspace{-2em} 
\end{figure}

\section{Results} \label{sec:results}

\subsection{Dataset and Implementation Details}
For all experiments we use $2048$-dimensional features from ResNet-101 \cite{he-16}.
We compare our method on three zero-shot benchmark datasets:
CUB-200-2011 \cite{wah-11} (CUB), Animals with Attributes 2
\cite{xian-18} (AWA2) and the SUN attribute dataset \cite{patterson-12} (SUN). 
CUB is annotated with 312 part-based attributes; AWA2 and SUN have 85 and 102 attributes respectively.
These attributes are labelled for every class.
We use the standard train-test split from \cite{xian-18} for all benchmarks.

\noindent\textbf{Expert annotations of most similar base class:} 
Our problem setup requires that the annotator provide the most similar base class for 
each novel class. 
 Since CUB is a specialized domain requiring bird expertise, we worked with a professional bird watcher to annotate this information for every novel class.
AWA2 and SUN do not require expert knowledge,
so we collected the ``similar class'' information using 3 annotators.
We took a majority vote, and in cases when all 3 disagreed we asked
them to come to a consensus. 
We will release these expert
annotations publicly upon acceptance.

\noindent\textbf{Taxonomy}: For SUN, the taxonomy is already
available along with the dataset.
We manually created a taxonomy for AWA2 by looking at the family in
biological nomenclature.  For CUB, we use the general class name as the
parent in the taxonomy (Hooded Oriole $\rightarrow$ Oriole). 

\noindent\textbf{Base zero-shot learner}:
To show generalization of our method over different models we experiment with two base zero-shot learners: CADA-VAE\cite{schonfeld-19} and TF-VAEGAN \cite{narayan-20}.
CADA-VAE trains two variational auto-encoders on the base classes to learn a common
embedding space for attribute descriptions and images.  
It then trains novel class classifiers in this latent space.
All the hyperparameters for the architecture and training
are kept as prescribed by \cite{schonfeld-19}.  
TF-VAEGAN uses a VAE-GAN \cite{larsen-2016} to generate realistic features from attributes and uses these generated features from unseen class to train a classifier.
We show the results averaged over 6 different runs of the model.

\noindent\textbf{Metrics}:
We measure the per-class classification accuracy over unseen classes
and the harmonic mean of seen and unseen classification accuracy for generalized zero-shot learning.  
We plot these metrics as a function of the annotation budget for each approach.

\subsection{Is interactive ZSL accurate and time-efficient?}
We first ask how our proposed interactive setup compares with prior work on ZSL in terms of accuracy and the burden of annotation.
We compare our proposed setup against the following baselines:

\begin{figure}
\centering
\includegraphics[width=0.45\linewidth]{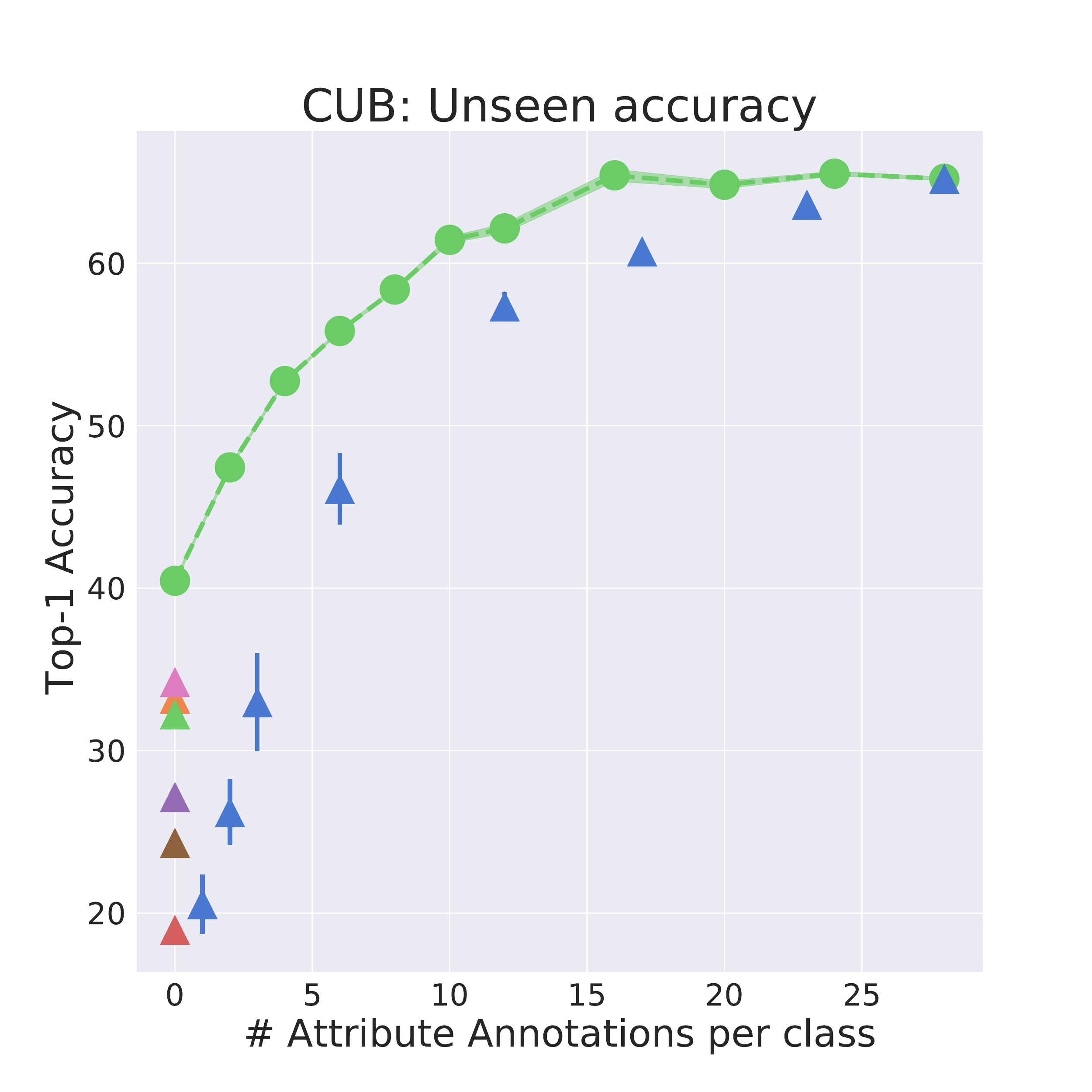}
\includegraphics[width=0.45\linewidth]{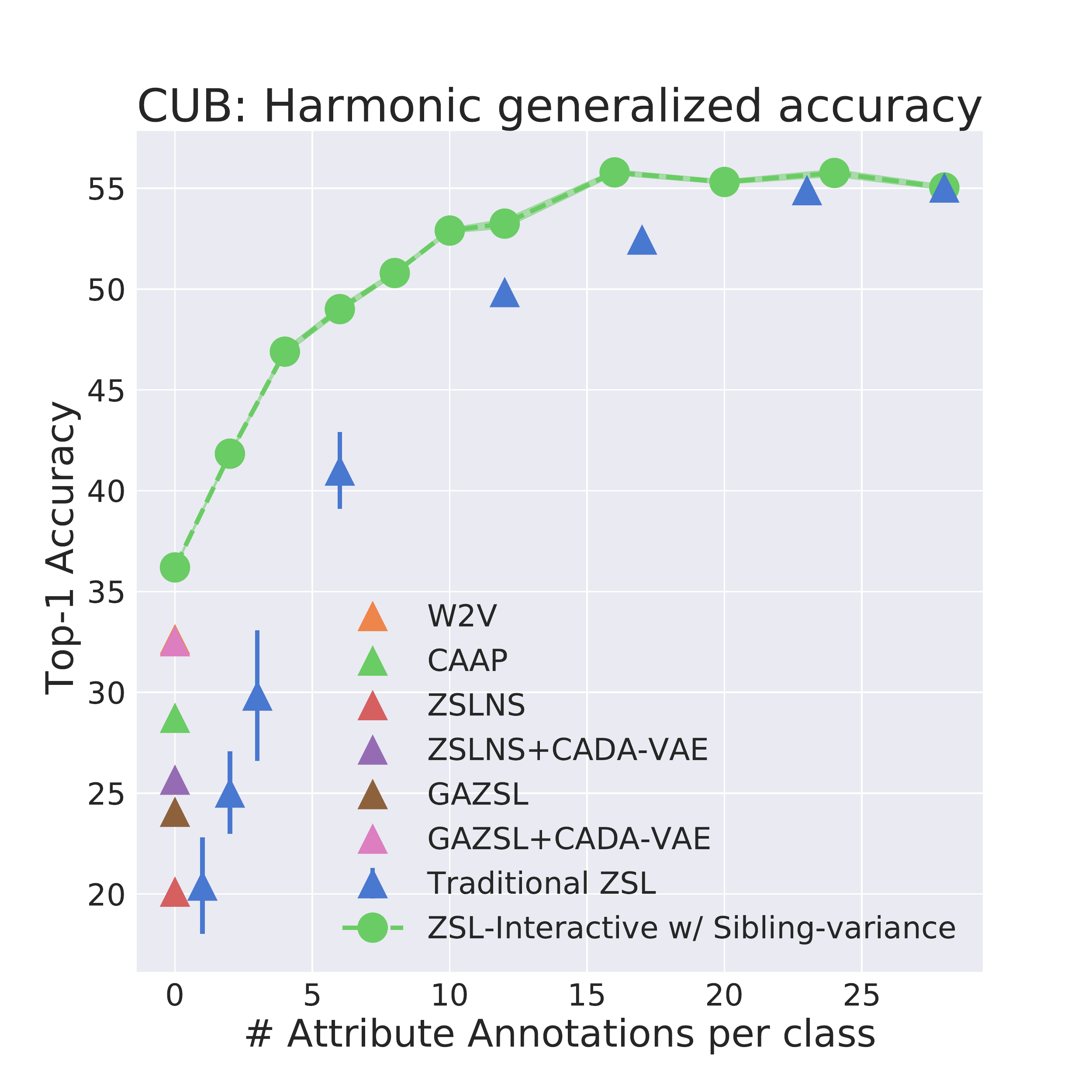}
	\caption{Performance of our method with TF-VAEGAN as base model. \zslinter performs better than the traditional ZSL and unsupervised baselines, proving the generalizability of \zslinter to other zero-shot models.}
\label{fig:quantitative_tfvaegan}
\vspace{-1em} 
\end{figure}

\noindent $\bullet$ \textbf{Traditional ZSL} with full attribute annotation.
Reducing annotation effort here can only be achieved by using a smaller attribute vocabulary.
Given an annotation budget, a correspondingly smaller subset of
attributes is chosen uniformly at random, and a new ZSL model is
retrained with the chosen attributes.  In deployment, all classes must
be described with this reduced vocabulary.

\noindent $\bullet$ 
\textbf{W2V}: This unsupervised ZSL approach uses word2vec embedding vectors of the classes instead of attribute vectors~\cite{akata-15}. 

\noindent $\bullet$ 
\textbf{CAAP}: This approach uses word embeddings for the classes and
the attributes to find the attribute vector for unseen classes \cite{alhalah-16}. 

\noindent $\bullet$ 
\textbf{ZSLNS~\cite{qiao-16} and GAZSL~\cite{zhu-18}}: use wikipedia
articles instead of attributes. Since the original papers use older
backbones for ZSL, we show the number for both the original model and
using ZSLNS (or GAZSL) extracted descriptors with newer backbones
(CADA-VAE and TF-VAEGAN).

While the last three are often deemed unsupervised, they do require
either a large corpus of text with attributes and classes to learn
word embeddings, or carefully curated text articles typically edited
by experts, thus indirectly requiring significant expert time, which we aim to minimize.

Note that reported numbers for unsupervised ZSL approaches typically
use a different class split that ensures that each novel class has a
closely related base class.  Instead, we use the proposed splits in
the attribute-based ZSL benchmark~\cite{xian-18}. We have used the
original code provided by the authors and changed the splits wherever
available.

We compare our best performing acquisition function (\emph{\sibvar}),
to the above baselines.  We plot the accuracy as a function of the
amount of annotation in Figure~\ref{fig:quantitative} (CADA-VAE) and
Figure~\ref{fig:quantitative_tfvaegan} (TF-VAEGAN). We find:

\noindent $\bullet$ 
\textbf{The expert annotated closest base class is extremely
informative}. With just that one annotation, one can recover almost
two-thirds of the performance of a full zero-shot learning system with
access to hundreds of attributes. Note that using a randomly sampled
class as the ``most similar class'' instead of the expert annotated
one yields very poor accuracy
($<$20\%) indicating that the expert-provided information is critically important.

\noindent $\bullet$ \textbf{Our approach can dramatically reduce annotator burden}. On CUB and SUN, our approach with only a third of the annotations (just 10 interactions per class) is as good as traditional ZSL with \emph{all} of the annotations. 

\noindent $\bullet$ \textbf{Our approach generalizes to other ZSL models}. Our method performs better than the baselines even when TF-VAEGAN is used as the base model Figure~\ref{fig:quantitative_tfvaegan}. This shows that as new ZSL methods develop, our method should generalize and can be used out-of-the-box with them. See supplementary for TF-VAEGAN results on other datasets.

\noindent $\bullet$ 
\textbf{Partial attributes more informative than unsupervised ZSL}.
While some methods perform better than our method with less
information on AWA2, our method beats all the baselines on CUB and
SUN. It is relatively hard to find discriminative information in
corpora like wikipedia for fine-grained categories like those in SUN
and CUB. This is the reason attribute-based systems are essential
when performing ZSL in fine-grained domains. Just providing most
similar class annotation (\emph{i.e., the field guide approach}) turns out to be significantly more useful
than using text or word embedding as attributes.

\begin{figure}[h!]
\centering
\includegraphics[width=0.45\linewidth]{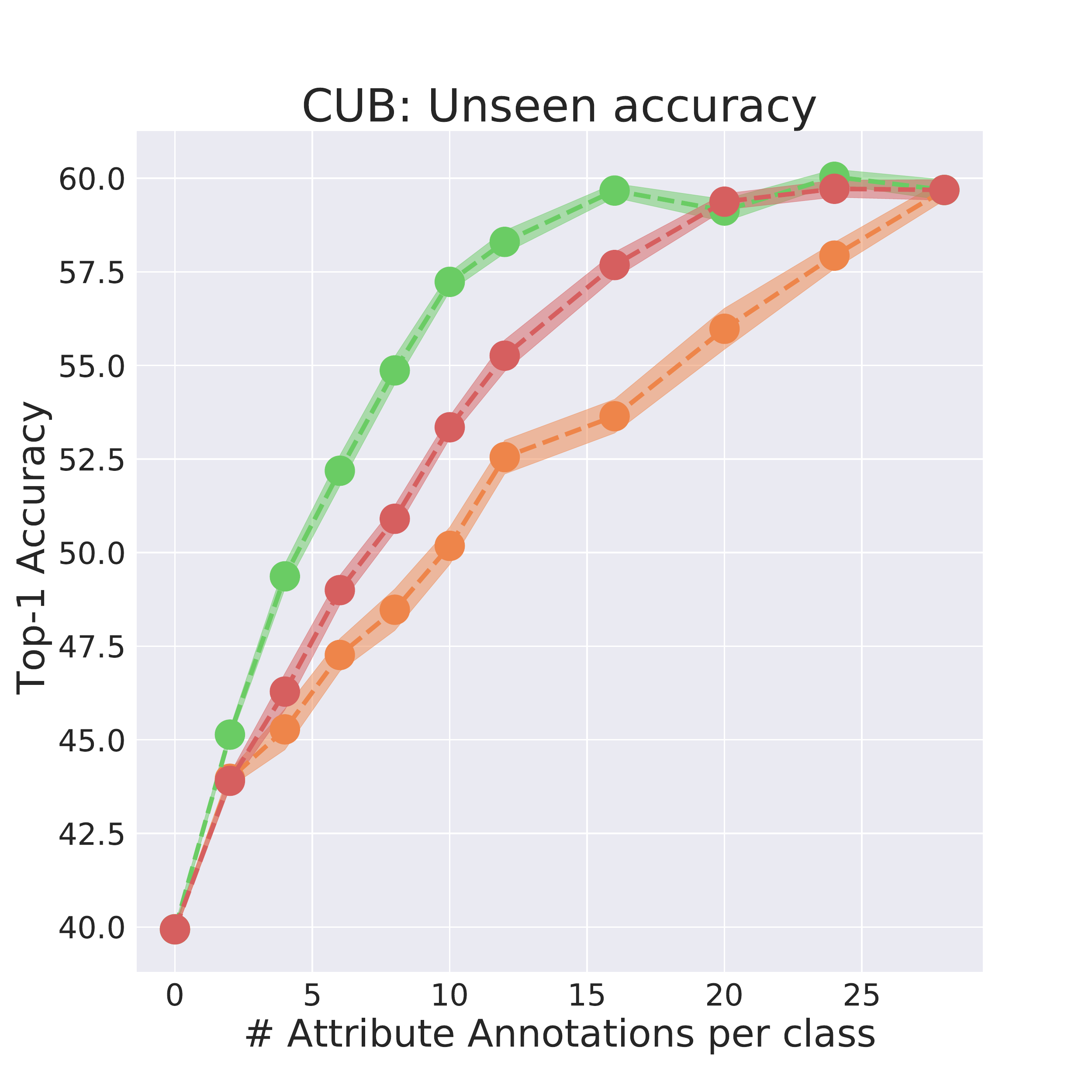}
\includegraphics[width=0.45\linewidth]{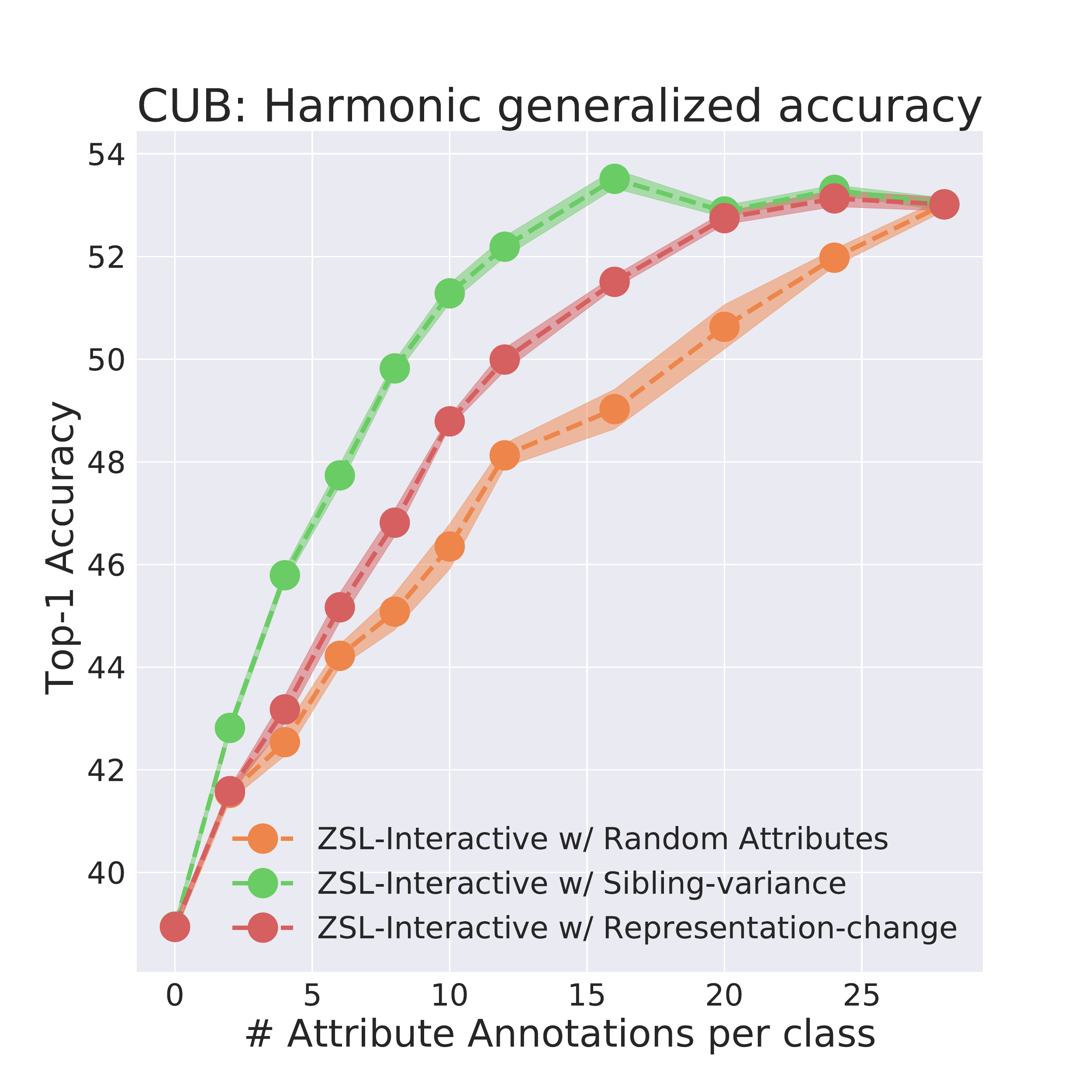}
	\caption{Performance of the two acquition functions with
CADA-VAE on CUB. Both functions perform better than the random
acquisition function. {\emph\sibvar} performs better than
\emph{\encchange}, but the latter does not require taxonomy information. See supplementary for results on other datasets.}
\label{fig:acquisition} 
\vspace{-1.5em} 
\end{figure}

\subsection{How do different acquistion functions perform?}
Our method performs better than baselines primarily for two reasons, the field-guide interface and an intelligent acquisition function.
We evaluate how different acquision functions perform with attribute annotation costs.
Among the various query strategies we have proposed, in general, the best performing query strategy is \emph{\sibvar}. 
\emph{\encchange} tends to have a weaker performance than
\emph{\sibvar} but it does not require the taxonomy to function and hence is still useful.
Figure~\ref{fig:acquisition} compares the 2 acquistion function against a random acquistion function as a baseline on CUB with CADA-VAE (see supplementary for other models and datasets).
Both of our proposed query strategies far outperform random attribute selection as well as the traditional ZSL pipeline, \emph{no matter what the labeling budget is}. This is true both for the unseen-only evaluation as well as the general evaluation.

\subsection{Can we do better with images?}
Figure~\ref{fig:oneshot} shows the performance of our approach when one image is given by the annotator along with the interactive attributes annotations for CUB.
Almost all of our conclusions from the previous section carry over, with the exception that 
\emph{\encchange} starts weaker than the baseline of choosing attributes to query randomly, but performs better than it in the later stages.
Additionally, note that \emph{\imagebased} performs on par with the \emph{\sibvar} without using any additional taxonomy information.
This suggests that using the inconsistency between image-derived and imputed attributes to determine what attributes are useful is a viable approach to interact with annotators.
See supplementary for results on other benchmarks.

\begin{figure}[h!]
\centering
\includegraphics[width=0.45\linewidth]{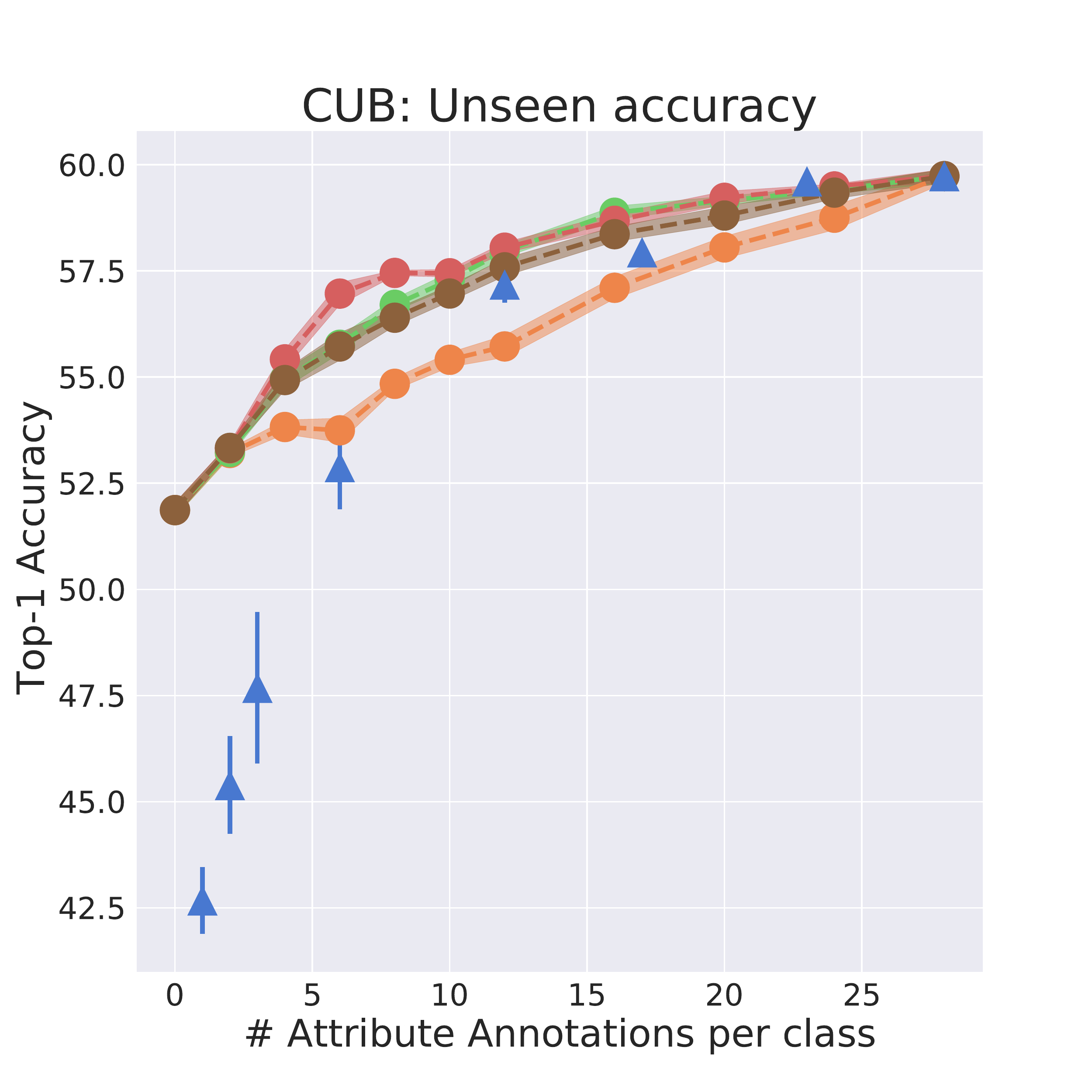}
\includegraphics[width=0.45\linewidth]{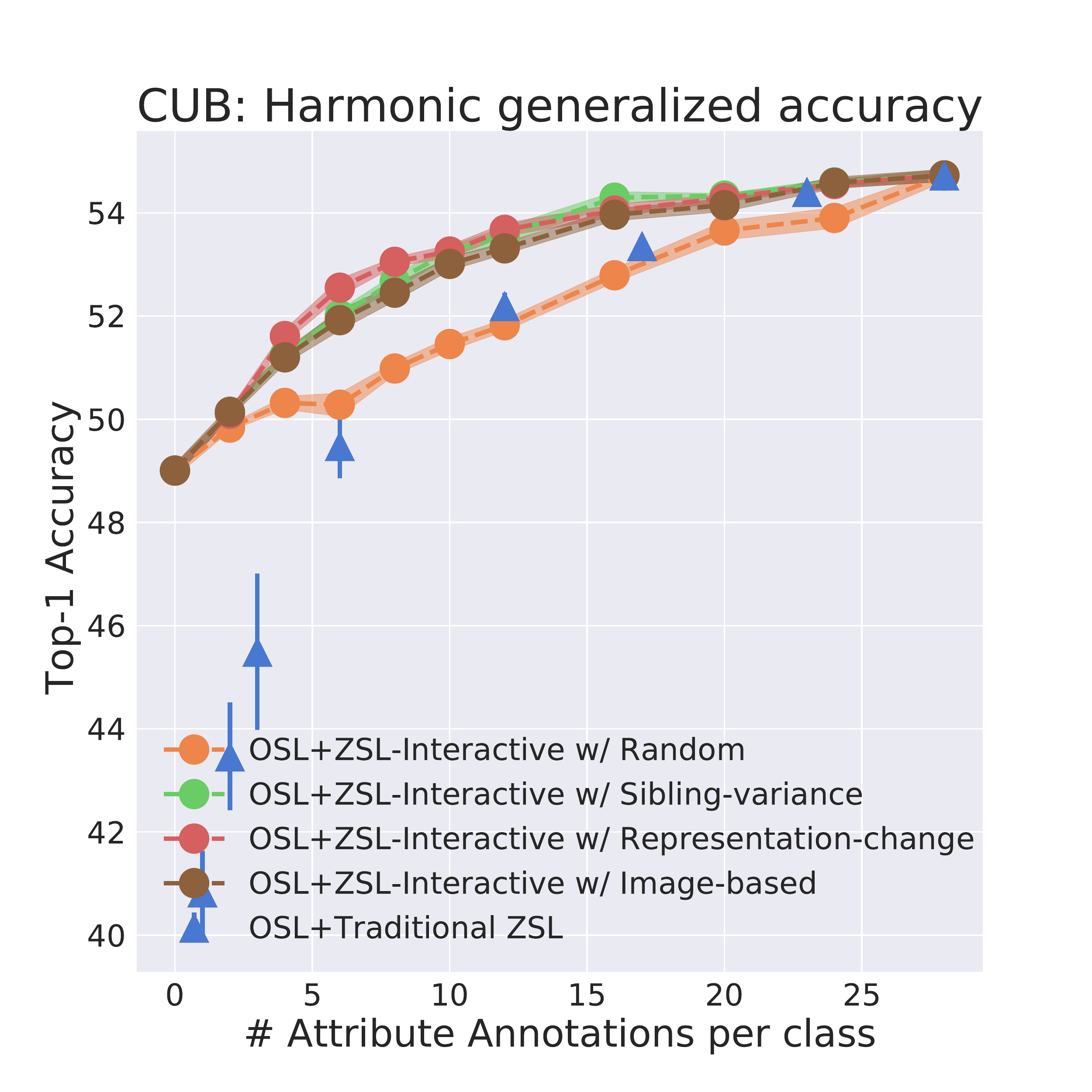}
	\caption{Performance of our method, zero+one-shot
setting, where the annotator provides a single image for the novel class
along with the interactively queried attributes. All methods perform
better than the baselines; \emph{\imagebased} performs on par
with \emph{\sibvar} without requiring an additional taxonomy over the
base classes.}
\label{fig:oneshot}
\vspace{-1em} 
\end{figure}

\subsection{Are acquisition functions better than experts?}
One might question how our interactive
method compares to collecting information from an expert.
As discussed in section~\ref{sec:intro} a field-guide provides what the expert finds to be distinctive attribute differences without any interaction. 
We evaluate how this compares to the learner actively querying.

To this end, for 20 of the 50 novel classes the CUB dataset, we additionally ask an expert to identify the 10 most informative attributes that distinguish each novel class from its most similar base class, in order of importance.
We use this infomation along with the similar class to construct the expert attribute baseline.

\begin{figure}[h!]
\centering
\includegraphics[width=0.45\linewidth]{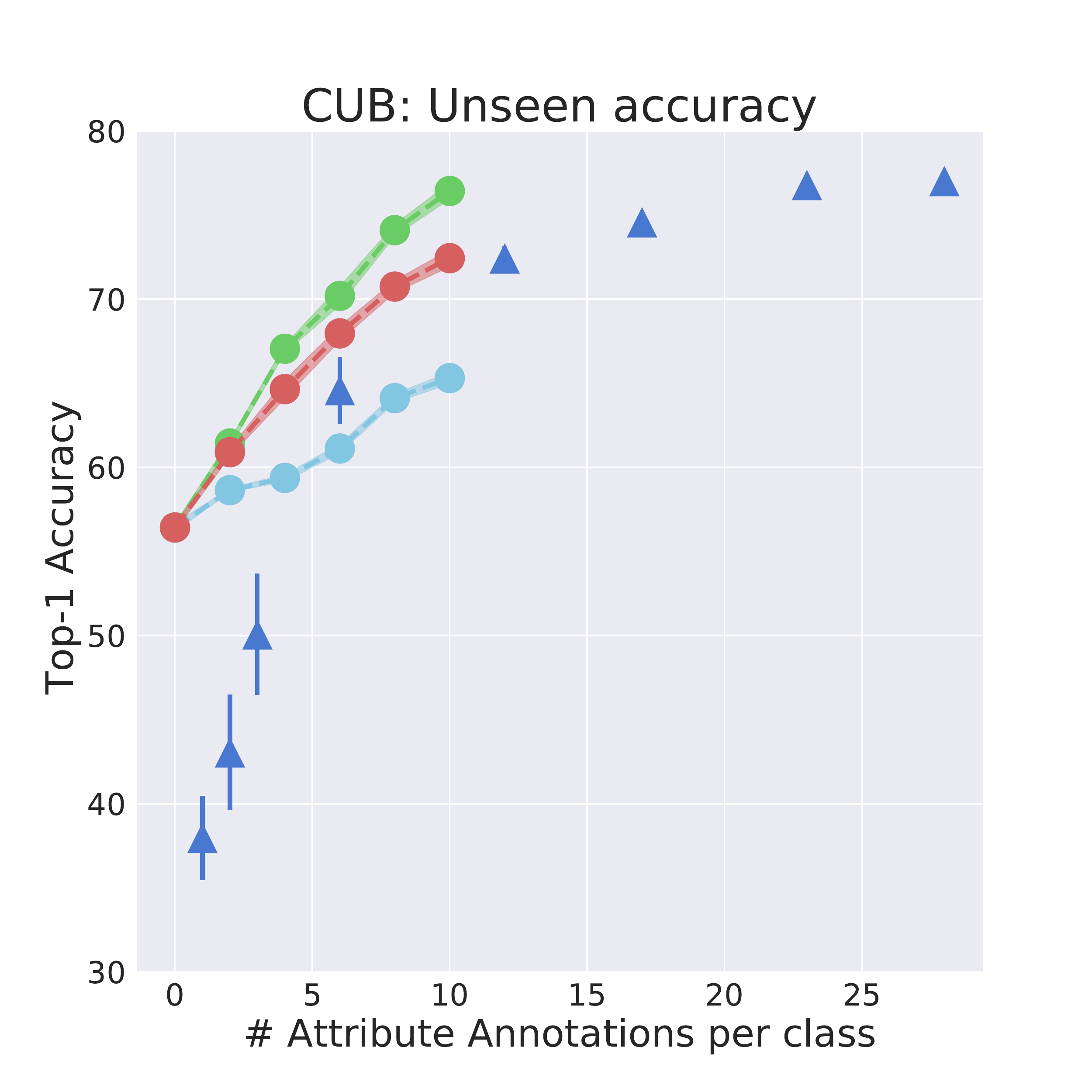}
\includegraphics[width=0.45\linewidth]{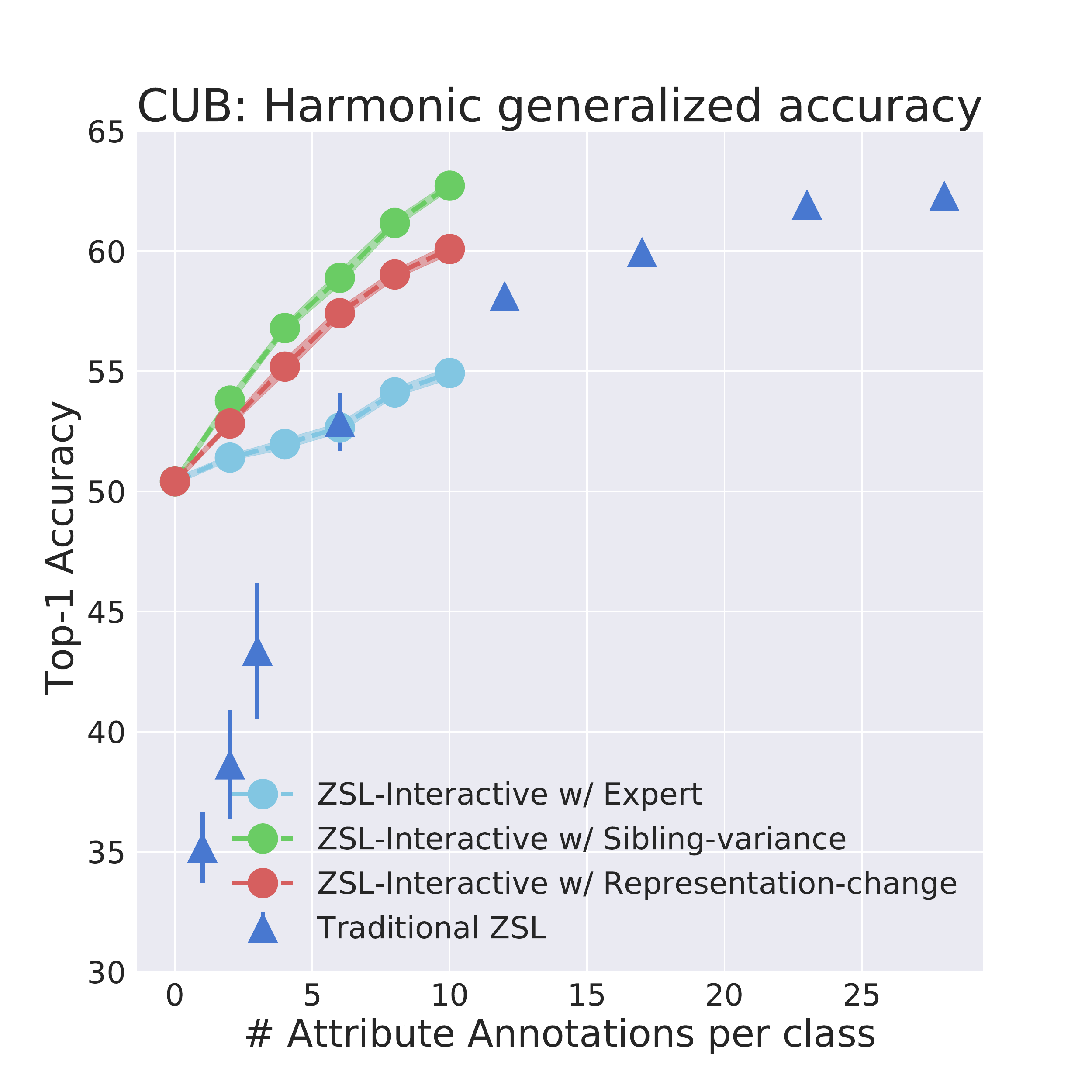}
	\caption{Performance of interactive methods against the expert
selected attributes for CUB (20 novel classes). 
The learner selecting attributes performs better than the expert providing the attributes.}
\label{fig:expert}
\vspace{-1em} 
\end{figure}

We find that the interactive methods are significantly better at asking for useful
attributes than what the domain expert gives (Figure~\ref{fig:expert}). 
This is a surprising result, as it shows that zero-shot learners learn about the domain differently than a human expert.
In the set selected by experts, the experts give importance to attributes like ``bird size'' and ``eye color''. 
While this information is useful to classify and differentiate between birds for a bird watcher, the model finds it difficult to understand the size of the bird as it is a relative property. 
Other attributes like ``eye color'' are very small in images (or
obstructed) so the network cannot 
necessarily utilize that knowledge during representation learning.
Hence, providing these attributes to the network is not very informative and it is better to let the learner select informative attributes.

\subsection{Is the query selection good?}

We look at the types of attributes queried by our method
\emph{\sibvar} for different parent categories of CUB.
The attributes queried should vary based on the sibling classes.
Figure~\ref{fig:attribute}, shows the top-3 attributes queried first
by the \emph{\sibvar} method for Swallows and Cuckoos.  It also shows
the top-3 attributes picked by measuring variance over all 
classes.  While bill length and overall shape have high variance over all
classes, within Swallows (and within Cuckoos) the variance is not big
and therefore should not be annotated first.  For swallows, it can
be seen from the image that throat, crown and forehead color varies
within category, and hence should be acquired first.  For Cuckoo,
underpart, crown and belly color varies (See supplementary for 
other examples and datasets).

\begin{figure}[t]
\centering
\includegraphics[width=.8\linewidth]{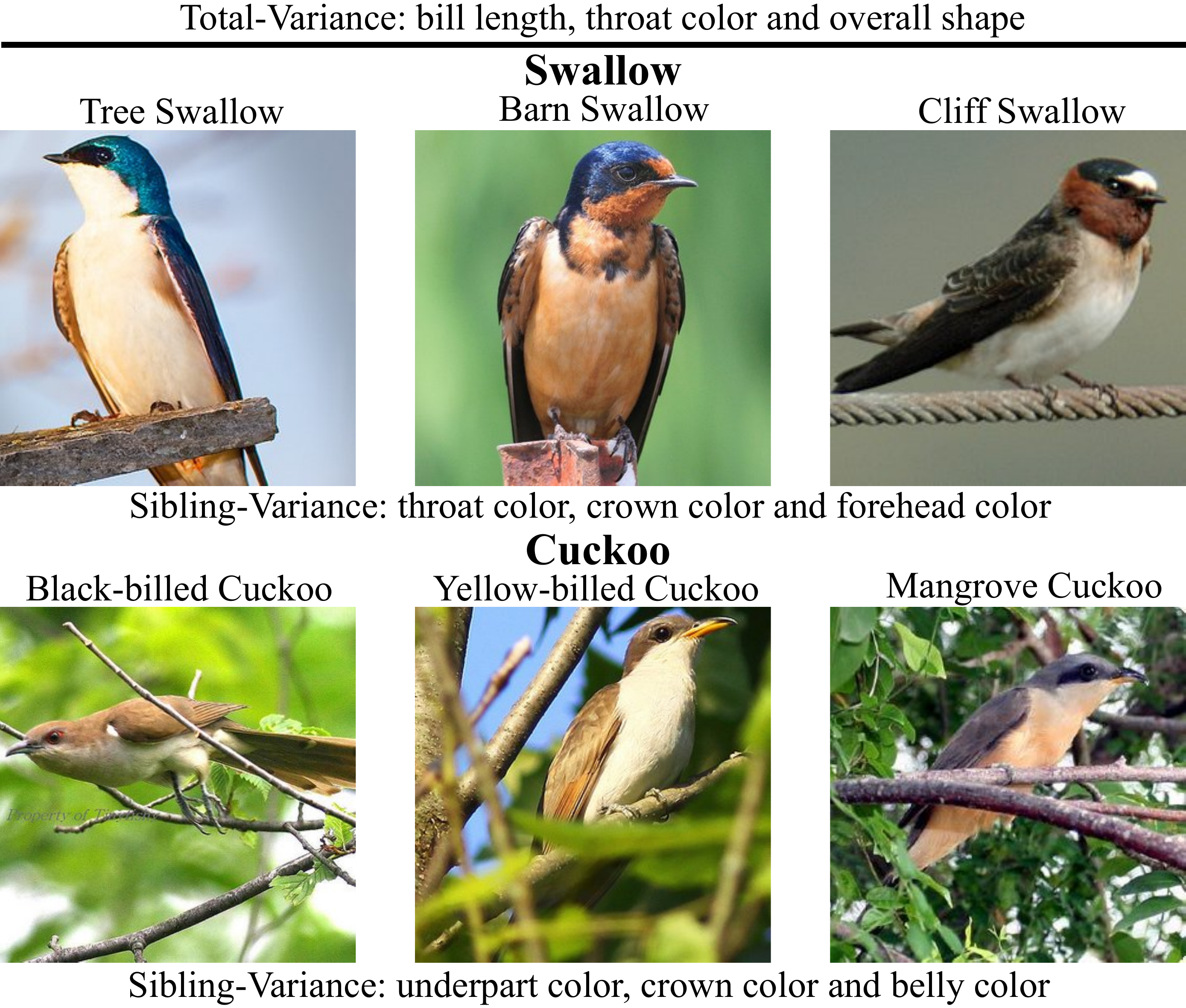}
\caption{Attributes selected by \emph{\sibvar} for a parent class in
the taxonomy and the attributes selected by measuring variance over all classes (top).
As can be seen from the images within the categories, overall shape
and bill length do not change much and measuring variance to pick
attributes across all classes is not informative and our approach of looking at the siblings is necessary. 
}
\label{fig:attribute}
\vspace{-1em} 
\end{figure}

\begin{figure}[h!]
\centering
\includegraphics[width=0.45\linewidth]{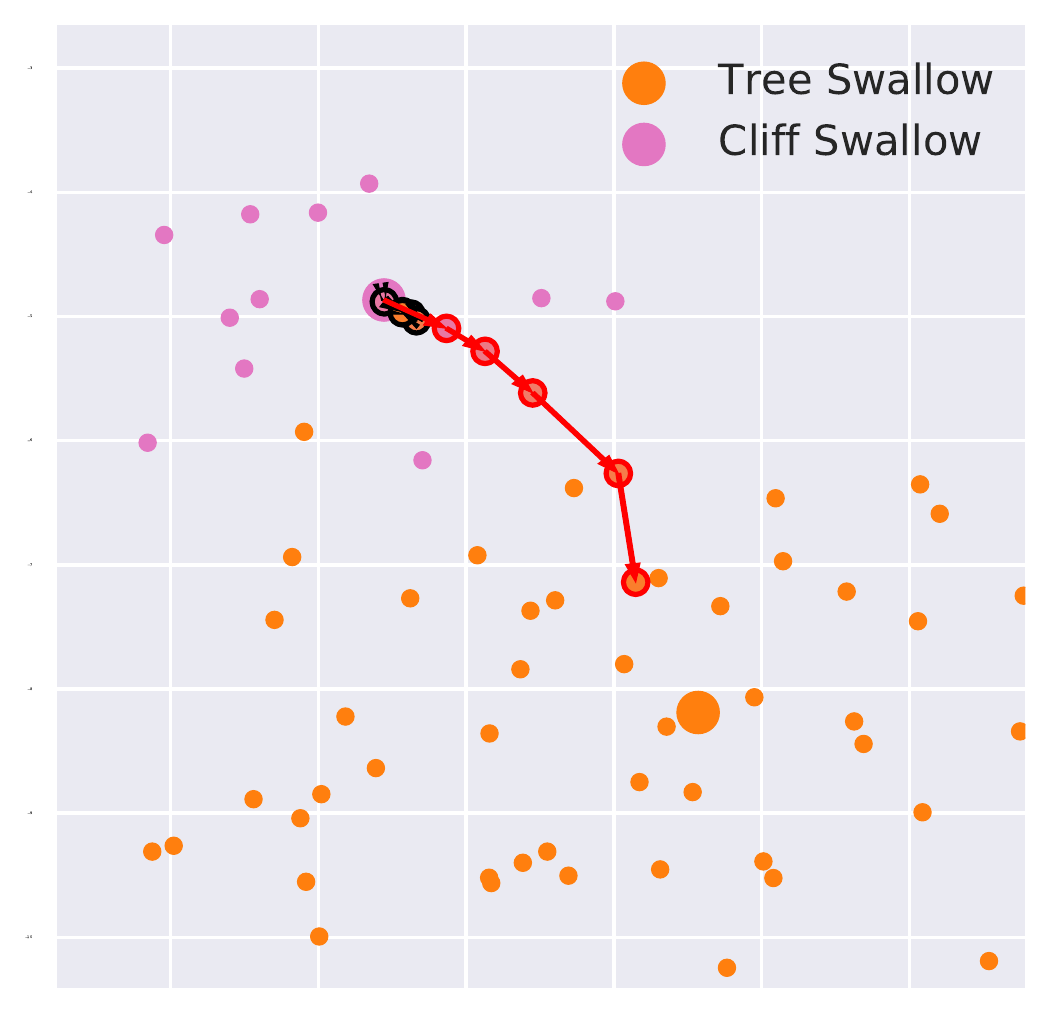}
\includegraphics[width=0.45\linewidth]{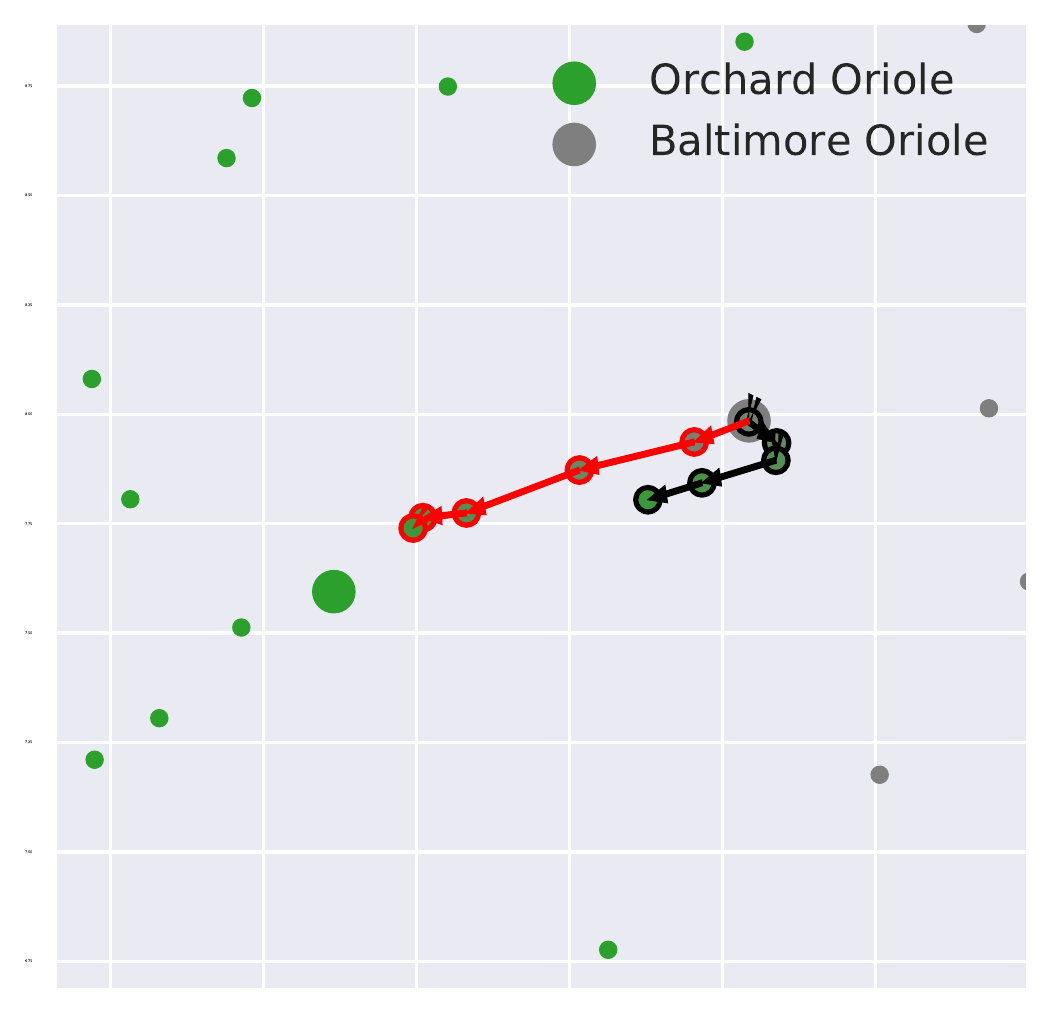}
\caption{t-SNE visualization for 2 novel CUB classes (first class in legend) and
its closest base classes (second) in the latent space of
CADA-VAE. Smaller dots represent test images and larger dots represent class attribute embedding.
Red edges show the progression of novel class attributes as learners interact using \emph{\sibvar}. Dots with black edges show the progression with \emph{random} function. }
\vspace{-1em}
\label{fig:tsne}
\end{figure}

We also visualize how learning progresses with interactive questions and reponses.
Figure~\ref{fig:tsne} shows the t-SNE visualization of 2 CUB classes and the most similar base classes. 
The full attributes descriptor (bigger dot) for a class is surrounded
by images of that class represented by smaller dots of the same color.
Dots with red and black edges show the progression of novel class
attributes as the learner interactively gains more information using
the \emph{\sibvar} and \emph{random} acquisition function
respectively. 
We see that \emph{\sibvar} yields a faster progression from the similar base class attribute to the novel class.
For example, the progression of the attribute descriptor encoding
using \emph{\sibvar} for ``Tree Swallow'' reaches the full attribute
descriptor quicker, whereas with the \emph{random} function the
attribute is still close to the base class ``Cliff Swallow''.
More examples are in the supplementary.

\subsection{How much does the taxonomy help?}
As seen from the results in previous sections, \emph{\sibvar} lets us
select informative attributes.  In this section we evaluate if
having a taxonomy is a requirement for this to work.
Rather than measuring variance within the sibling classes we measure
variance over all  classes. 
This variant will always prioritize some types of attributes over
others irrespective of the class,  and ignore local variations within
siblings. Therefore,
we expect it will perform worse than with known taxonomy information.

\begin{figure}[h!]
\centering
\includegraphics[width=0.45\linewidth]{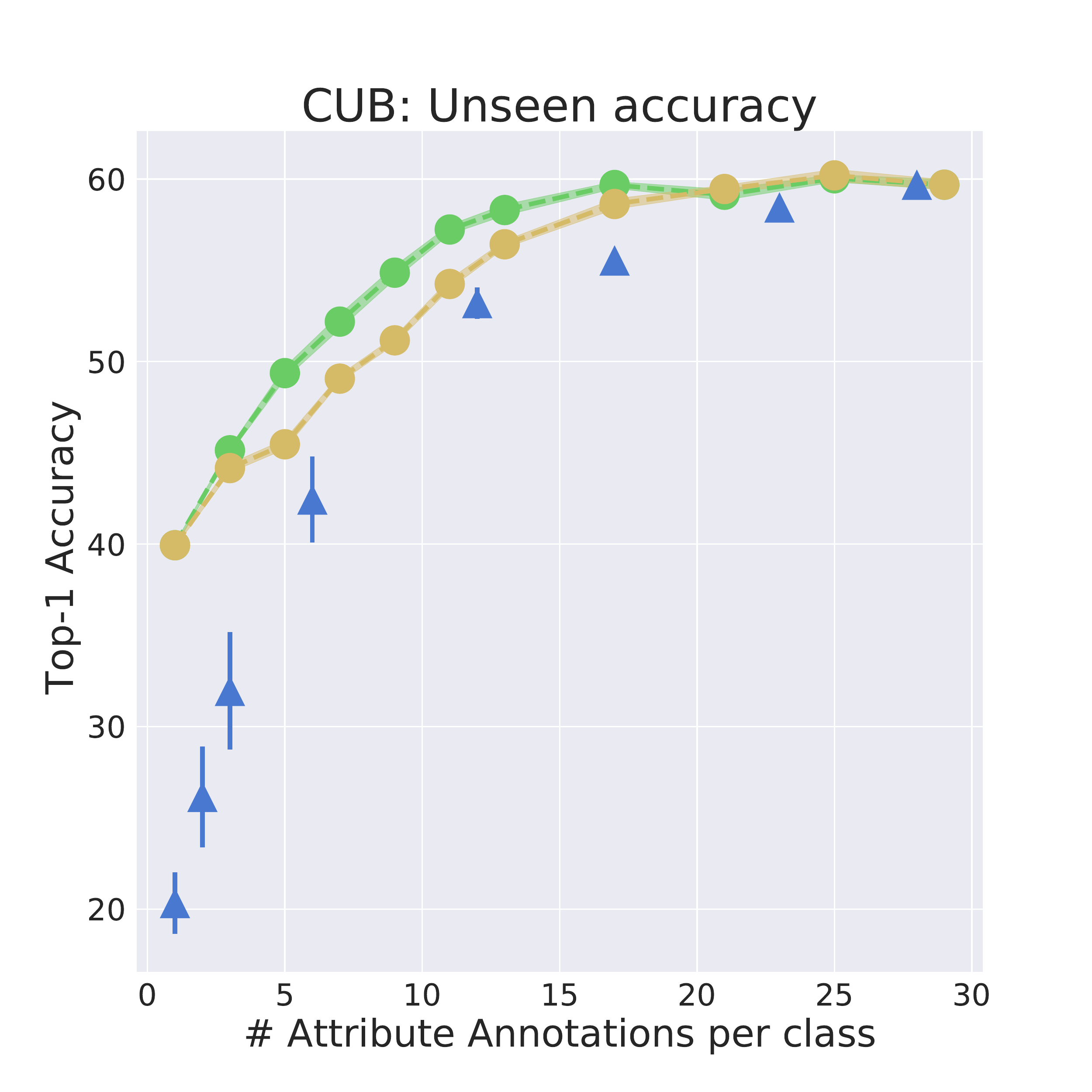}
\includegraphics[width=0.45\linewidth]{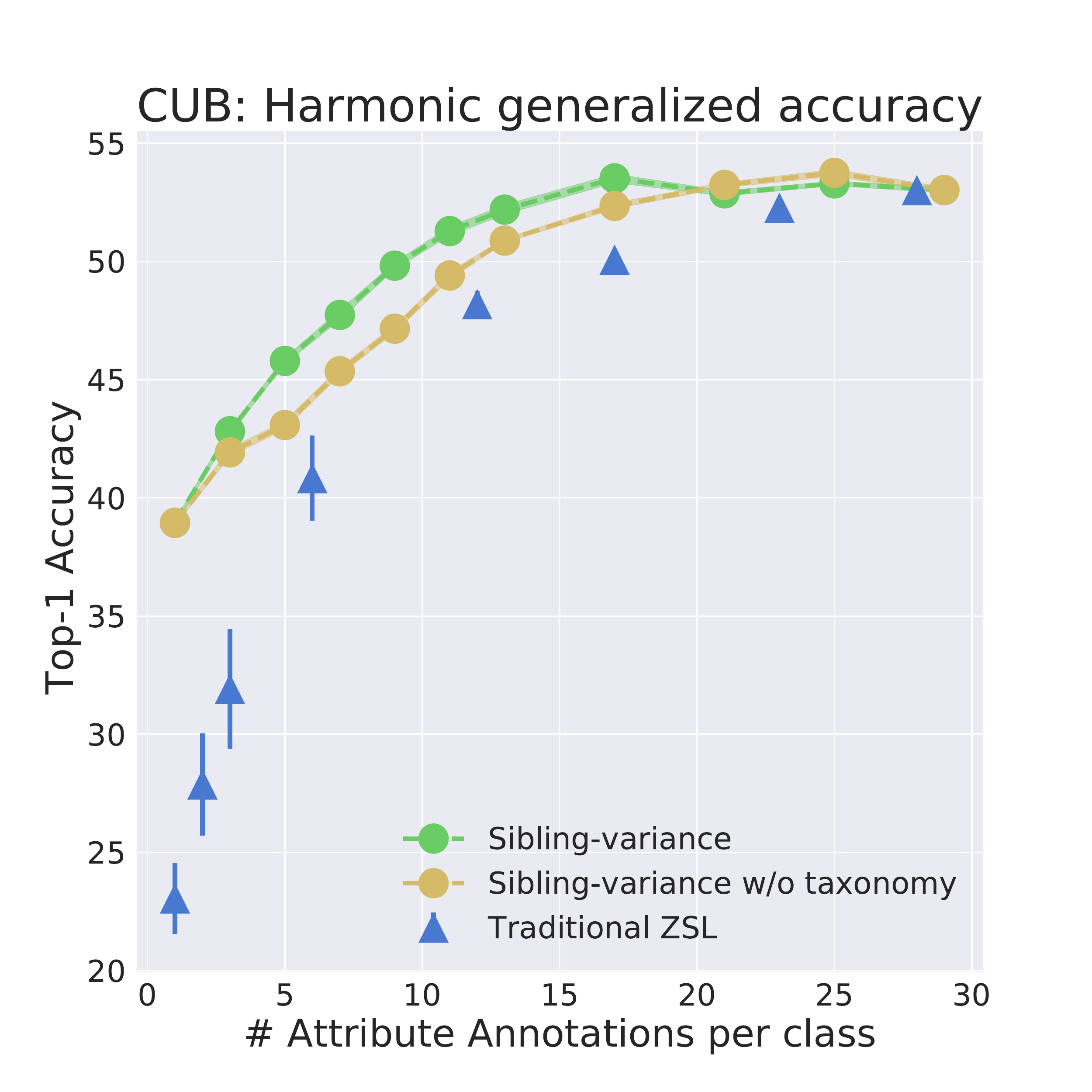}
\caption{Comparing \emph{\sibvar} against a variant where the taxonomy is unknown on CUB. 
The model loses accuracy if sibling classes are not used to measure \emph{\sibvar}.}
\label{fig:notaxo}
\vspace{-1em} 
\end{figure}

Figure~\ref{fig:notaxo} compares the method without taxonomy
information against the model where the taxonomy is known. 
As expected, this method loses performance because the local variation
of a class cannot be measured, and hence those attributes cannot be selected.
But even without the taxonomy the method performs better than \zsl and
is useful for cases when the taxonomy is not known or difficult to acquire
(See supplementary for performance on other datasets.).  Finally, we
also show that our methods are not very sensitive to similar base
class selection as long as the expert chooses a class that is not
wildly different looking (see supplementary).

\vspace{-0.5em}
\section{Conclusions}
In this work we show that an interactive {\em field-guide-inspired}
annotation approach identifies informative attribute queries, 
and can achieve high performance by judiciously
using an annotation budget.  We present different ways to
acquire informative sparse annotations from an annotator and show that
is is better to let the machine choose the attributes to ask for
(even when compared to an expert from the domain). Given
these promising results, there are many avenues of future work: can
one get even more cost-efficient by choosing a different number of
attributes for different classes based on class confidence? 
We also need to bridge the gap between the attribute
understanding of humans and machines, as our results show that human
experts and neural-networks do not find the same attributes equally
useful.

\vspace{0.5em}

\noindent\textbf{Acknowledgements.} We thank our funding agencies TCS, NSF 1900783, and DARPA LwLL program (HR001118S0044) for their support.

\newpage
{\small
\bibliographystyle{ieee_fullname}
\bibliography{actzsl-bibliography}
}

\renewcommand\thefigure{A\arabic{figure}}

\newpage
\appendix
\renewcommand\appendixpagename{Supplementary Material}

\appendixpage
\addappheadtotoc
\counterwithin{figure}{section}

\section{Overview}
In this supplementary material we look at some more results that could not be presented in the main paper.
Our code can be found at \href{https://github.com/utkarshmall13/Field-Guide-ZSL}{github.com/utkarshmall13/Field-Guide-ZSL}
We show attributes queried by \emph{\sibvar} on SUN and AWA2 in Sec.~\ref{sec:attvis}.
Sec.~\ref{sec:acqu} shows the performance of the two acquisition
functions on AWA2 and SUN.
In Sec.~\ref{sec:oneshot} we present results for when the interactive
learner uses a single image for SUN and AWA2.  
In Sec.~\ref{sec:tfvg} we compare the performance of our method with
TF-VAEGAN on the AWA2 and SUN. 
Sec.~\ref{sec:notaxo} shows the effect of not using a taxonomy in \emph{\sibvar} for AWA2 and SUN.
In Sec.~\ref{sec:similar} we show the learner's behavior when classes other than the annotators first choice are chosen for SUN and AWA2.
Sec.~\ref{sec:tsne} presents more t-SNE visualization examples of the
learning progression for novel class descriptors on all three
datasets.  We also {\bf strongly} encourage the reader to refer to the
\textbf{\href{https://www.cs.cornell.edu/projects/field-guide/static/videos/supp-video.mp4}{supplementary video}} for better visualizations of the t-SNE
progression.

\section{More Qualitative Evaluation}\label{sec:attvis}
\begin{figure}[h!]
\centering
\includegraphics[width=0.8\linewidth]{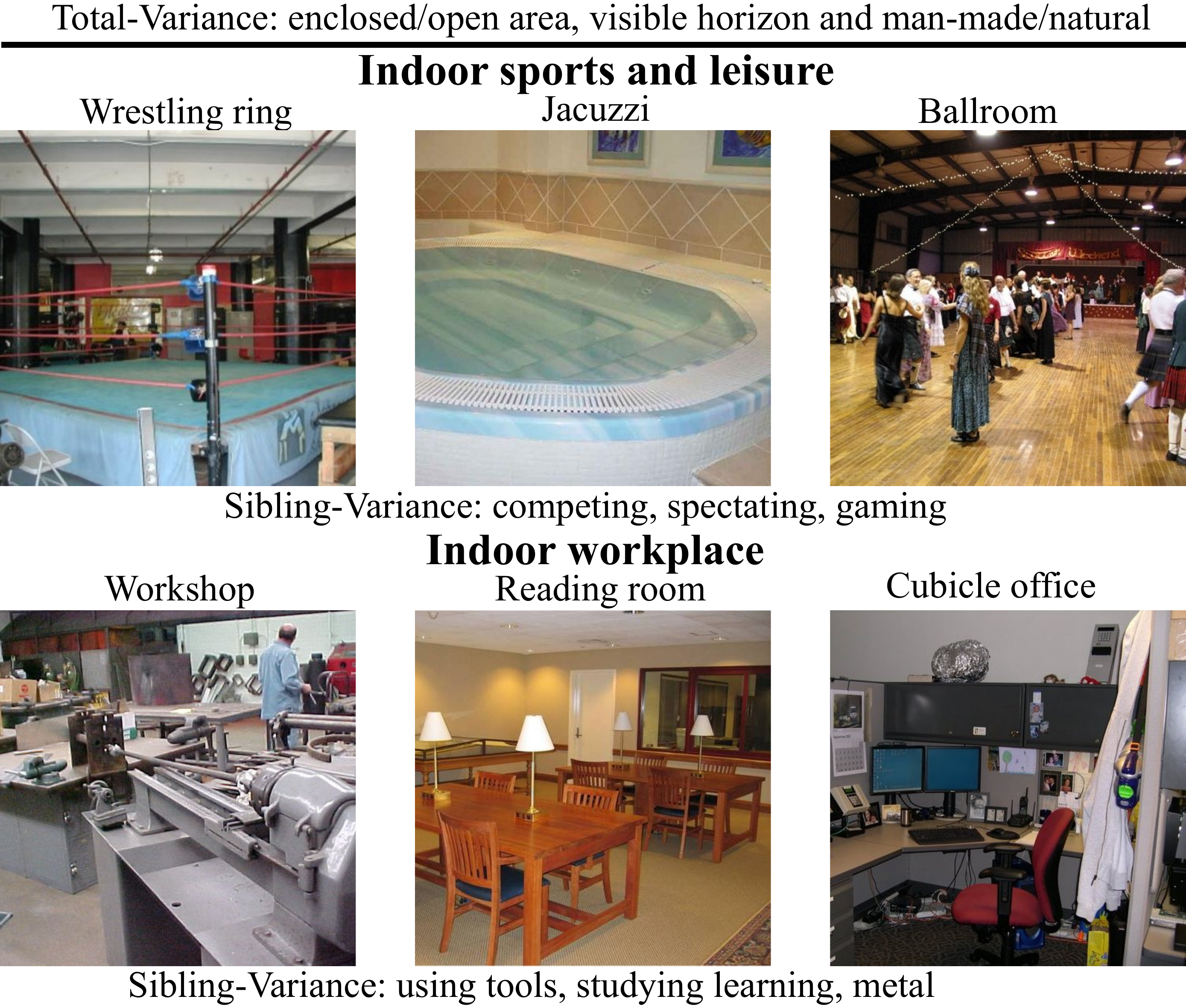}
\includegraphics[width=0.8\linewidth]{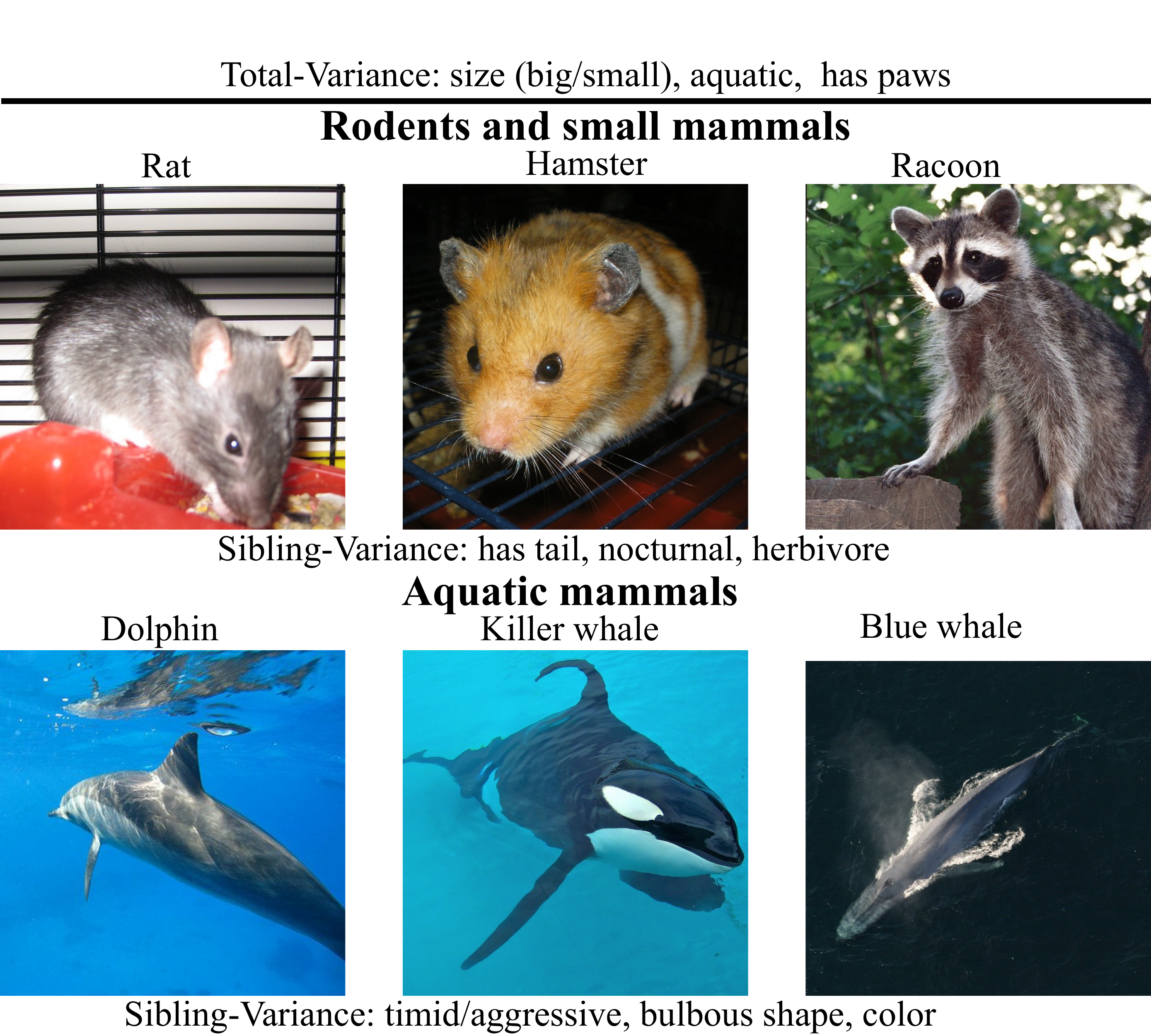}
\caption{Attributes selected by \emph{\sibvar} for a parent class in
the taxonomy and the attributes selected by measuring variance over all classes (top) for SUN and AWA2.
}
\label{fig:attribute}
\end{figure}

Figure~\ref{fig:attribute}, shows the attributes queried first by
the \emph{\sibvar} method for 2 supercategories of SUN and AWA2.  
It also shows the attributes picked by measuring variance over all the classes.
For SUN, attributes like ``enclosed/open area'' or ``man-made/natural'' may help in disambiguating between very different classes, 
but within a supercategory they do not help.
For example, for the superclass ``Indoor sports and leisure'', all the classes are closed and man-made.
But attributes like ``competing'', ``spectating'' are more informative.
Similarly for indoor workplaces, attributes like ``using tools'' and
``studying/learning'' are very informative. Similar
patterns can be seen on AWA2. \emph{\sibvar} asks for attributes
informative within the superclass.

\section{Comparison of Acquisition Functions on AWA2 and CUB}\label{sec:acqu}
Figure~\ref{fig:acquisition} shows the performance of the two
attribute querying acquisition functions with the CADA-VAE model on AWA2 and SUN. 
Our acquisition functions perform significantly better than a random acquisition function, showing the value of our field-guide annotation. 

While the results for fine-grained dataset such as SUN are similar to that on CUB, 
for AWA2 \emph{\encchange} performs better than \emph{\sibvar} in the later stages.
This might be because the AWA2 model is trained for fewer coarse-grained classes with thousands of images and has a better representation and understanding of changing representation.

\begin{figure}[h!]
\centering
\includegraphics[width=0.45\linewidth]{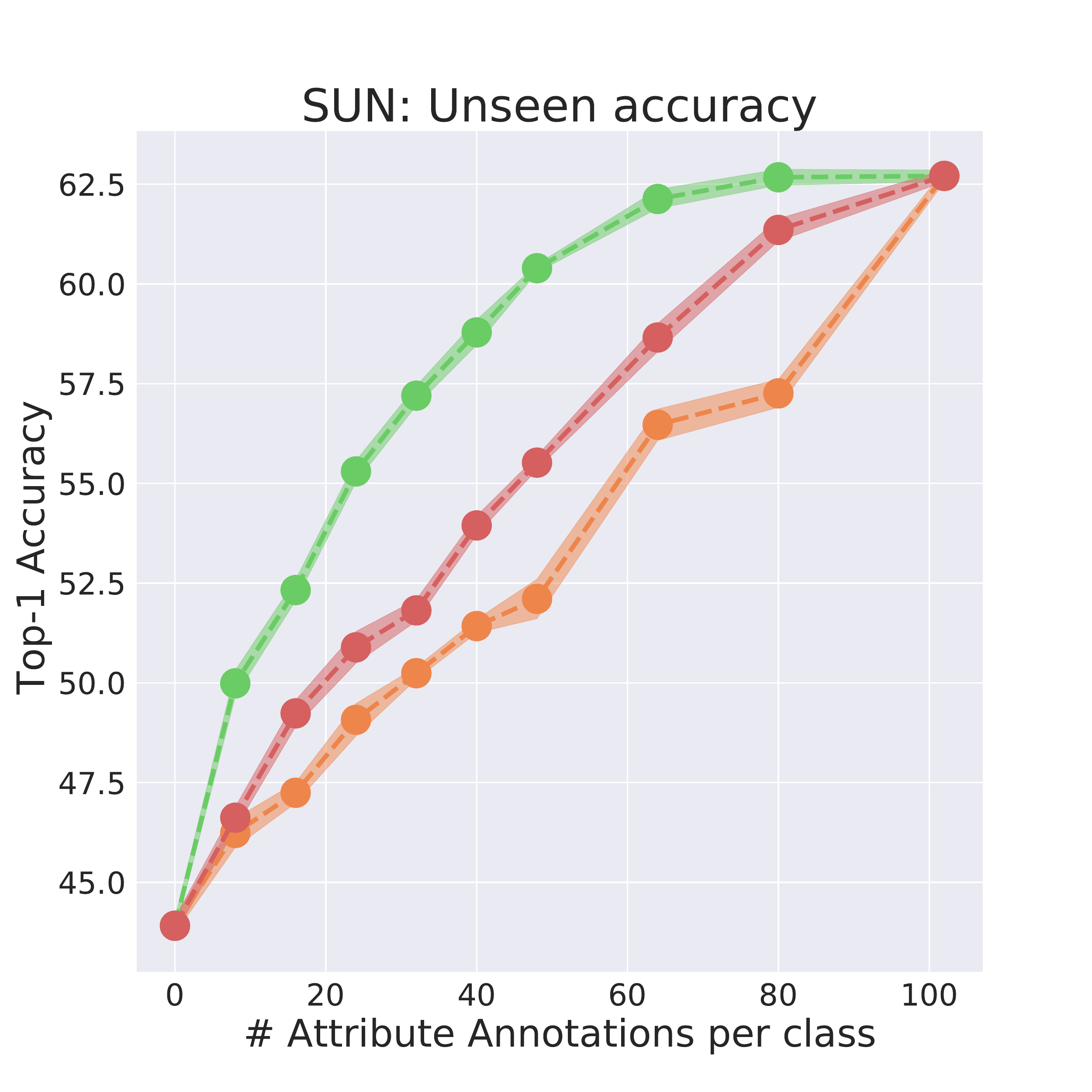}
\includegraphics[width=0.45\linewidth]{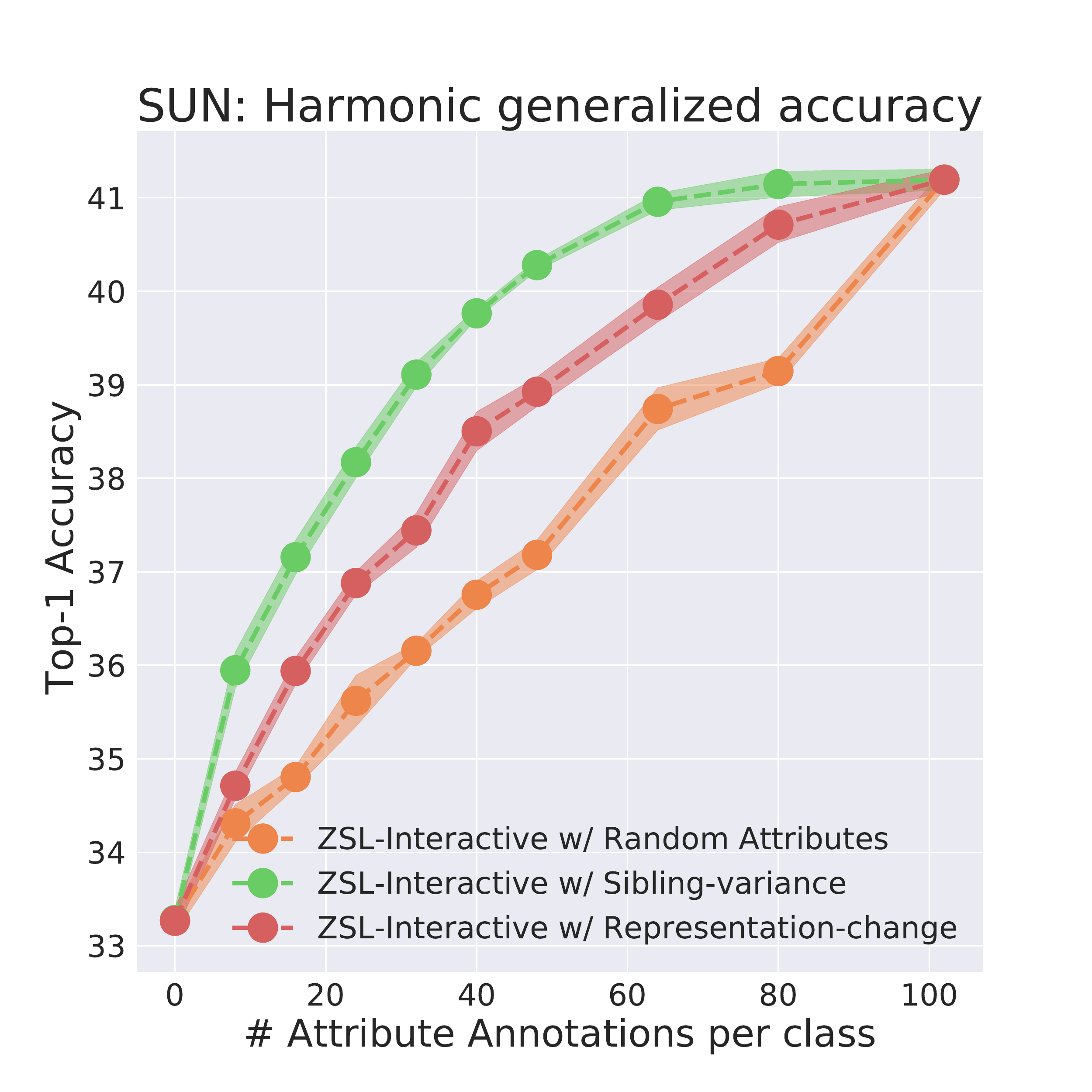}
\includegraphics[width=0.45\linewidth]{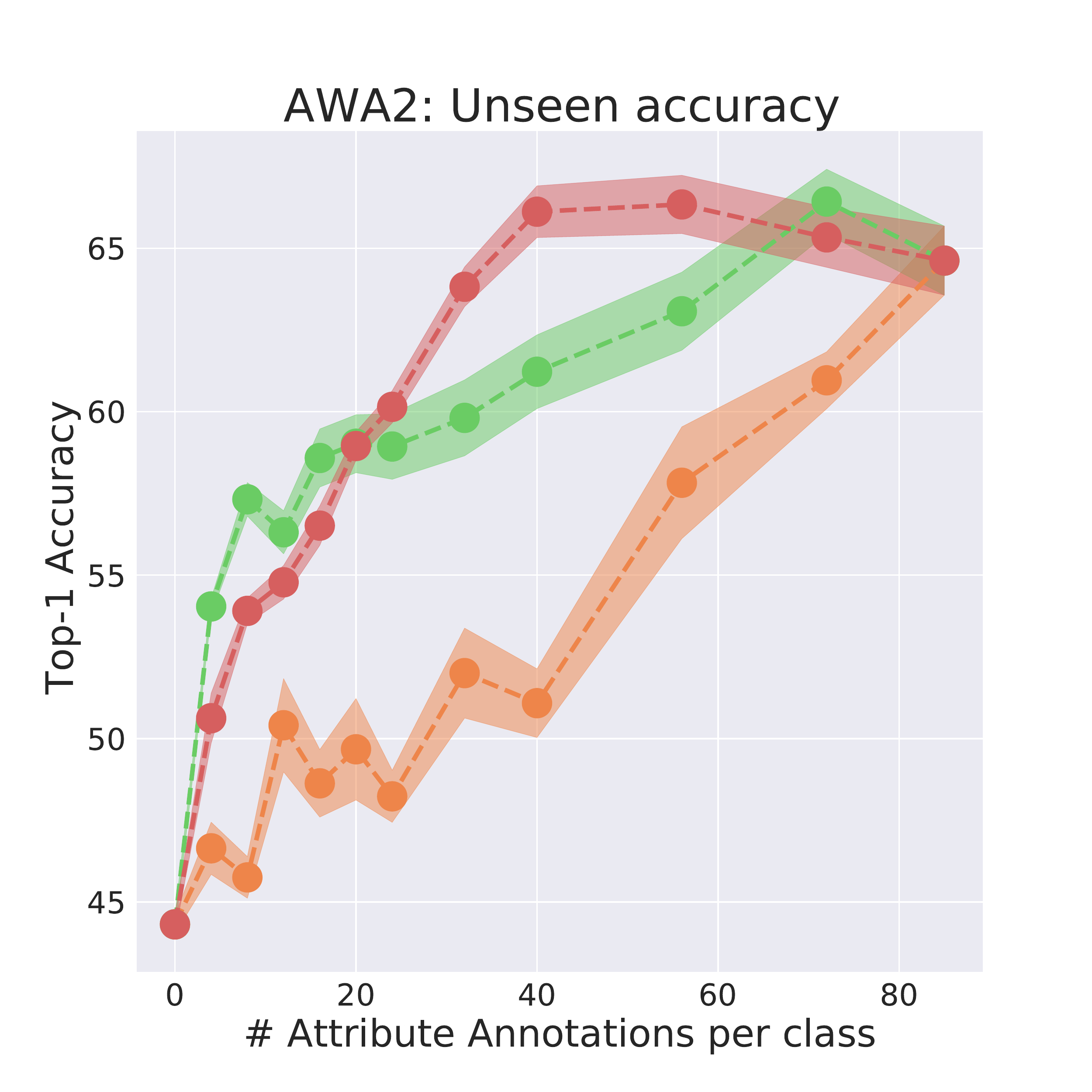}
\includegraphics[width=0.45\linewidth]{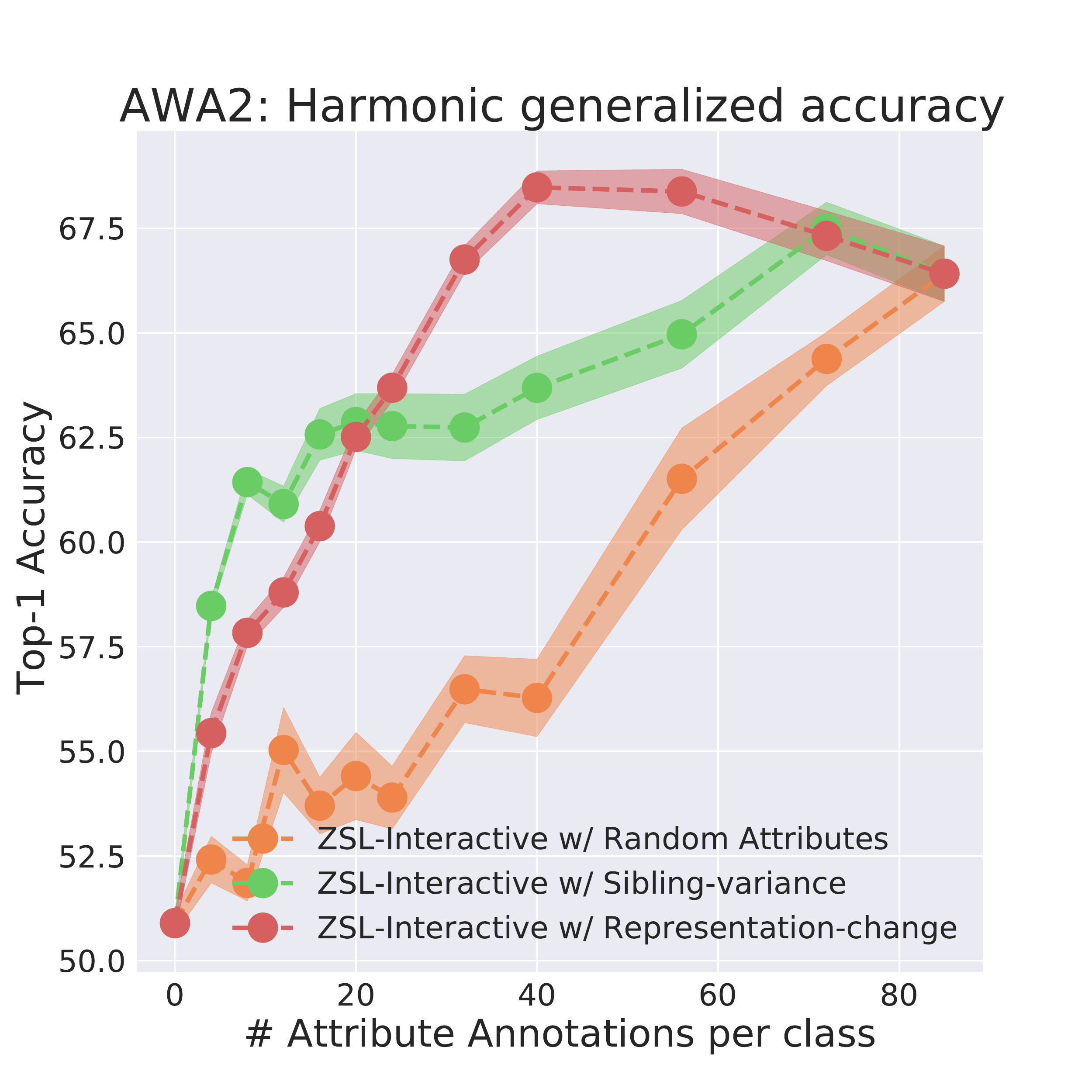}
	\caption{Performance of the two acquisition functions with
CADA-VAE on AWA2 and SUN. Both functions perform better than the random
acquisition function. On SUN, {\emph\sibvar} performs better than
\emph{\encchange}, but the latter does not require taxonomy information.
Results on AWA2 are different, in the earlier stages \emph{\sibvar} is
better than \emph{\encchange}, but in the later stages
\emph{\encchange} is better.
}
\label{fig:acquisition} 
\end{figure}

\section{\emph{\imagebased} Results for AWA2 and SUN}\label{sec:oneshot}
Figure~\ref{fig:oneshot} shows the performance of our approach when one image is given by the annotator along with the interactive attribute annotations for AWA2 and SUN.
As for CUB, all the methods perform better than the baselines. 
For SUN, the \emph{\imagebased} function performs on par
with \emph{\sibvar} without requiring an additional taxonomy. 
For AWA2, the \emph{\imagebased} function performs better than all the
methods we propose and the baselines. 
AWA2 has more training images and the classes are not very fine-grained.
This might be the reason why \emph{\imagebased} acquisition
functions work better for AWA2.

\begin{figure}[h!]
\centering
\includegraphics[width=0.45\linewidth]{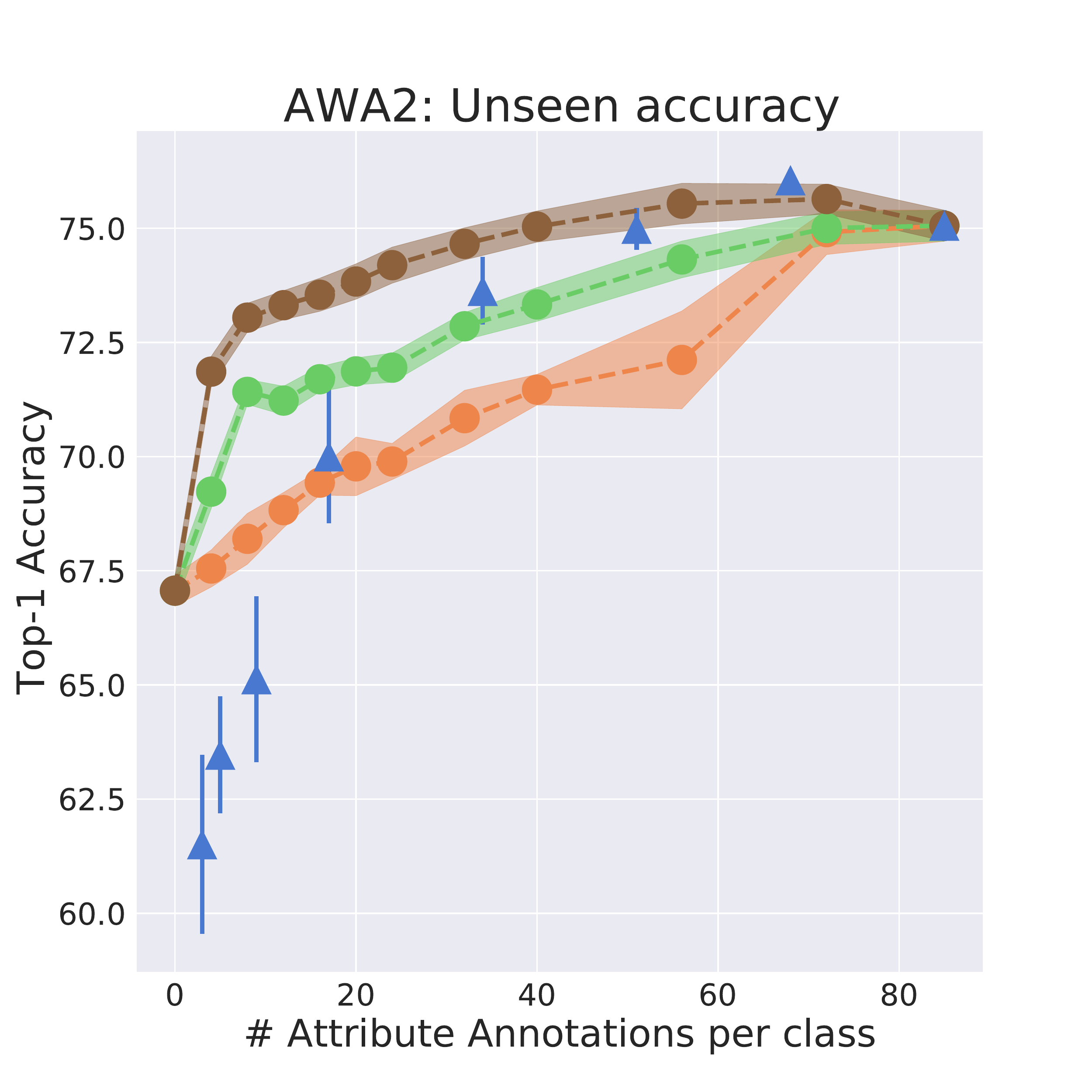}
\includegraphics[width=0.45\linewidth]{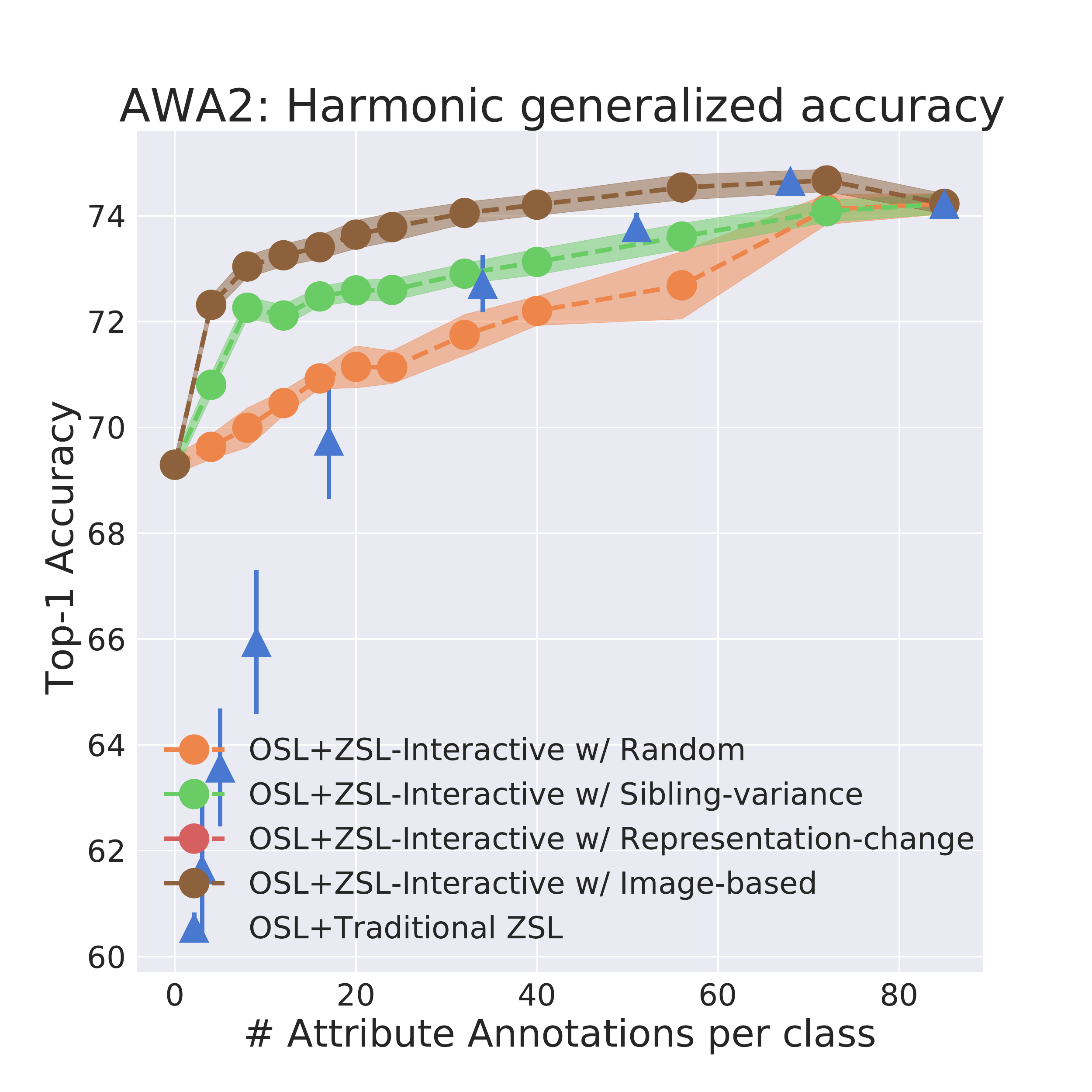}
\includegraphics[width=0.45\linewidth]{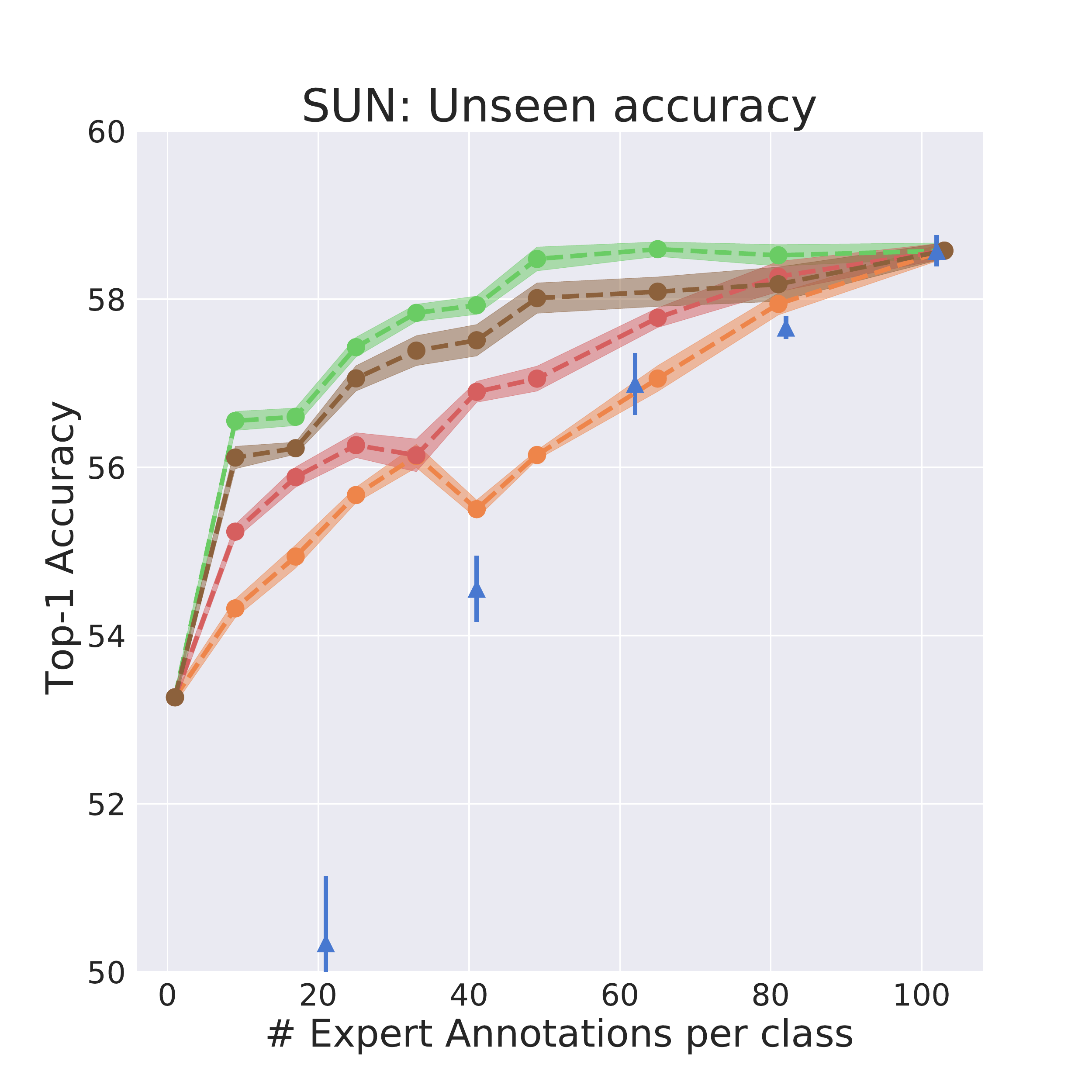}
\includegraphics[width=0.45\linewidth]{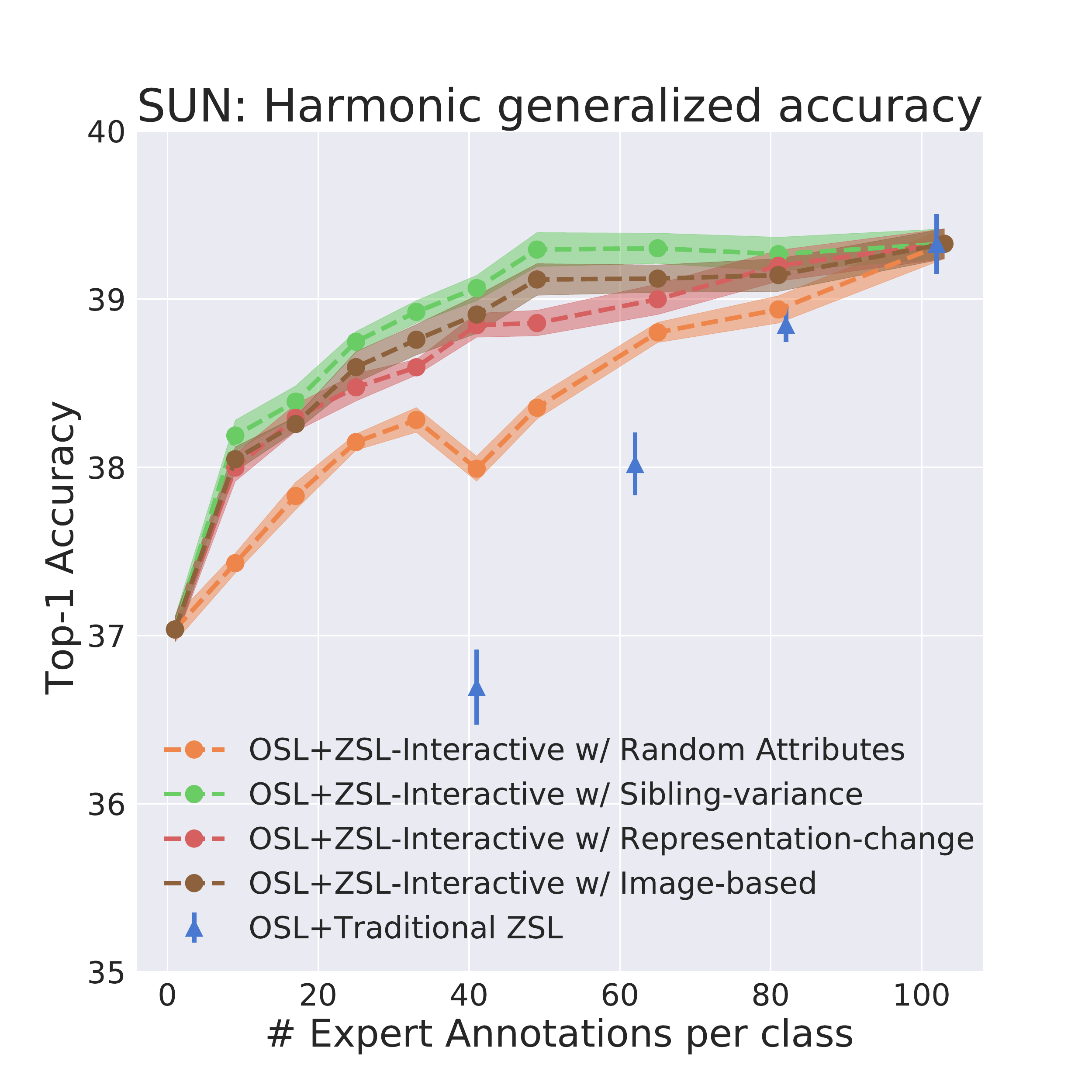}
\caption{Performance of our method under the zero+one-shot
setting when the annotator provides a single image for novel class
along with the interactive attribute values for SUN and AWA2. 
}
\label{fig:oneshot}
\end{figure}

\section{Performance of TF-VAEGAN on AWA2 and SUN}\label{sec:tfvg}
\begin{figure}[h!]
\centering
\includegraphics[width=0.45\linewidth]{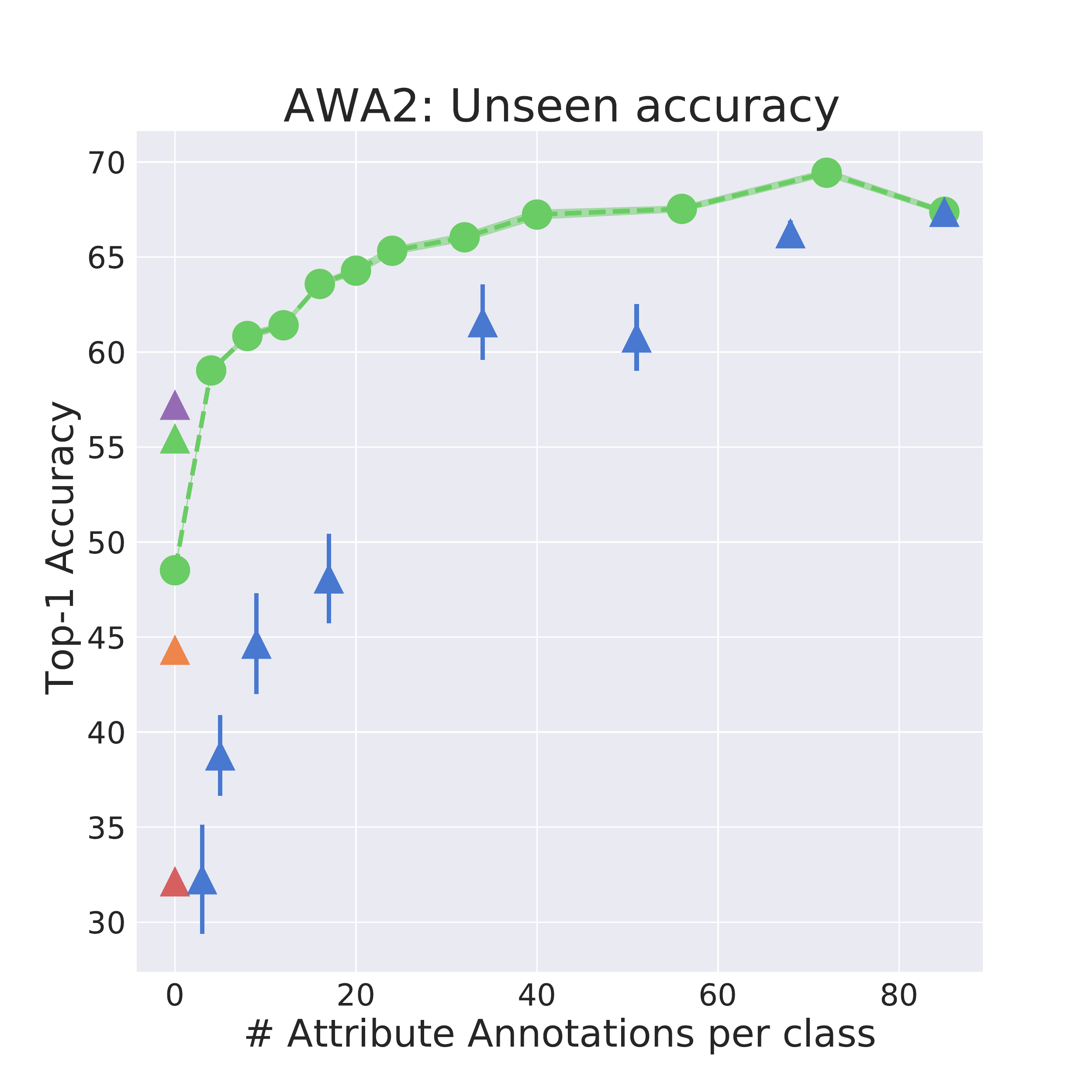}
\includegraphics[width=0.45\linewidth]{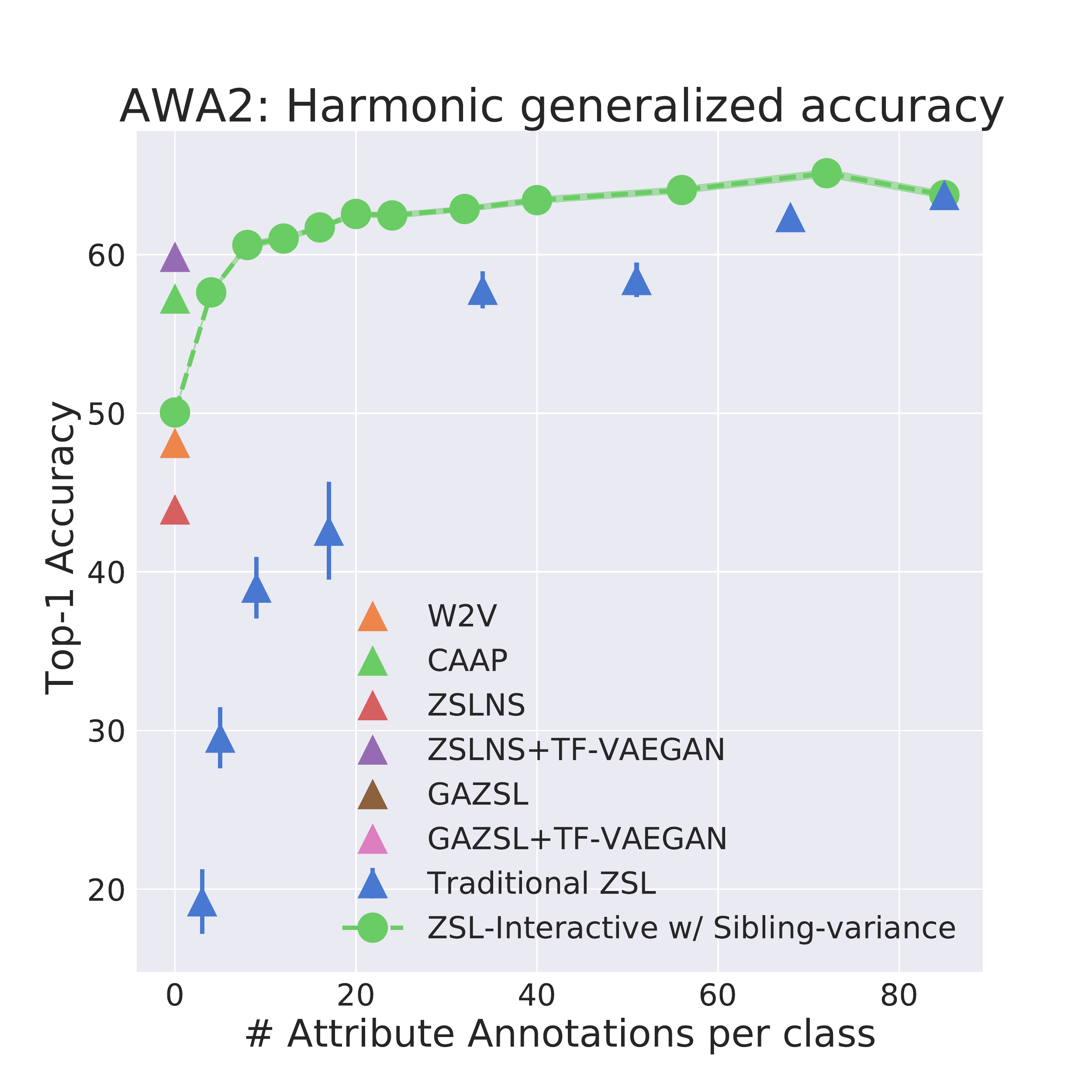}
\includegraphics[width=0.45\linewidth]{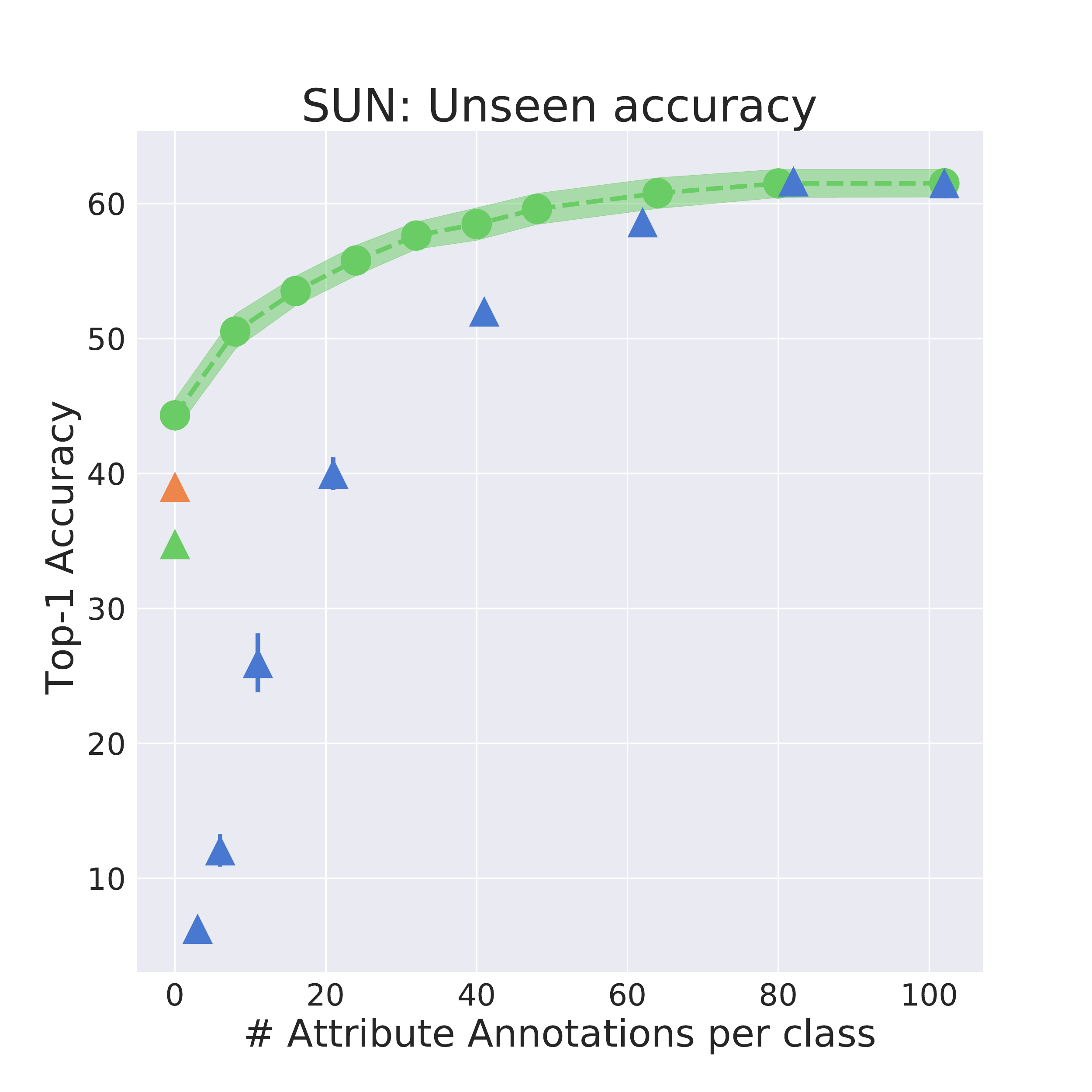}
\includegraphics[width=0.45\linewidth]{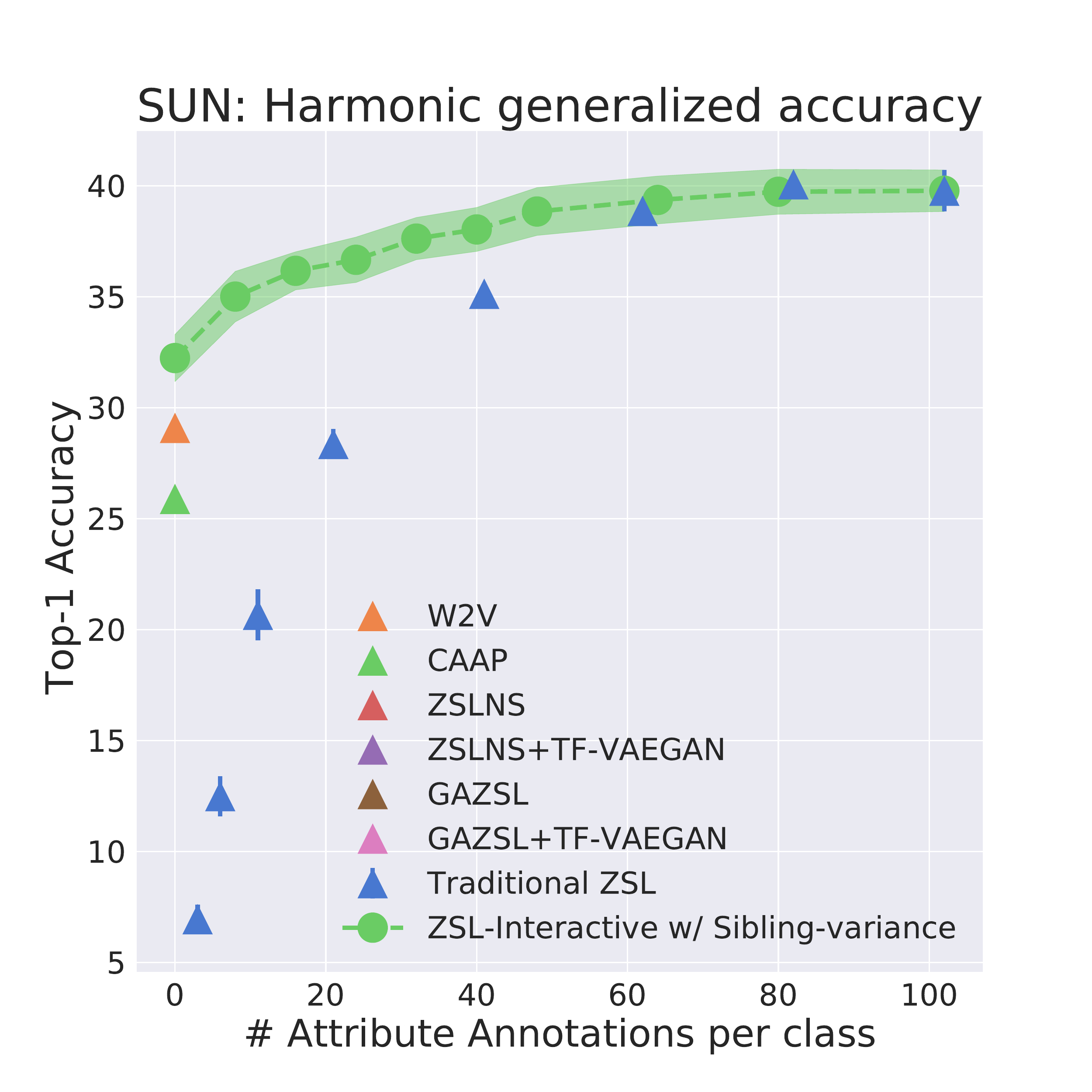}
\caption{Comparison of our method against unsupervised and traditional ZSL baselines with TF-VAEGAN as the base model.
Our method performs better than tradtional ZSL at the same attribute annotation cost for both AWA2 and SUN.
Similar to results for CADA-VAE in the main paper, our method works
better than the unsupervised baselines for SUN.}
\label{fig:tfvaegan}
\end{figure}

Figure~\ref{fig:tfvaegan} show the performance of our approach when the base model is TF-VAEGAN on SUN and AWA2. 
Our field-guide way of annotation works better then traditional ZSL baselines for both the dataset, proving the effectiveness and generalization of our method.
For fine-grained classes such as SUN and CUB, our method is .

\section{Performance on AWA2 and SUN Without Taxonomy}\label{sec:notaxo}
Figure~\ref{fig:notaxo} compares the method without taxonomy
information against the model where the taxonomy is known. 
The results follow the conclusion from the main paper. 
When there is no taxonomy information available, this method loses performance because the local variation of a class cannot be measured, and hence those attributes cannot be selected.
But even without the taxonomy the method performs better than \zsl and
is useful for cases when the taxonomy is not known or difficult to
acquire.

\begin{figure}[h!]
\centering
\includegraphics[width=0.45\linewidth]{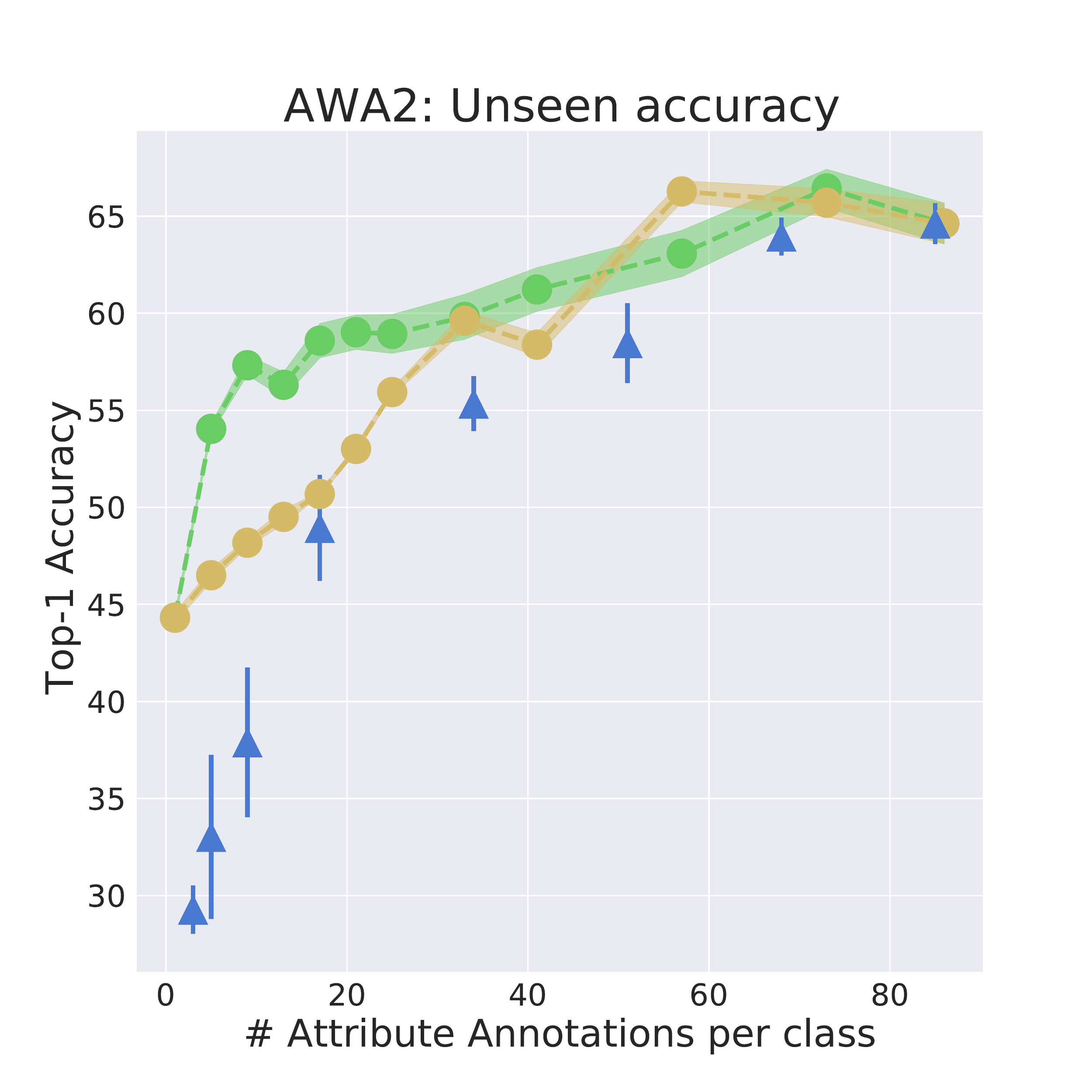}
\includegraphics[width=0.45\linewidth]{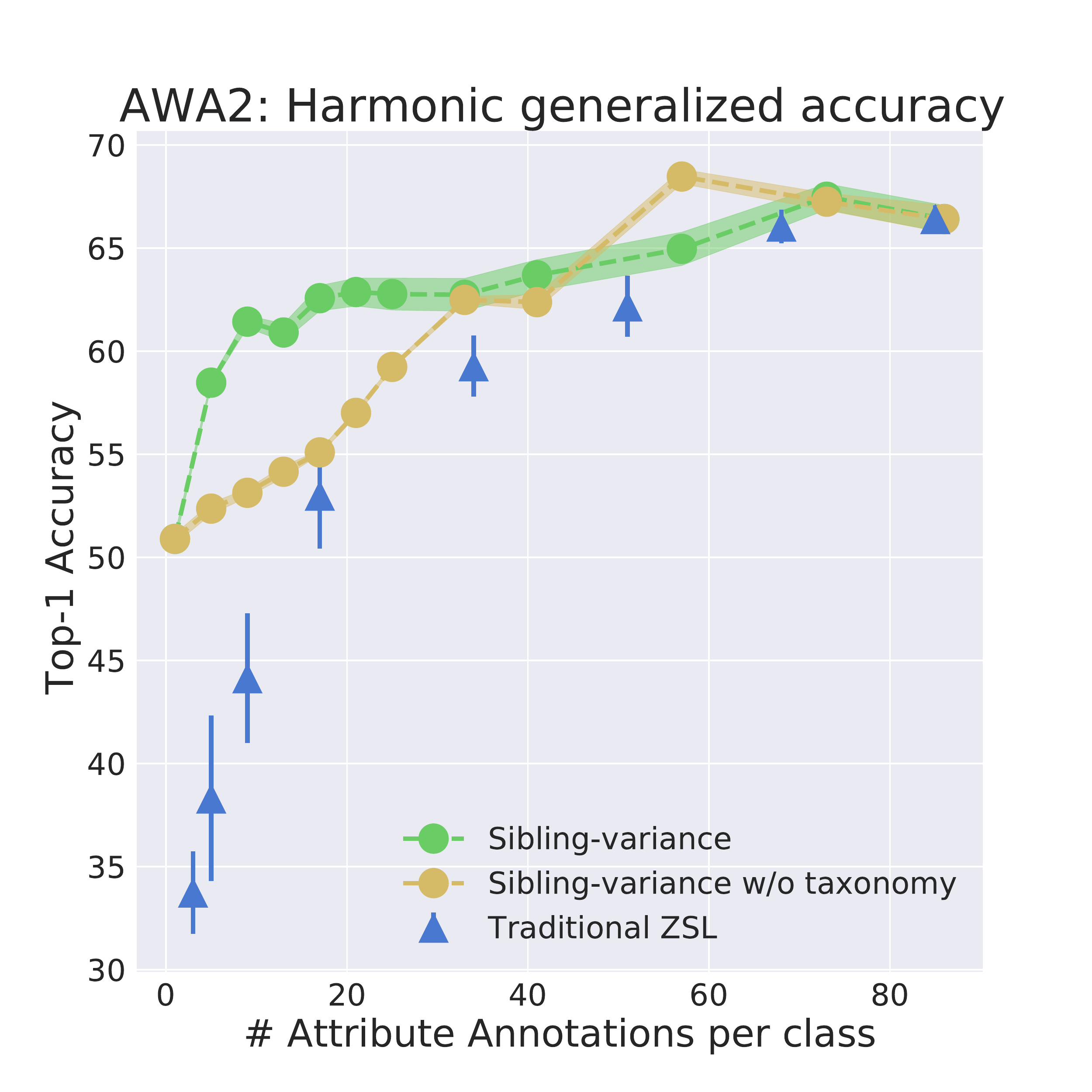}
\includegraphics[width=0.45\linewidth]{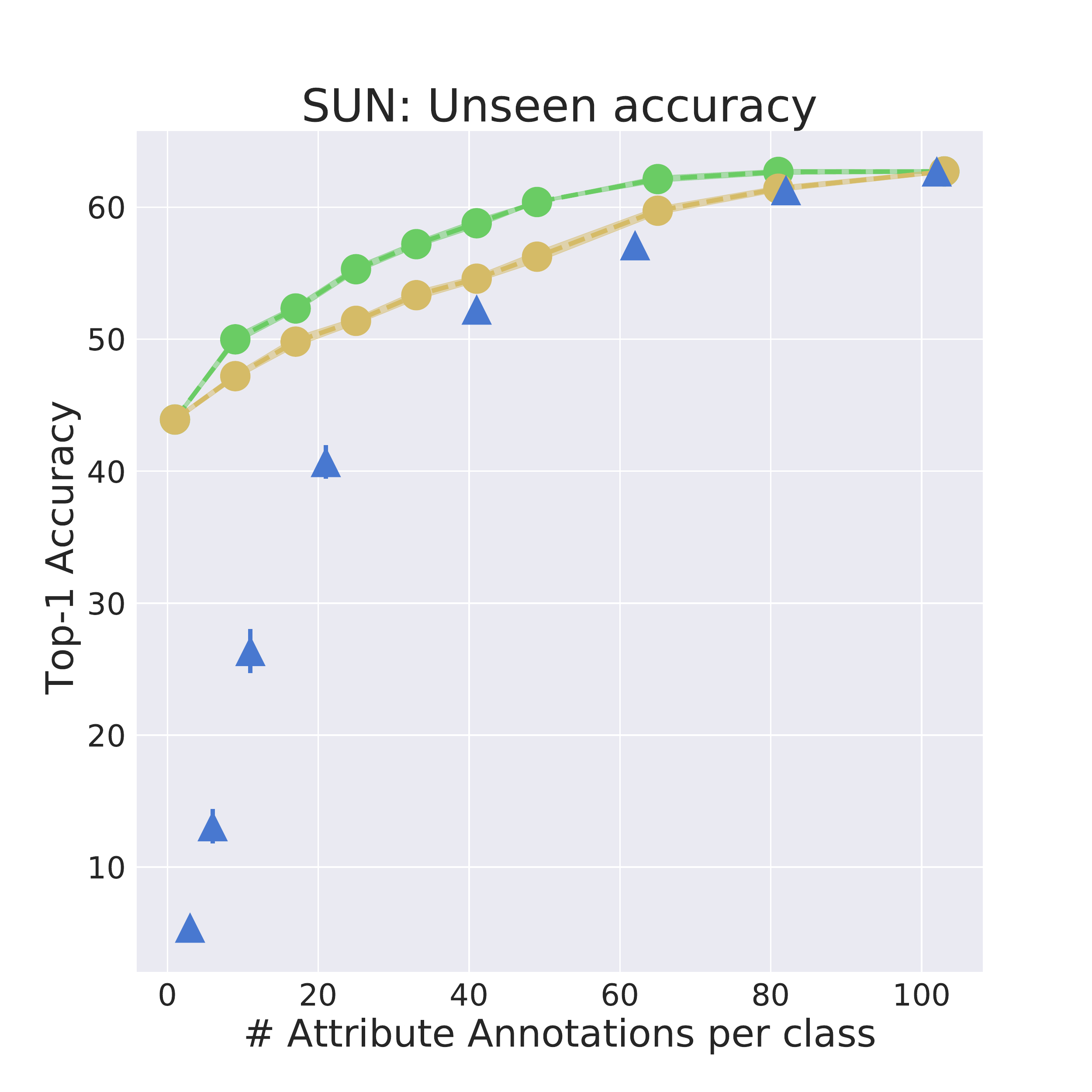}
\includegraphics[width=0.45\linewidth]{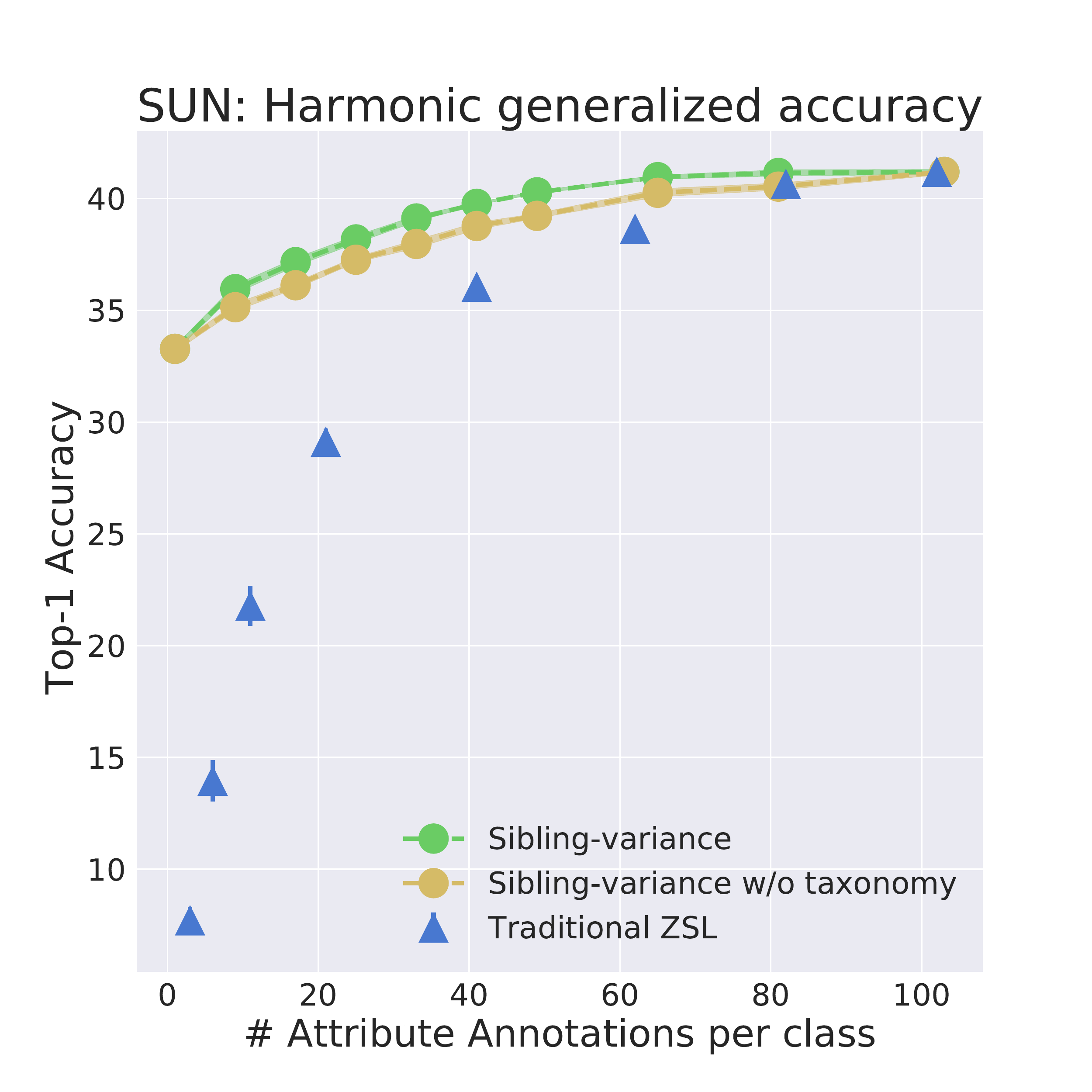}
\caption{Comparing \emph{\sibvar} against a variant where the taxonomy is unknown on AWA2 and SUN. 
The model loses accuracy if sibling classes are not used to measure \emph{\sibvar}.}
\label{fig:notaxo}
\end{figure}

\section{Effect on Performance When Changing Similar Base Class}\label{sec:similar}
The similar class given by the annotator is certainly more
important than each attribute annotation.  We look at
the effect of choosing another class: either a random class from the full set, or a sibling of the expert selection that is closest to the expert selected sibling in word2vec embedding space.

\begin{figure}[h!]
\centering
\includegraphics[width=0.45\linewidth]{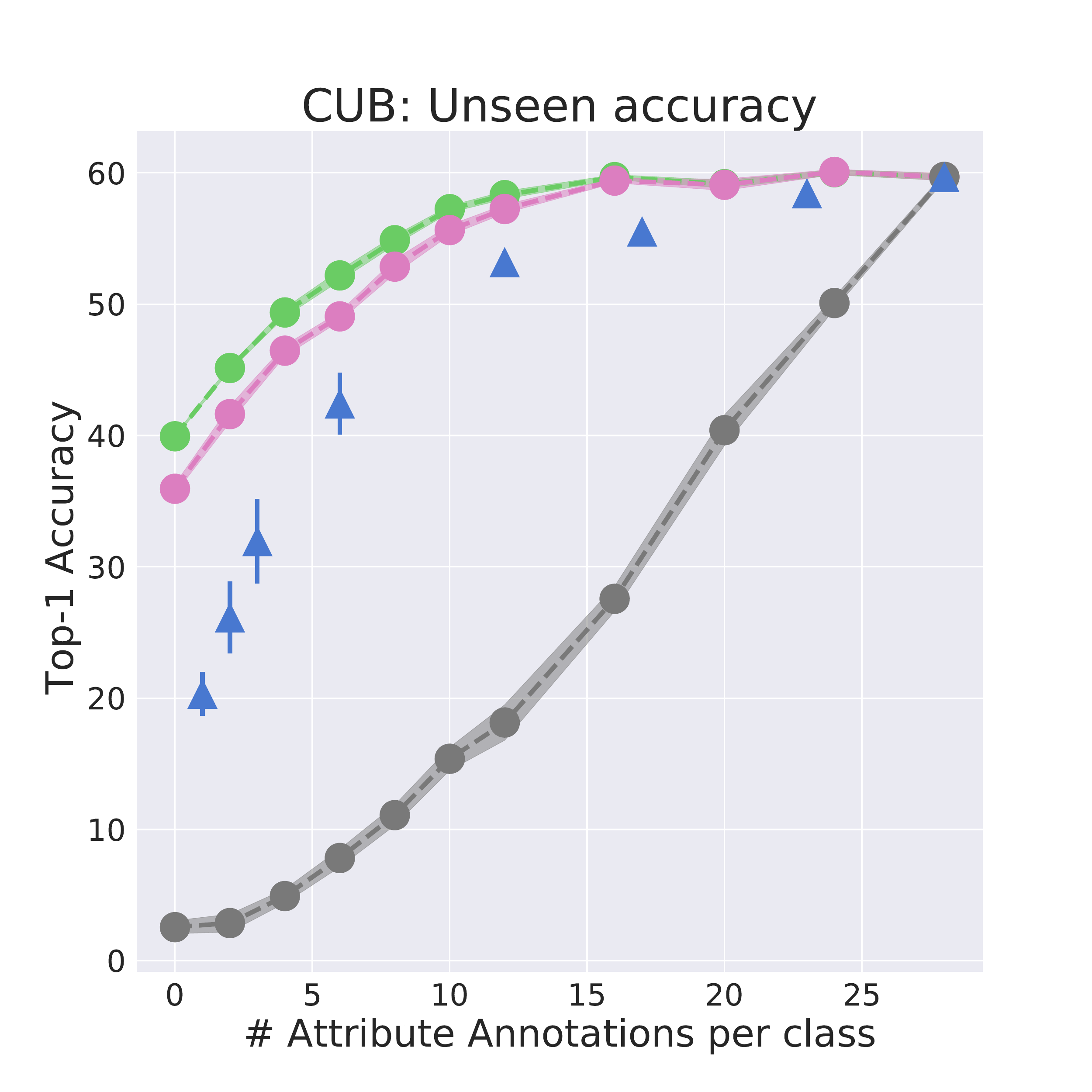}
\includegraphics[width=0.45\linewidth]{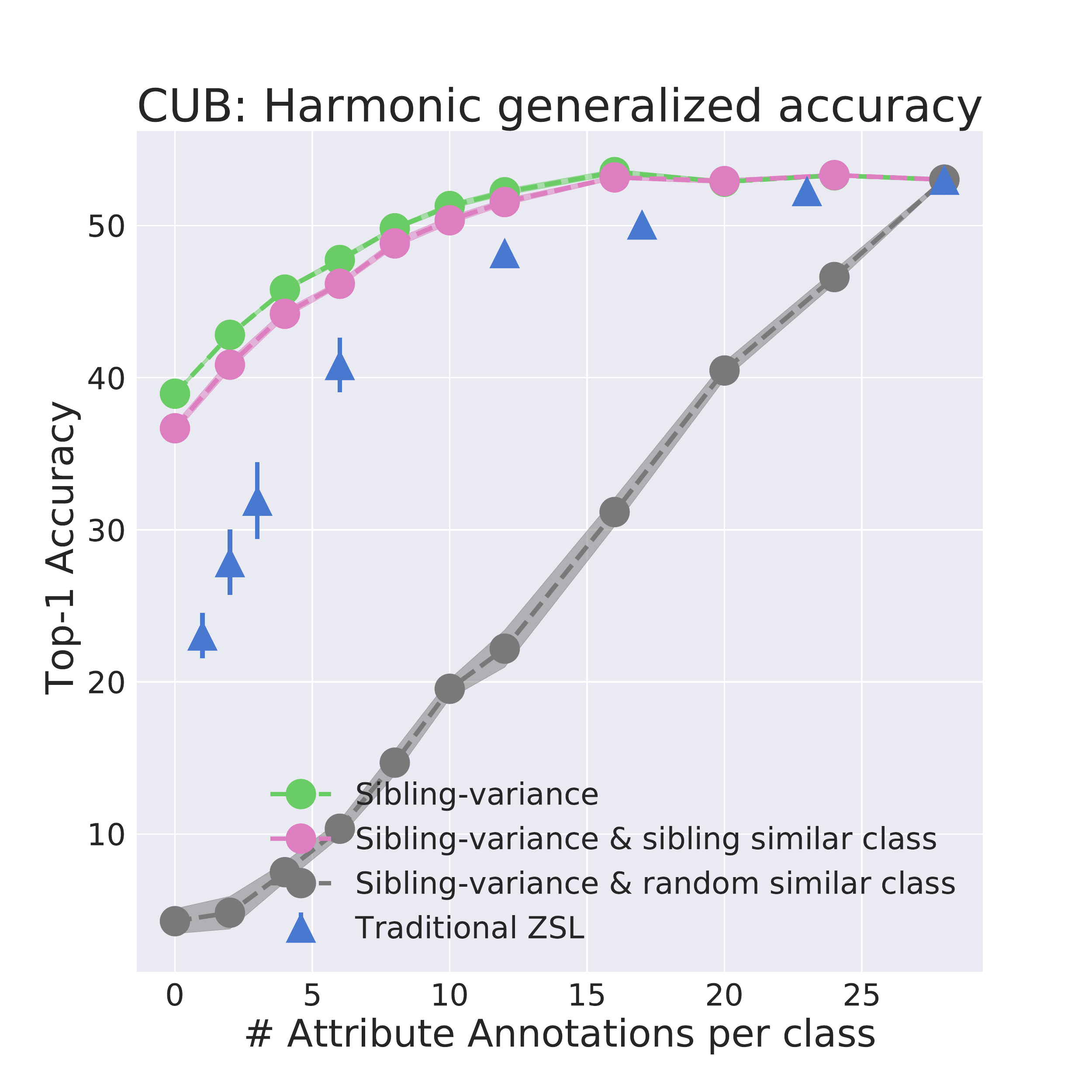}
\caption{Comparing different variants for selecting the similar class
$S(y)$ on CUB dataset. The model is not sensitive to the similar class as long as the selected class is not wildly different.}
\label{fig:similar1}
\end{figure}

Figure~\ref{fig:similar1} shows \emph{\sibvar} with these annotations along with the ZSL baseline on CUB.
The similar class chosen by annotators performs best.
When we choose a class that is close to this (sibling), the method performs slightly worse. 
This shows that although our method performance is affected if a
non-optimal nearest class is chosen, it is not very sensitive to it.
Both these variants do significantly better than randomly selecting a similar class. 
This suggests that the interactive model will perform well as long as
annotators do not provide a wildly different looking similar class.

\begin{figure}[h!]
\centering
\includegraphics[width=0.49\linewidth]{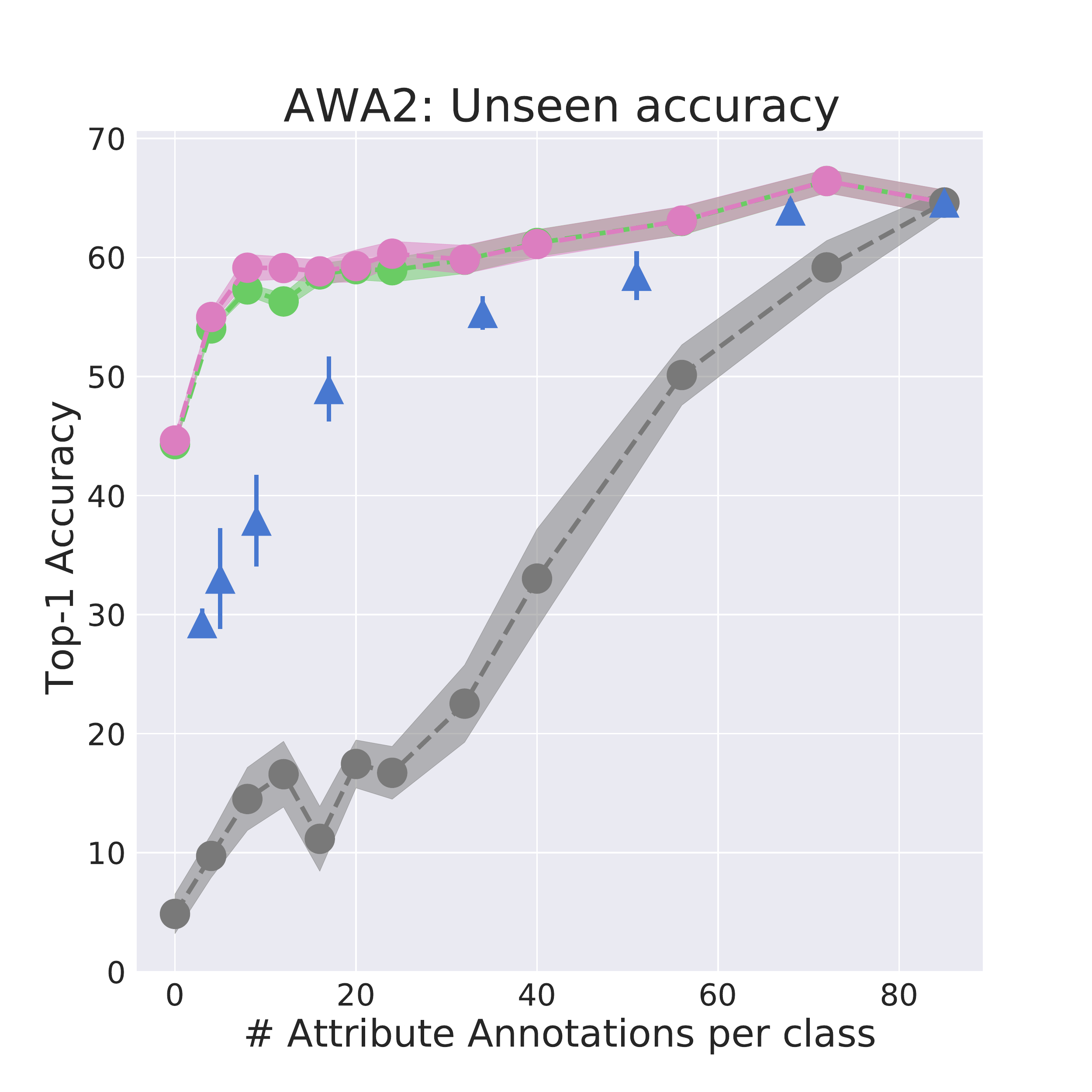}
\includegraphics[width=0.49\linewidth]{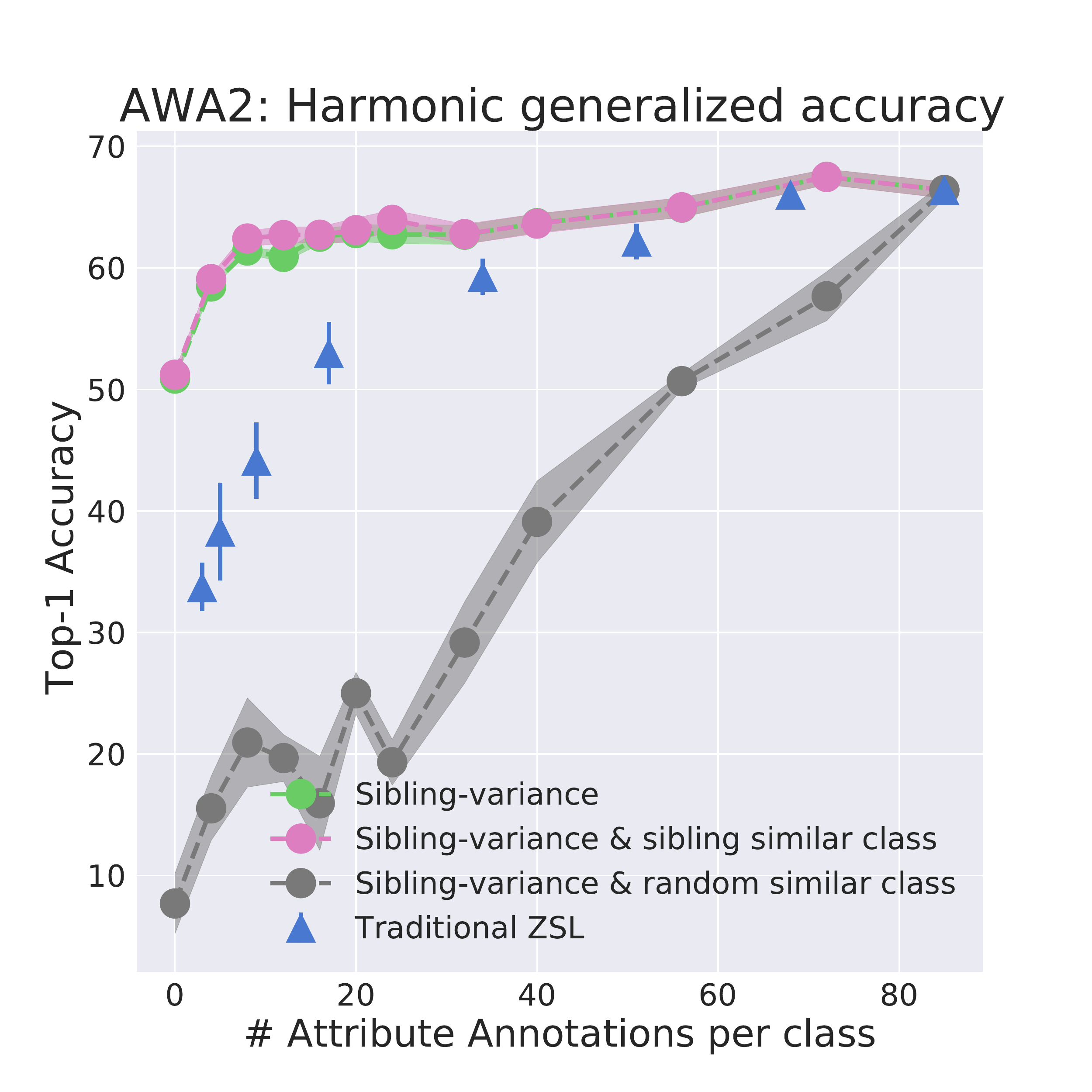}
\includegraphics[width=0.49\linewidth]{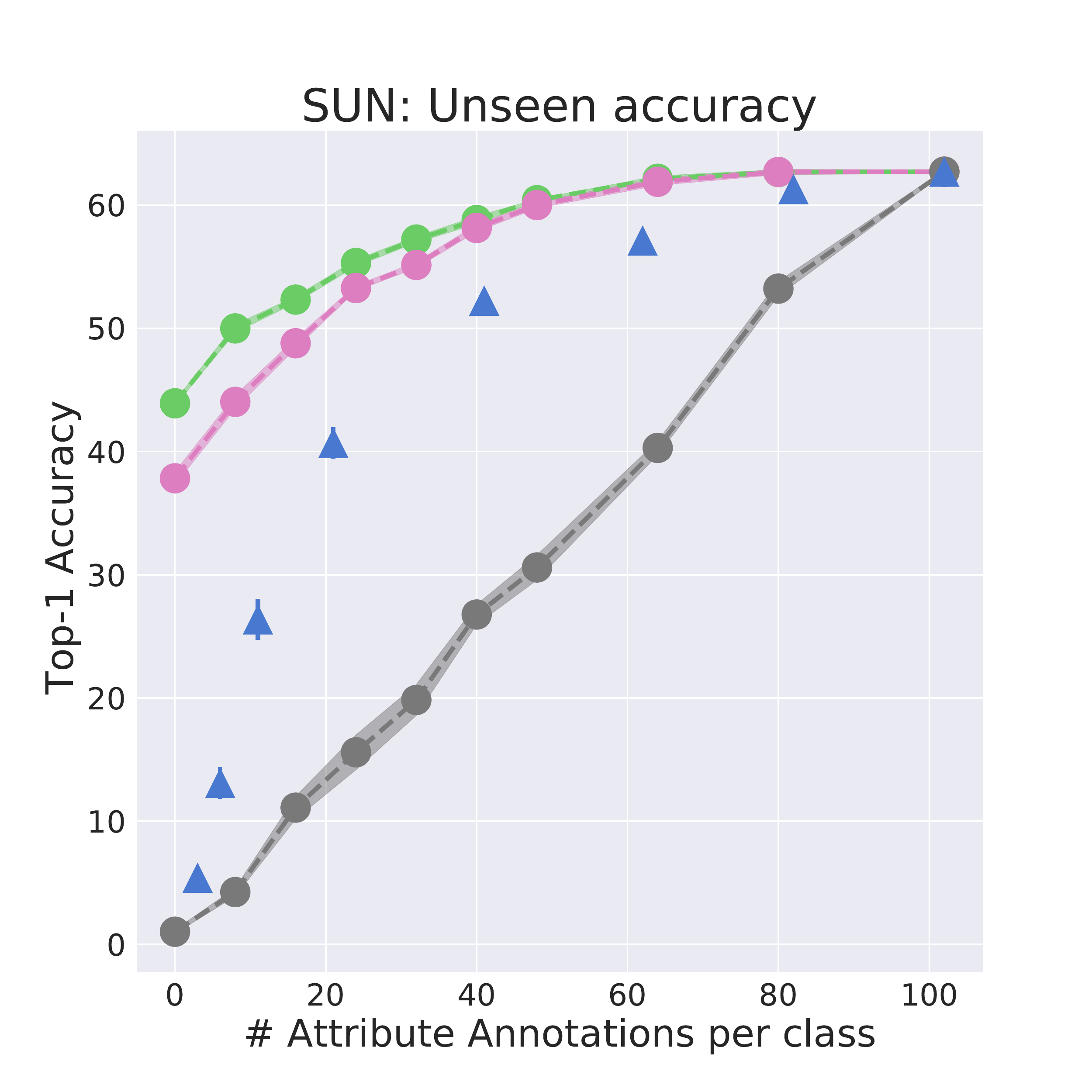}
\includegraphics[width=0.49\linewidth]{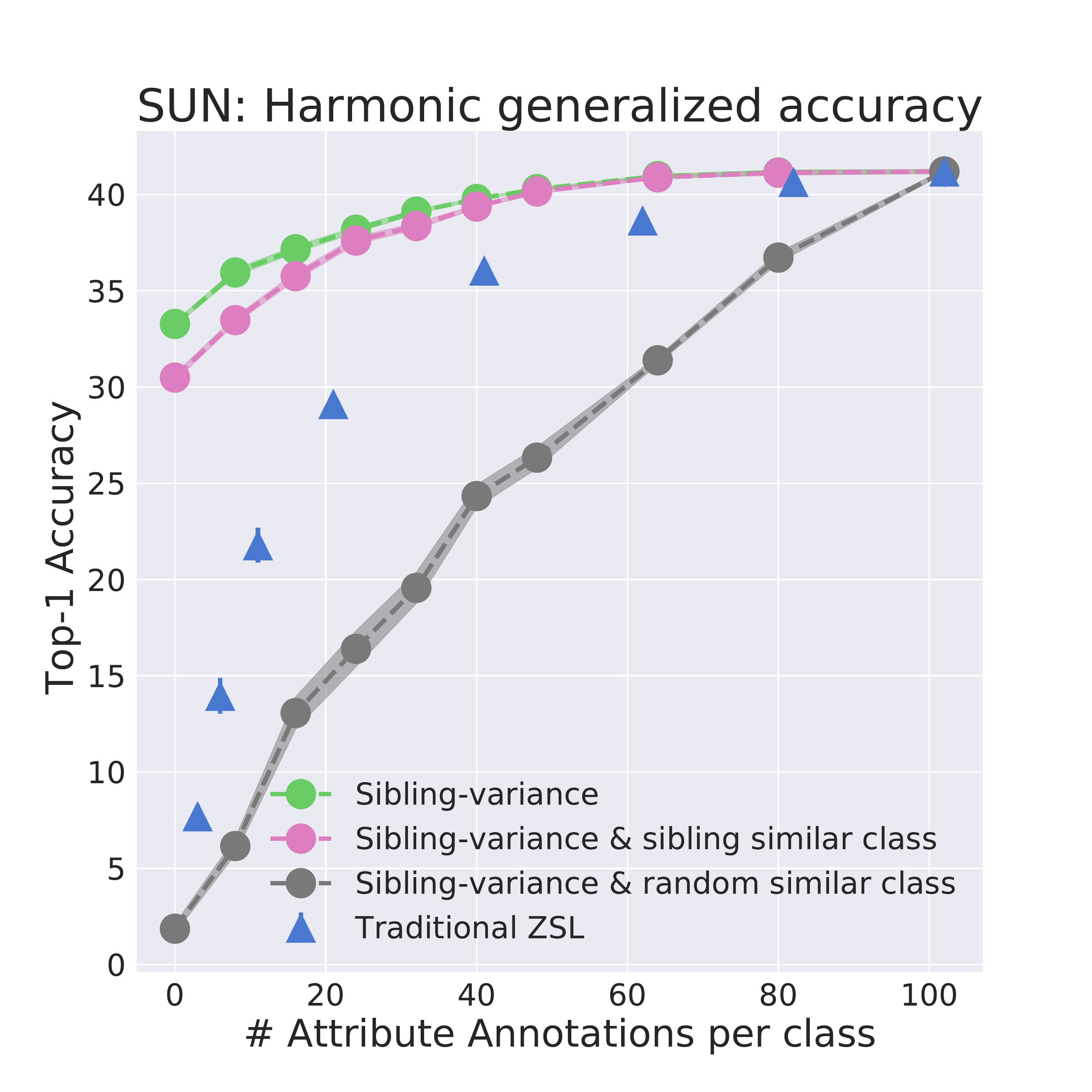}
\caption{Comparing different variants for selecting the similar class
$S(y)$ on AWA2 and SUN dataset. The model is not sensitive to the similar class as long as the selected class is not wildly different.}
\label{fig:similar}
\end{figure}

Figure~\ref{fig:similar} shows the results for \emph{\sibvar} on AWA2
and SUN.
The similar class chosen by annotators performs similar to when a
sibling base class is chosen for AWA2 and SUN. 
The performance is again not very sensitive to choosing the similar class as long as they are
not very different.

Along with the sensitivity to the choice of the similar class, we also evaluated our method with sensitivety to attribute values.
Note that incorrect or noisy attribute values will affect not just our proposed active ZSL but also the traditional non-interactive ZSL. 
With $10\%$ noise in the novel attributes, when all attributes are provided, both our method and traditional ZSL see a $\sim3\%$ drop in performance. 
With partial attribute annotations (5 per class), our proposed annotation strategy ($\sim1\%$ drop) fares much better than traditional ZSL annotation ($\sim 5\%$ drop).

\section{t-SNE Visualizations for More Classes and Dataset}\label{sec:tsne}

\begin{figure*}[h!]
\centering
\includegraphics[width=0.17\linewidth]{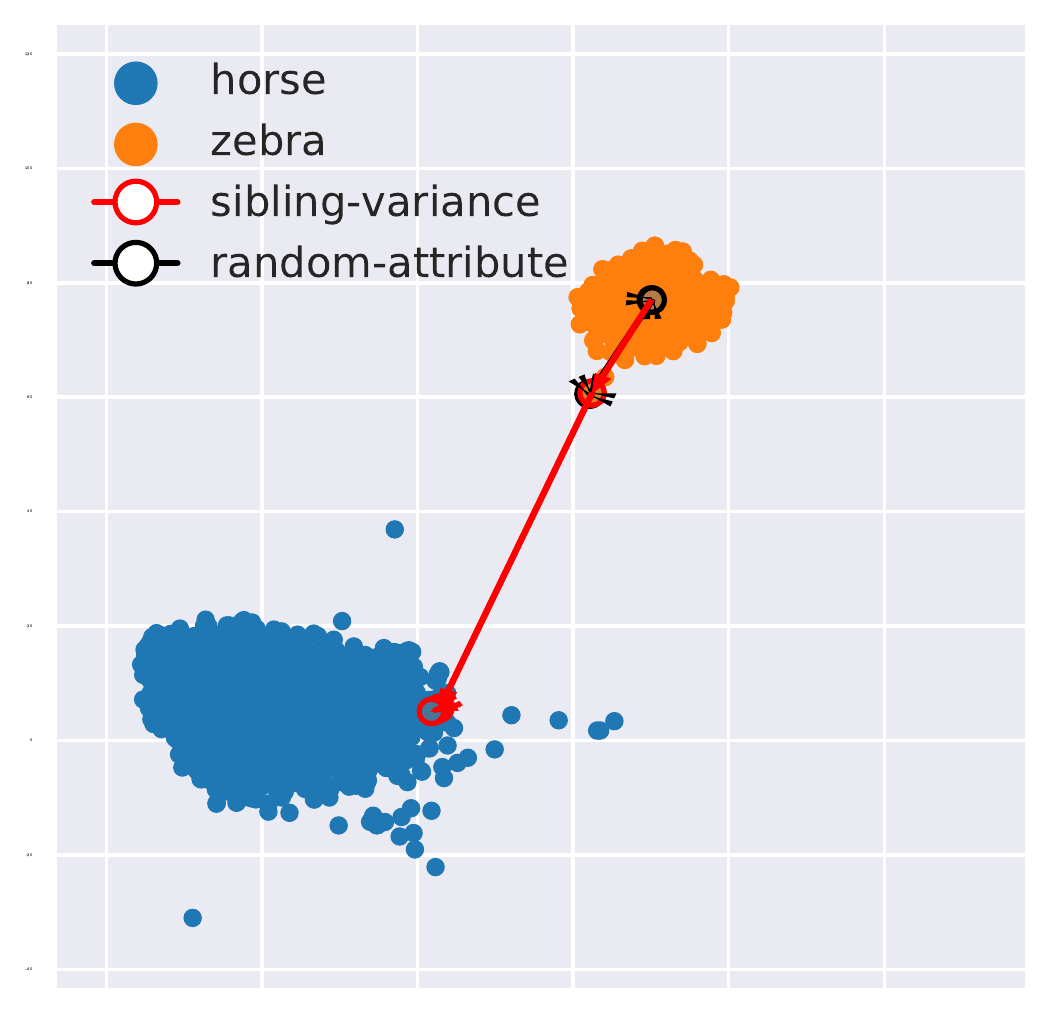}
\includegraphics[width=0.17\linewidth]{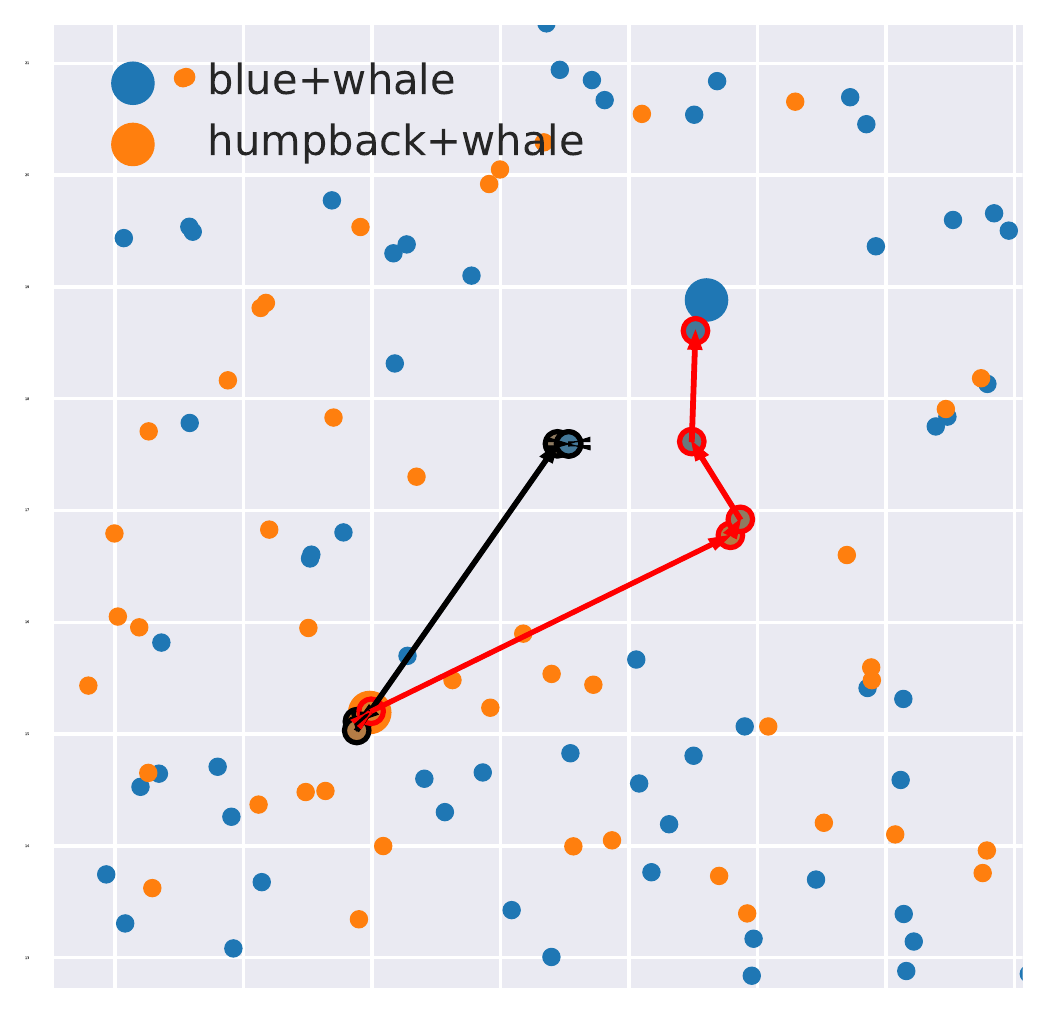}
\includegraphics[width=0.17\linewidth]{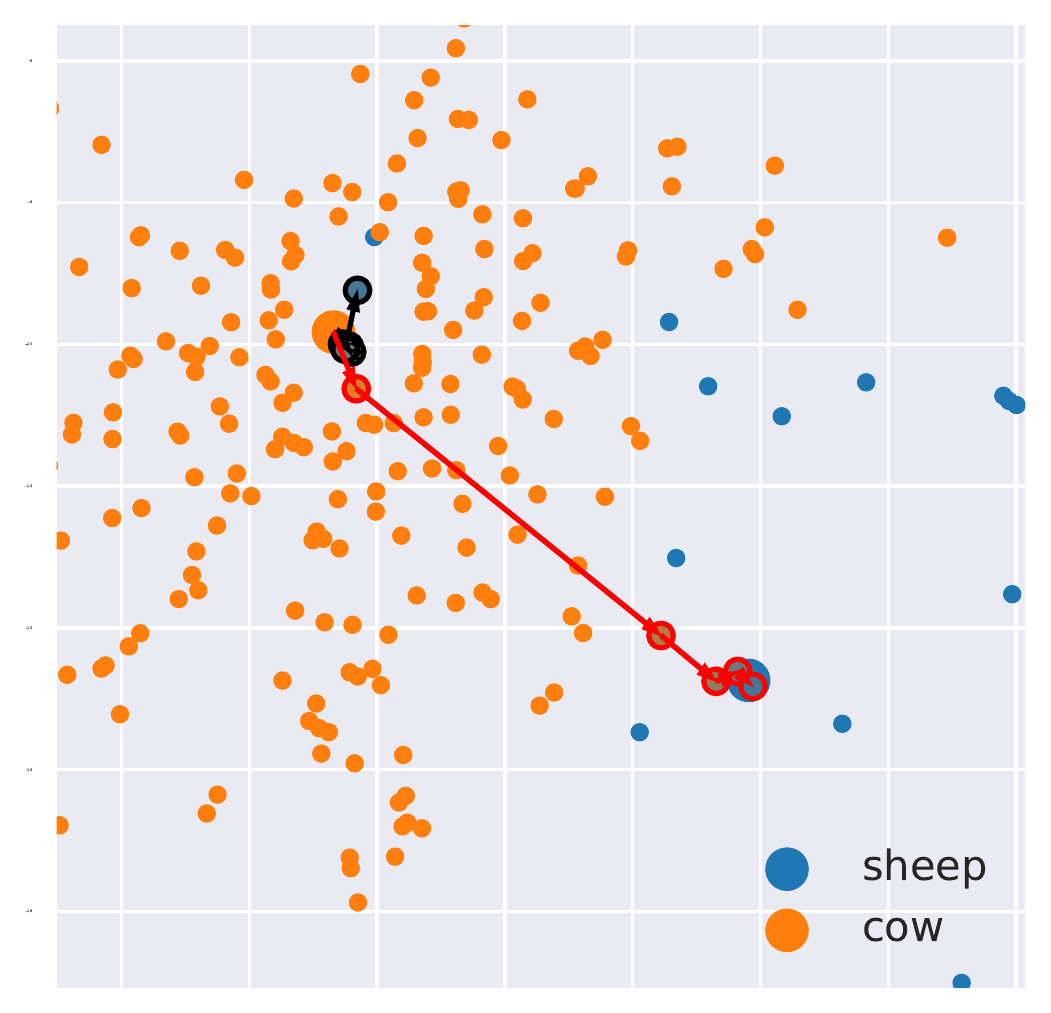}
\includegraphics[width=0.17\linewidth]{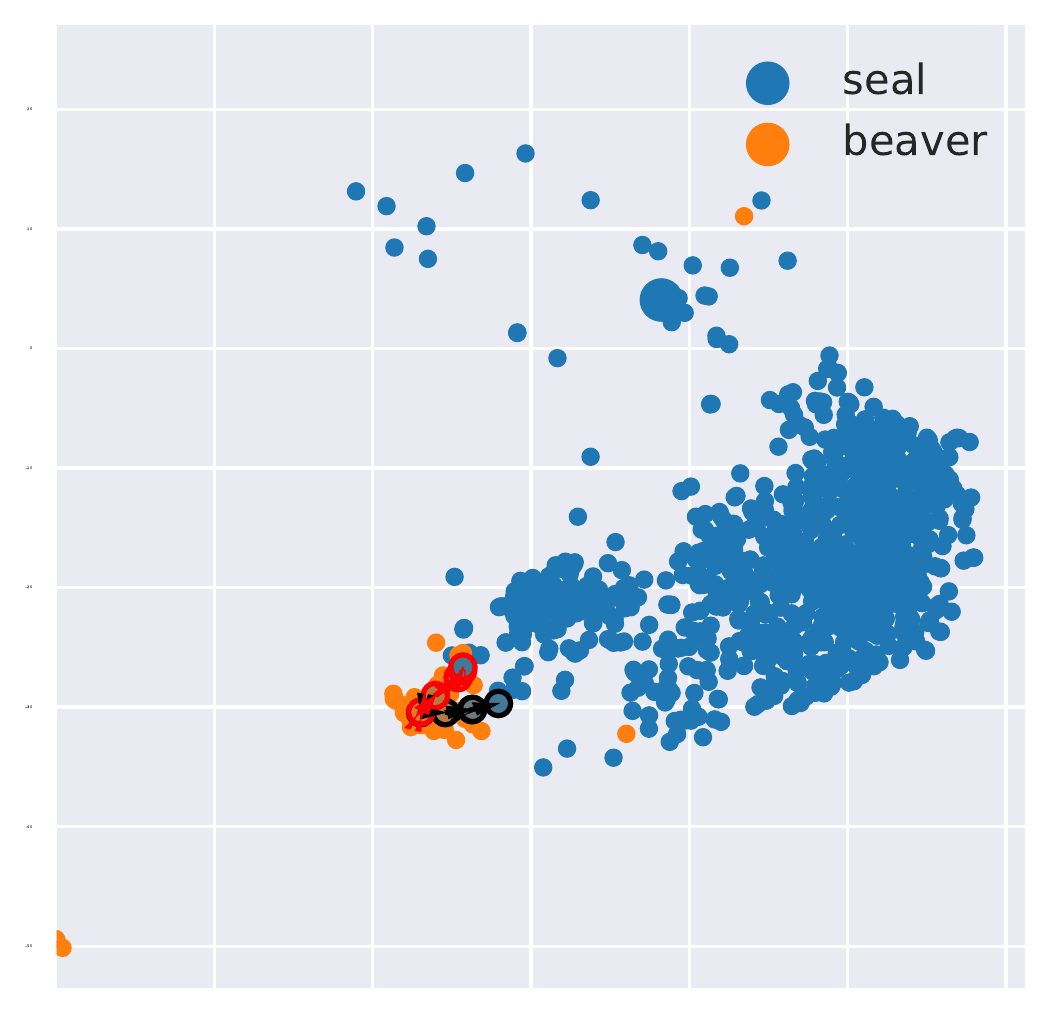}
\includegraphics[width=0.17\linewidth]{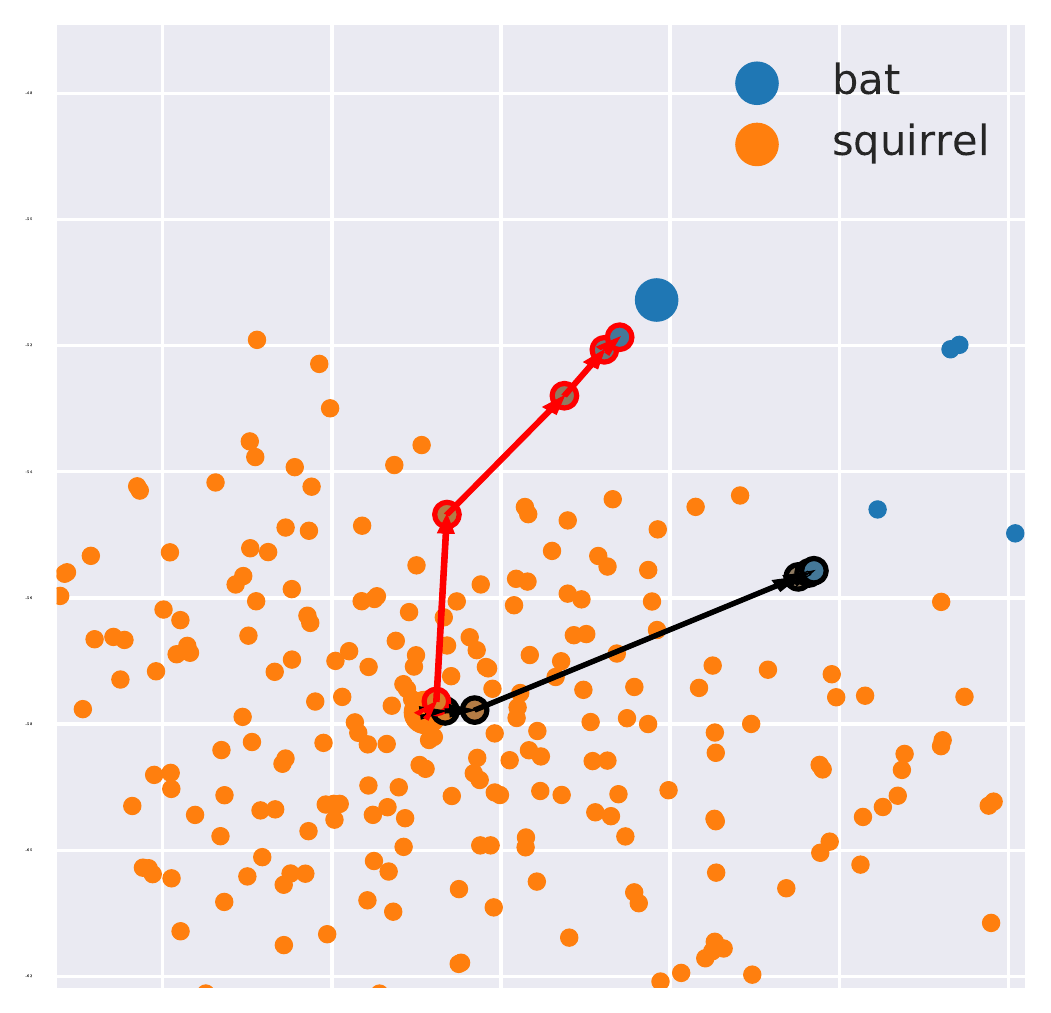}
\includegraphics[width=0.17\linewidth]{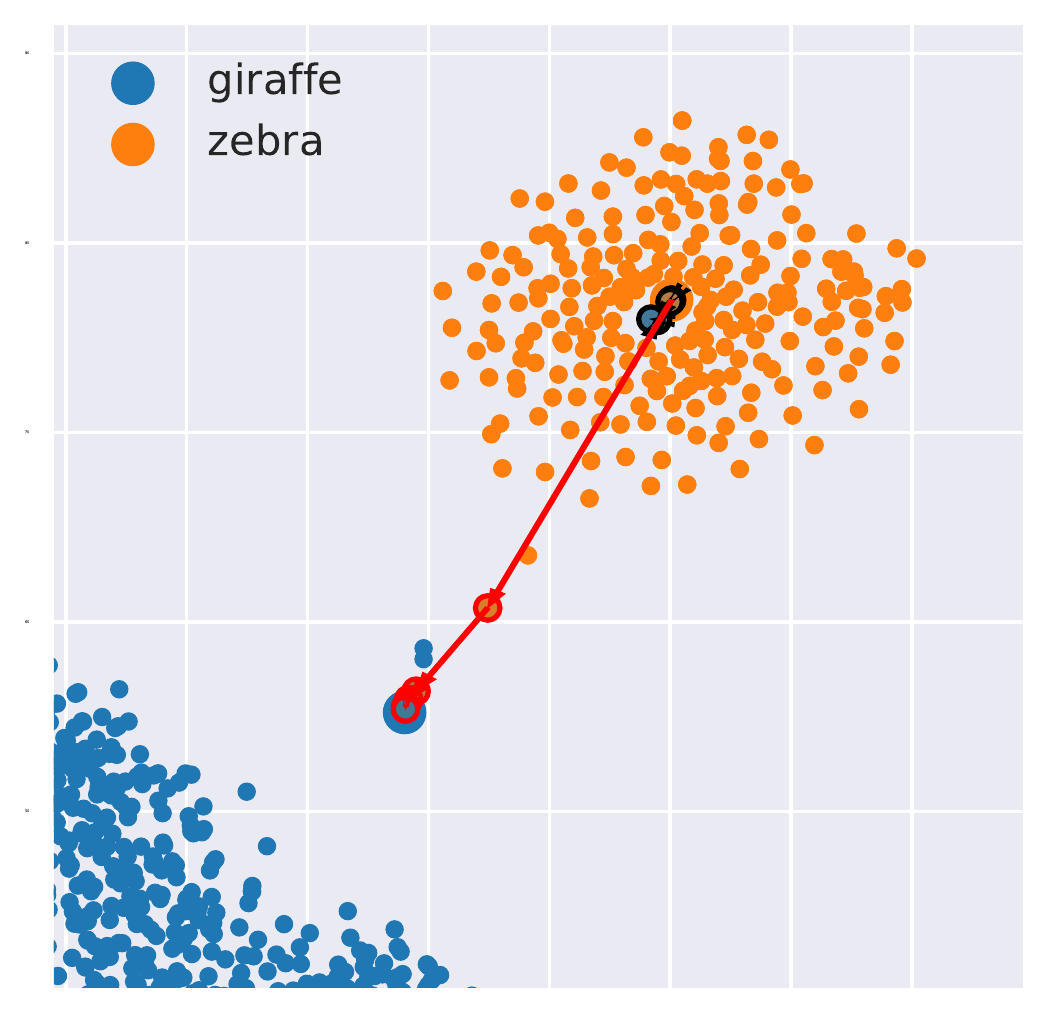}
\includegraphics[width=0.17\linewidth]{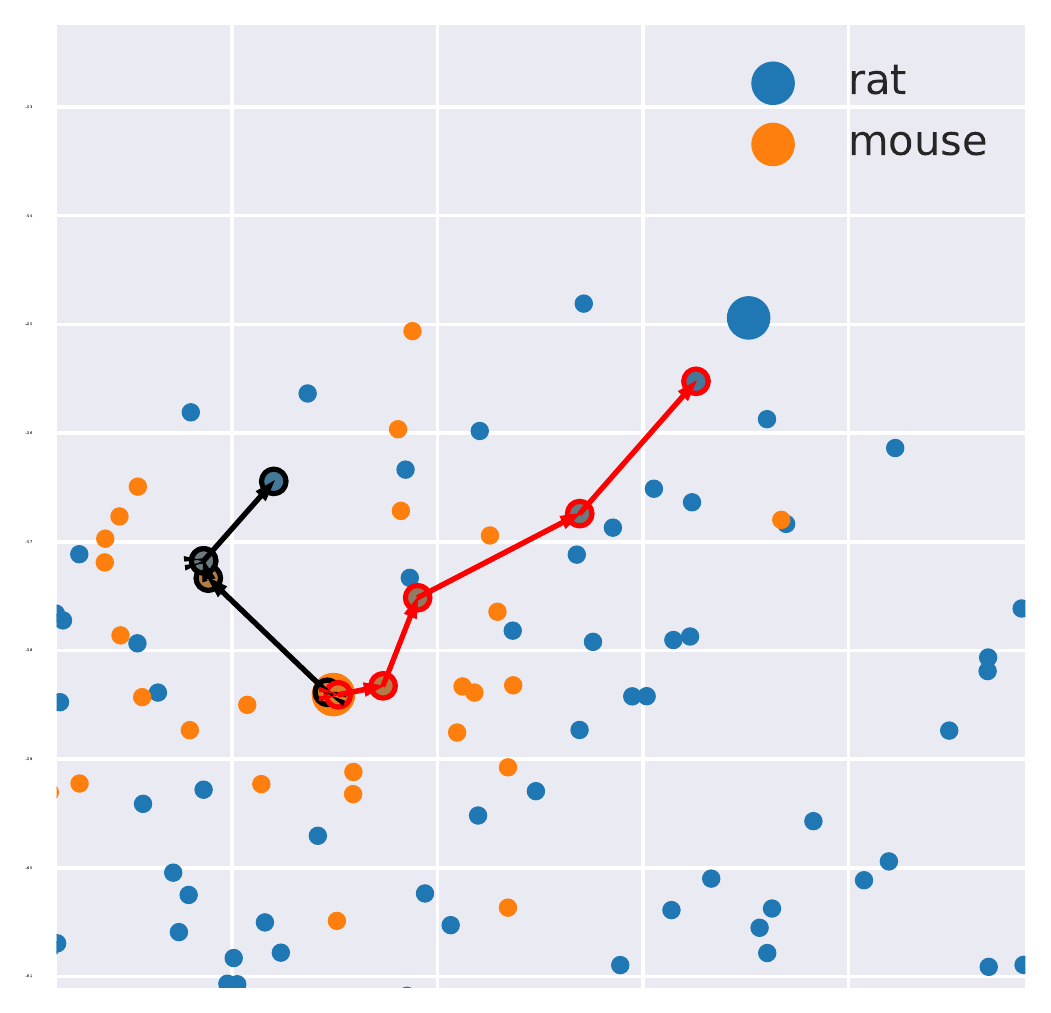}
\includegraphics[width=0.17\linewidth]{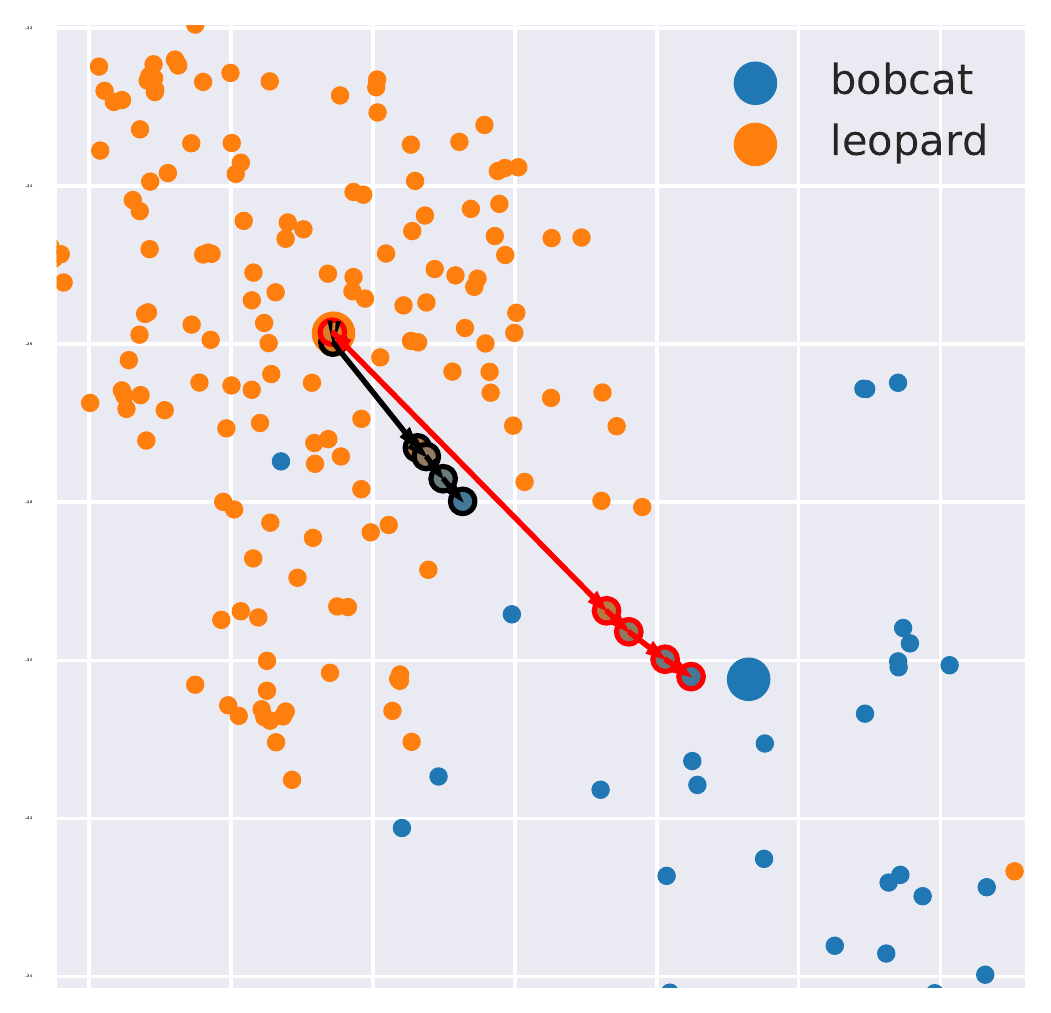}
\includegraphics[width=0.17\linewidth]{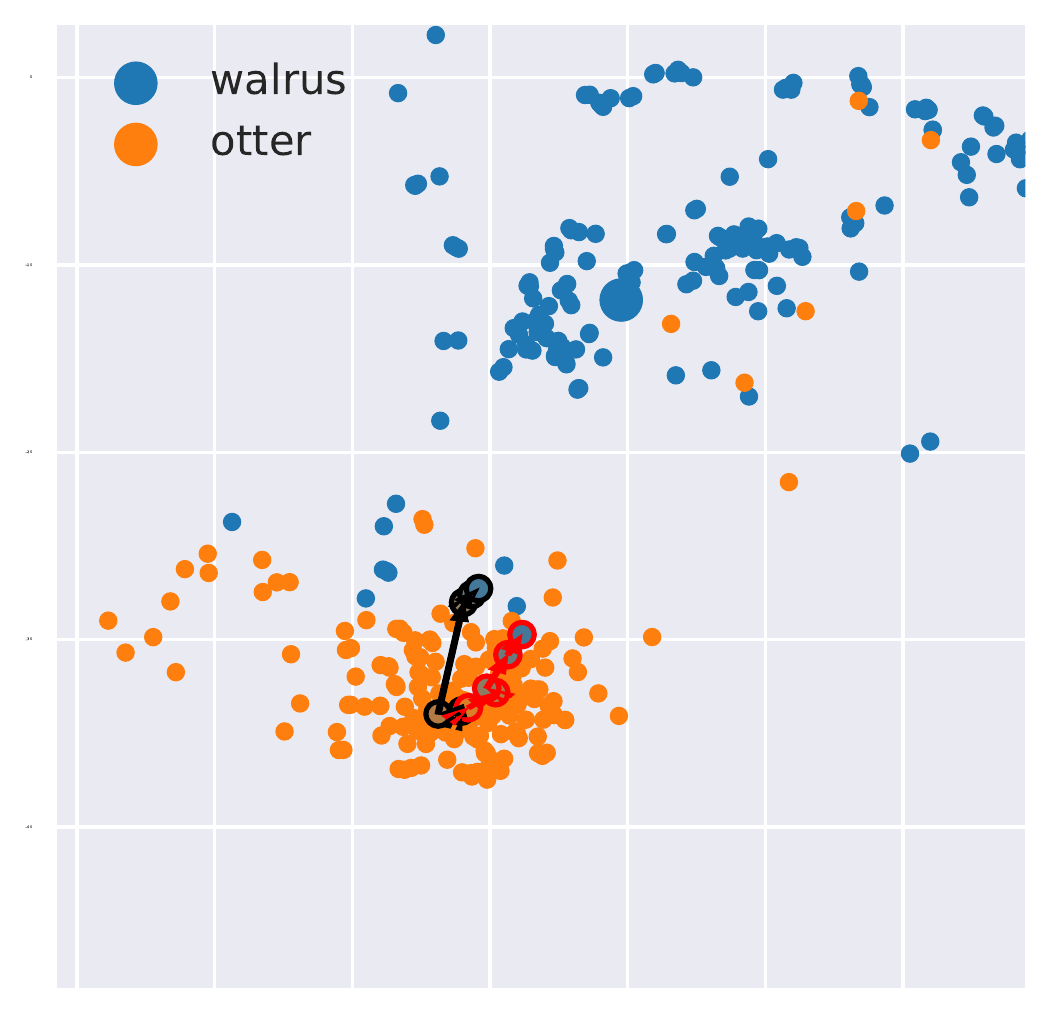}
\includegraphics[width=0.17\linewidth]{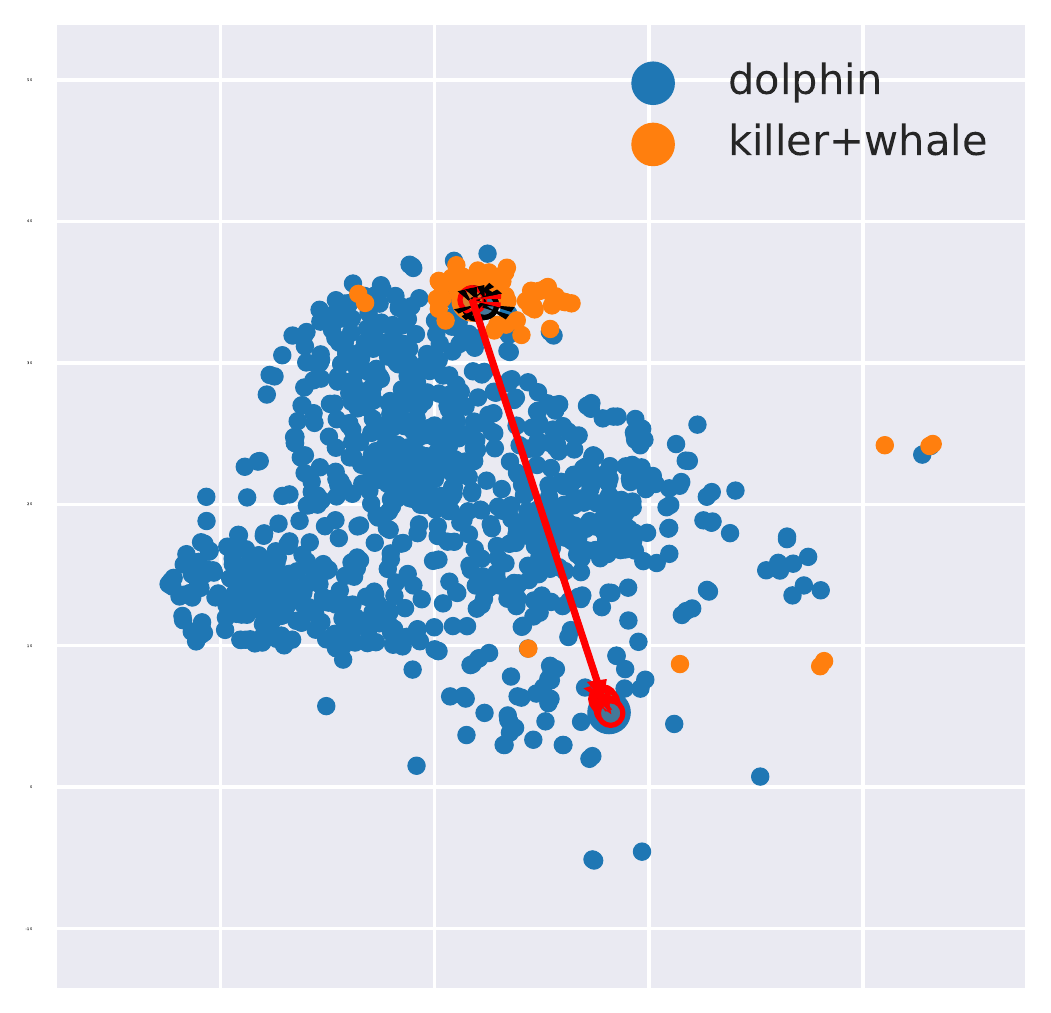}
\caption{t-SNE visualizations. For all 10 AWA2 novel classes and corresponding similar base classes. 
Smaller dots represent test images and larger dots represent class
attribute embeddings.
Red edges show the progression of novel class attributes as learners
interact using \emph{\sibvar}. Dots with black edges show the
progression with the \emph{random} function. 
Both methods start at the base class attribute descriptor, and aim to
reach to the novel class descriptor with as few interactions as possible. 
In most cases \emph{\sibvar} reaches closer to the novel class descriptor quicker in contrast to \emph{random}.
}
\label{fig:tsne_awa}
\end{figure*}

Figure \ref{fig:tsne_awa} show the progression of \emph{\sibvar} and random attributes for all 10 AWA2 novel classes. Note that in the standard split of AWA2, the classes are split in a way that sometimes no good similar classes could be found. For example, both seal and walrus are in the test split and hence the annotators chose beaver and walrus as similar classes. Similarly no good base class is there for giraffe and bat so the annotators had to chose zebra and squirrel. 
Nonetheless the faster progression towards novel classes' images and attributes can be seen for the classes when using \emph{\sibvar} over random attributes.

\begin{figure*}[h!]
\centering
\includegraphics[width=0.17\linewidth]{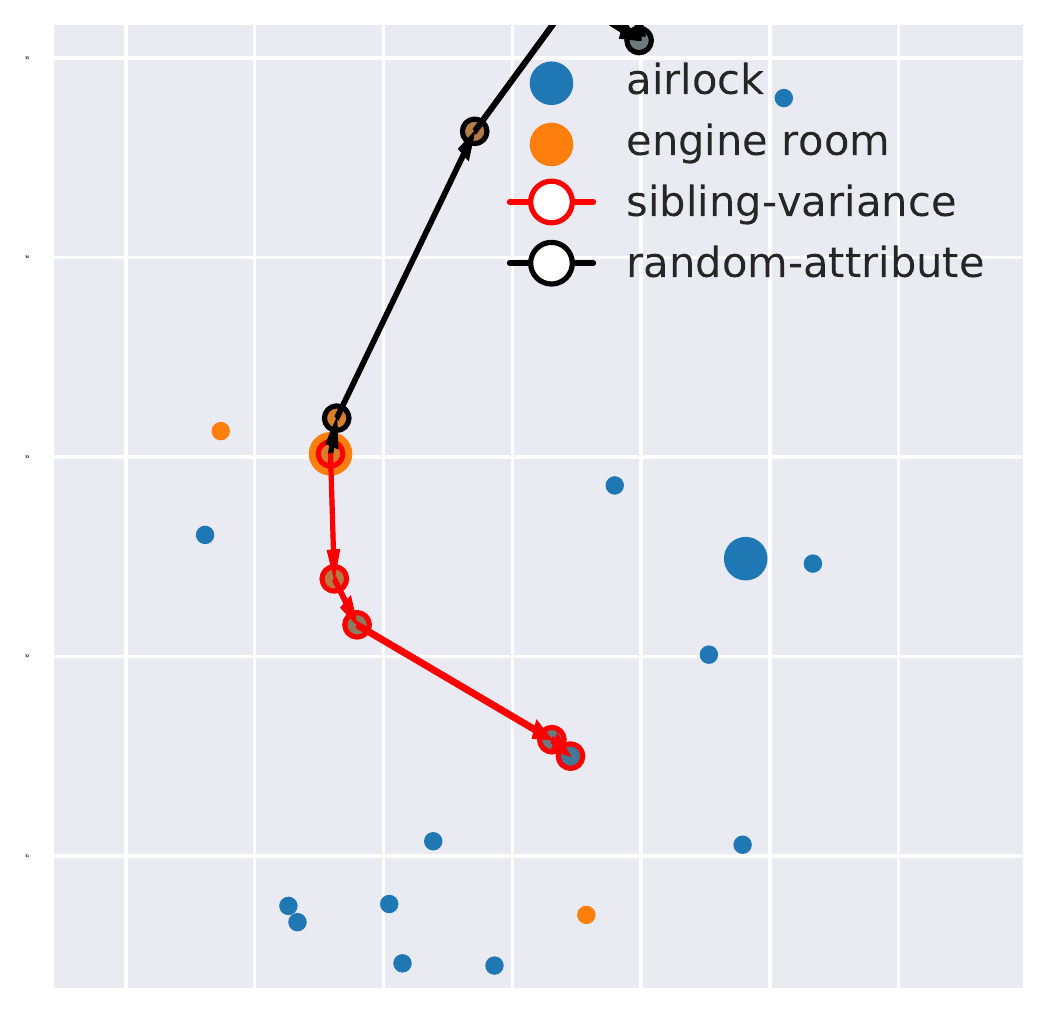}
\includegraphics[width=0.17\linewidth]{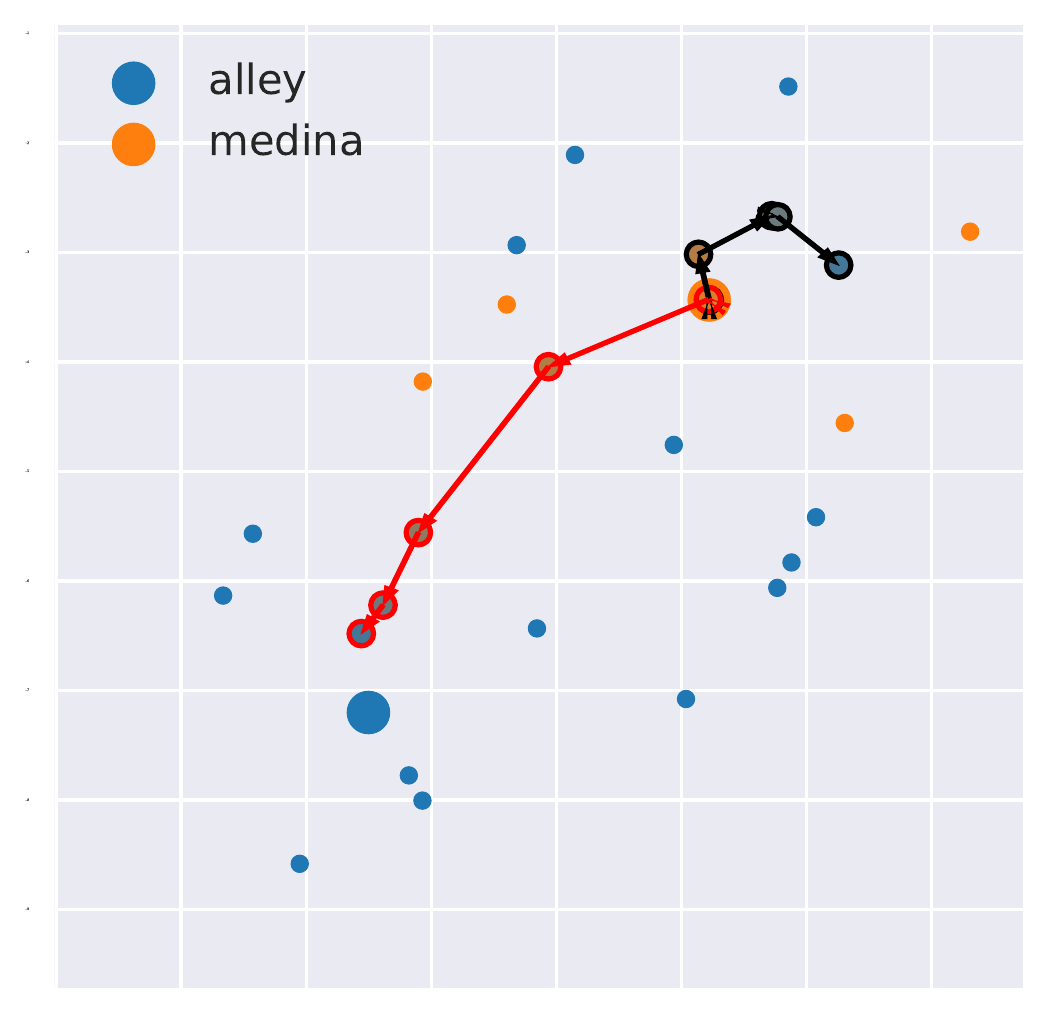}
\includegraphics[width=0.17\linewidth]{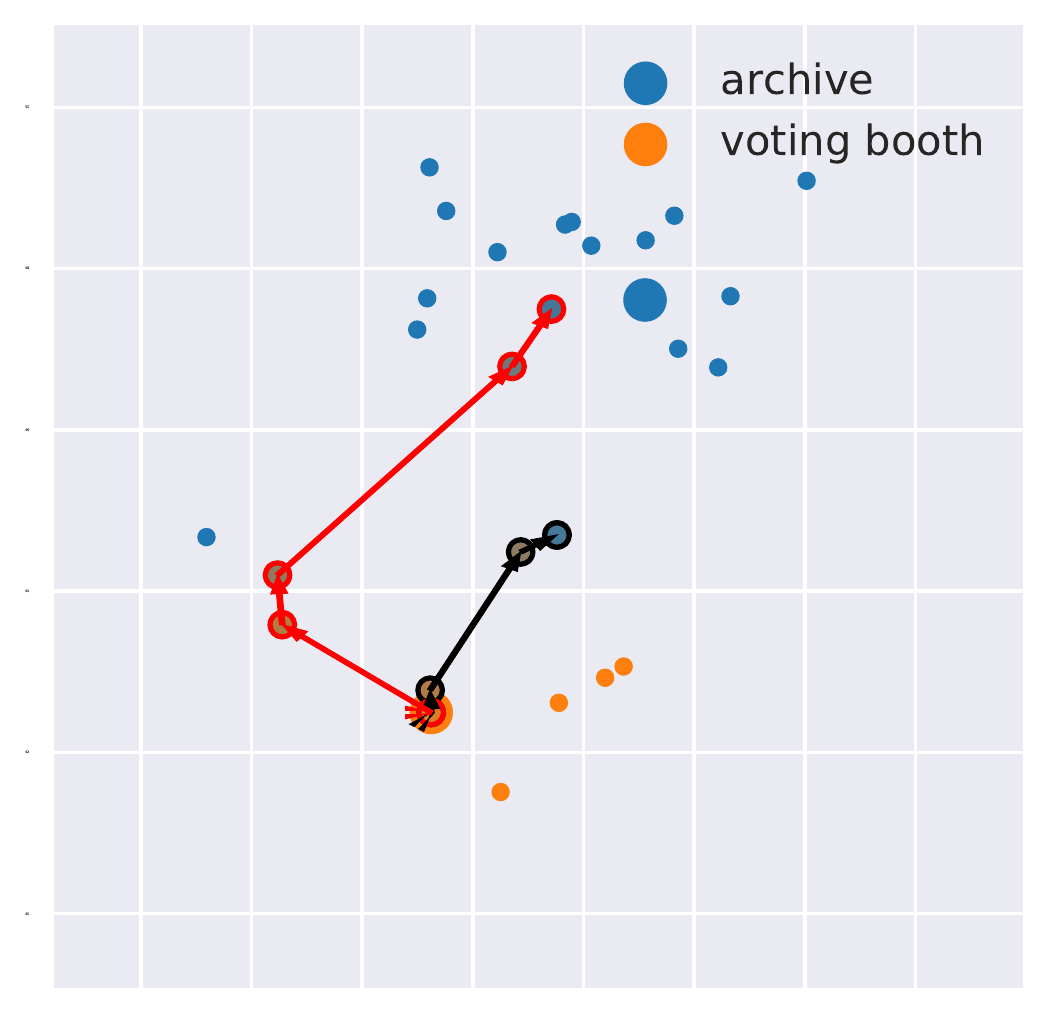}
\includegraphics[width=0.17\linewidth]{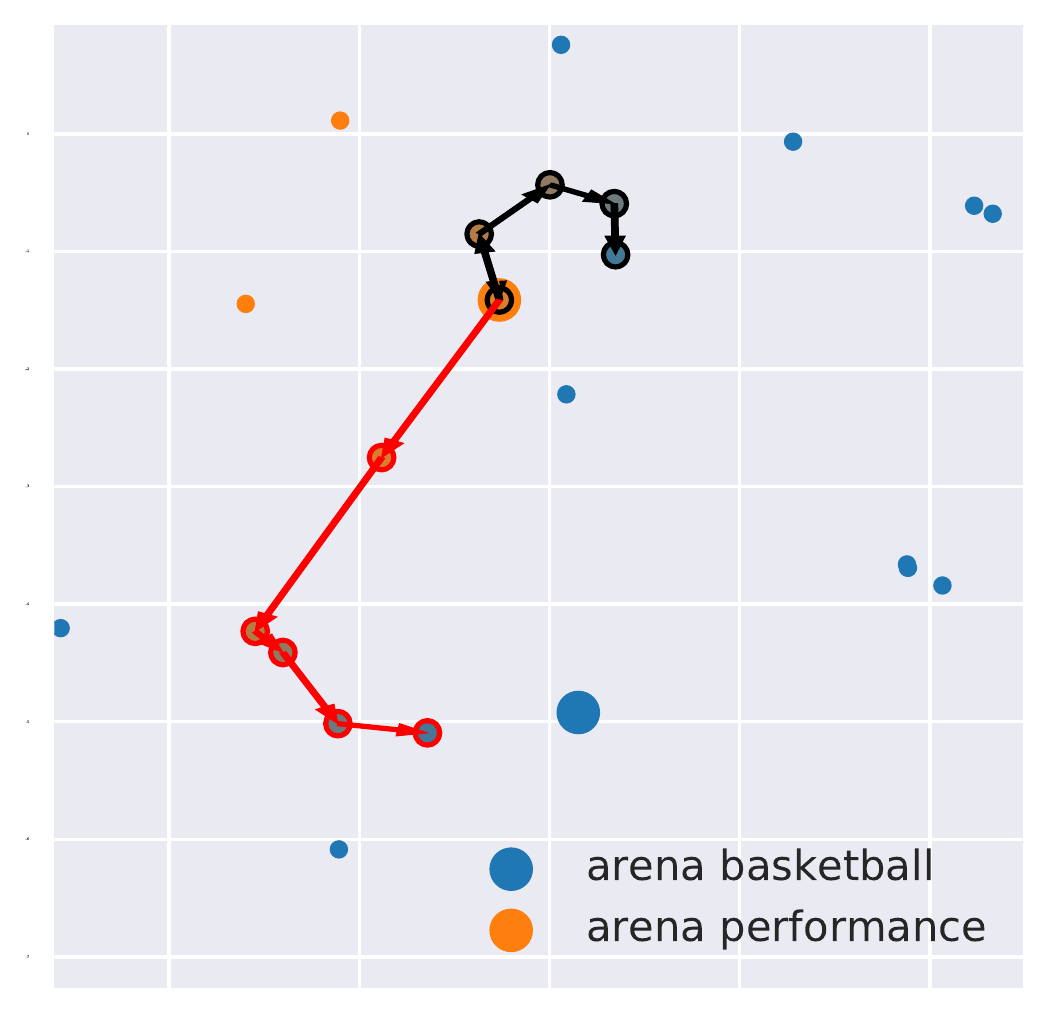}
\includegraphics[width=0.17\linewidth]{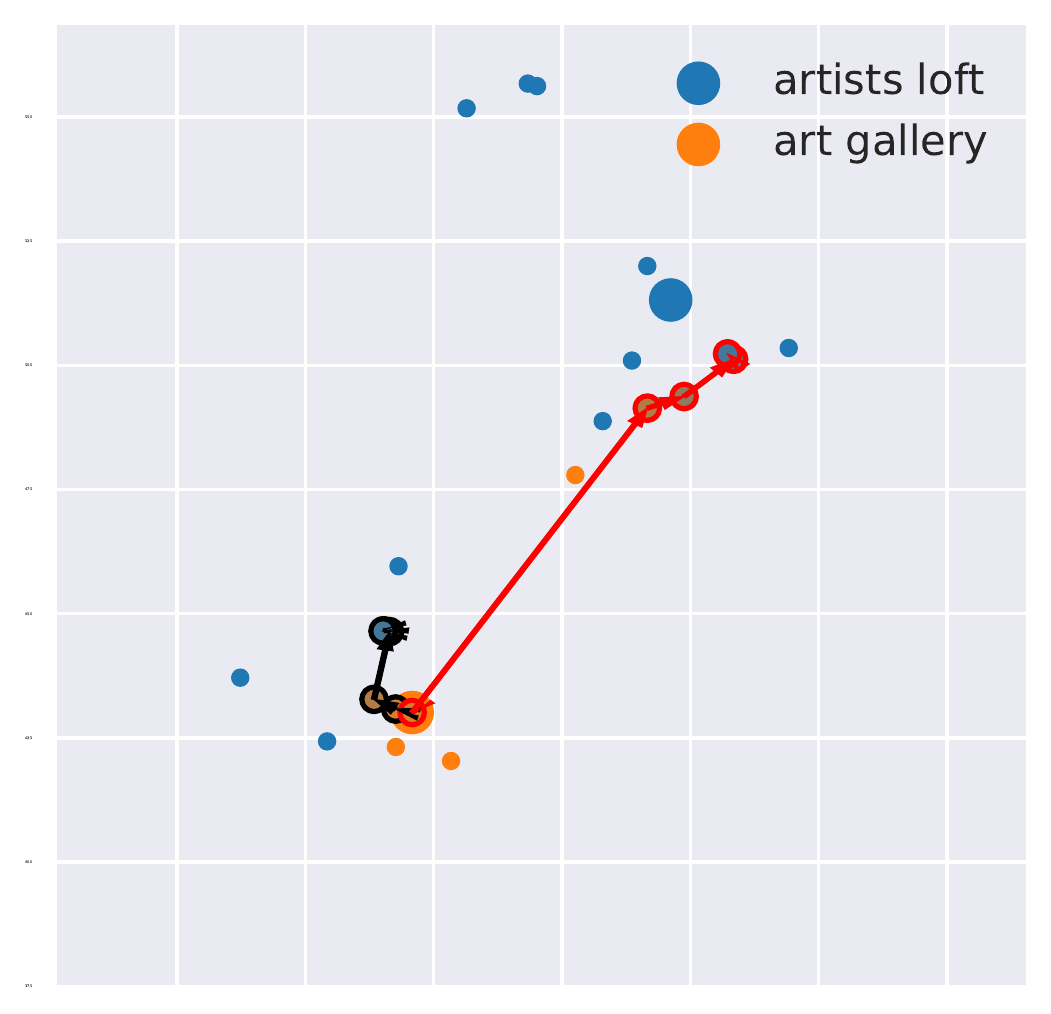}
\includegraphics[width=0.17\linewidth]{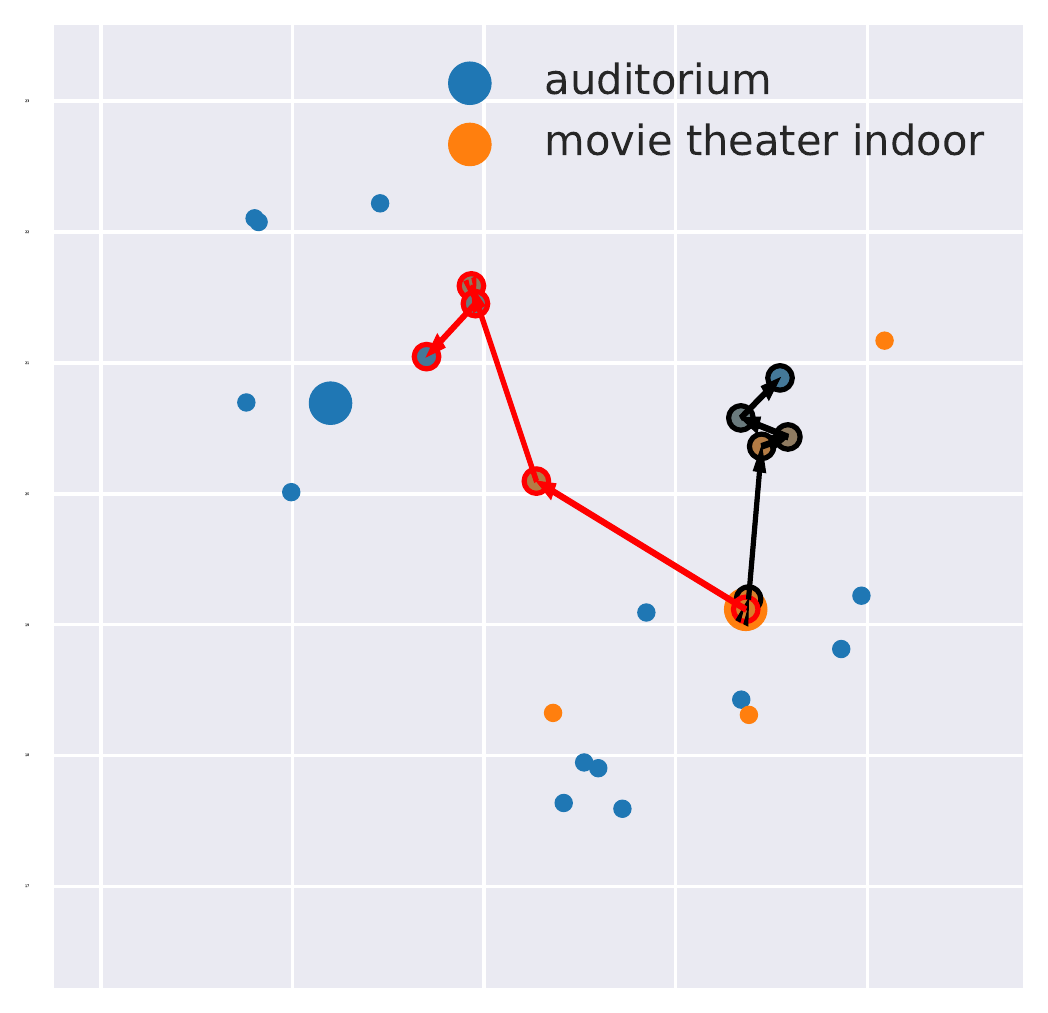}
\includegraphics[width=0.17\linewidth]{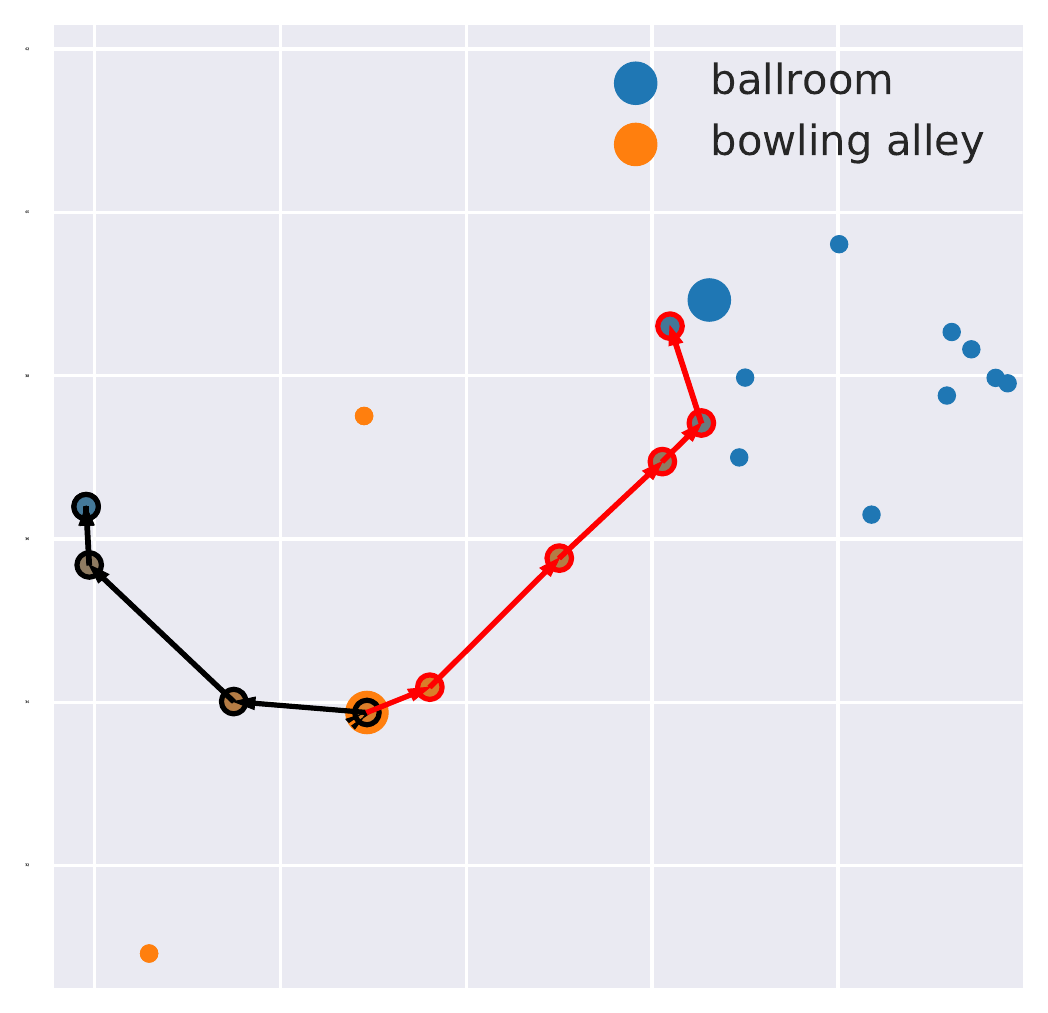}
\includegraphics[width=0.17\linewidth]{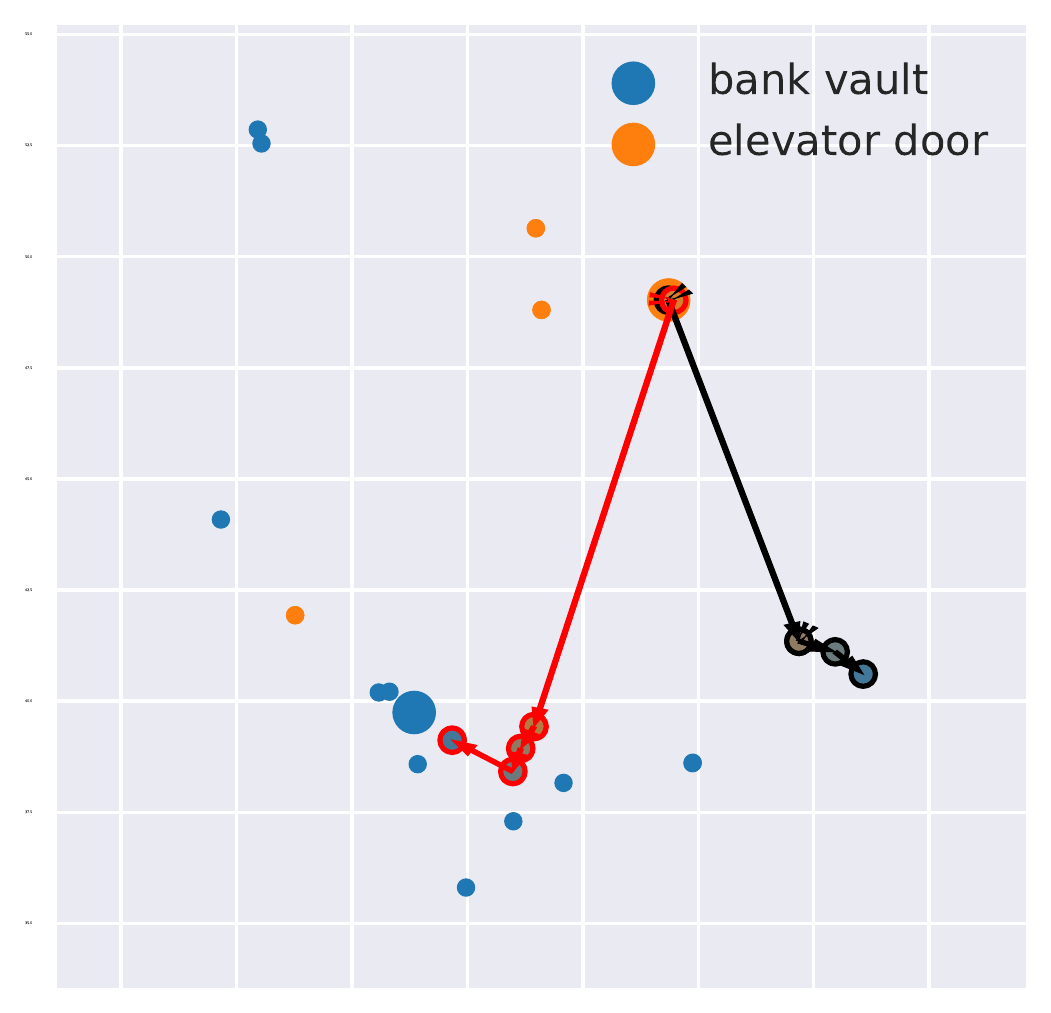}
\includegraphics[width=0.17\linewidth]{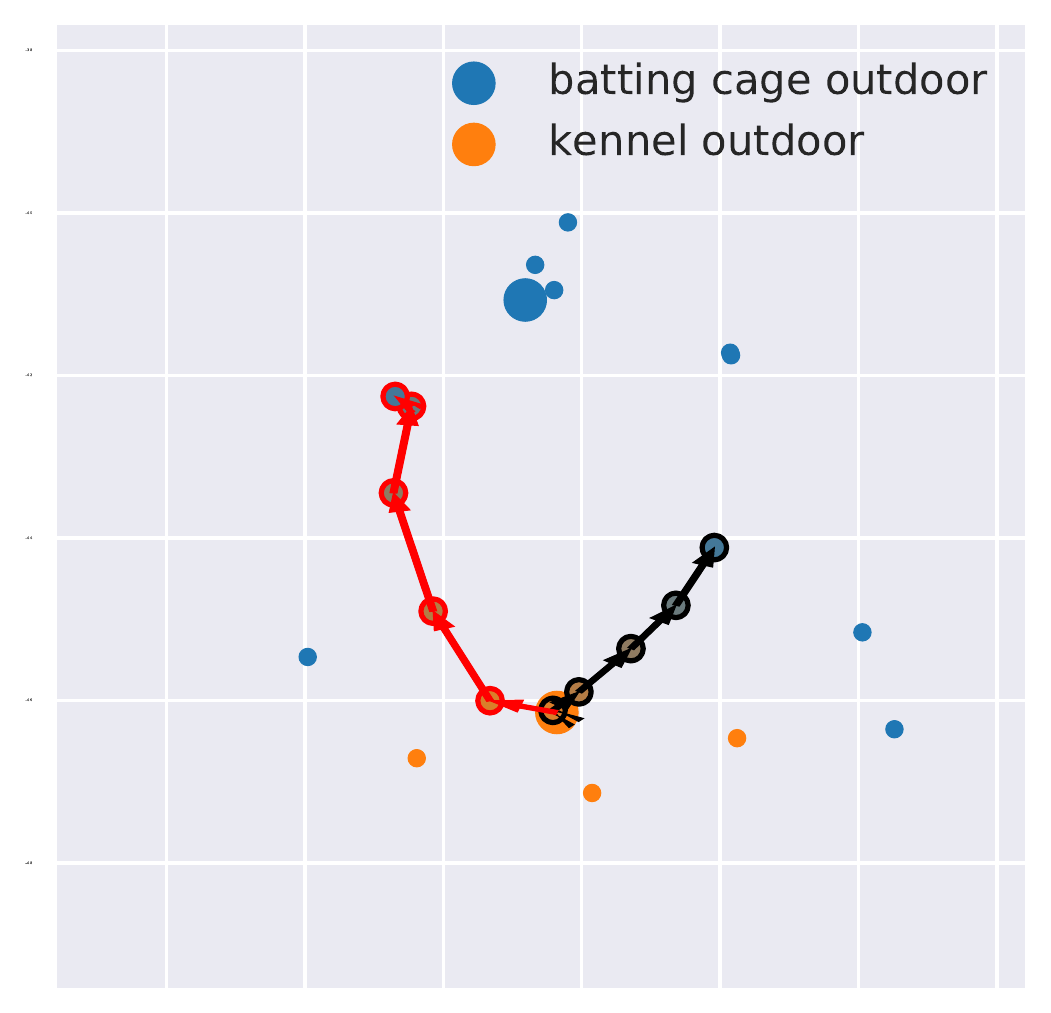}
\includegraphics[width=0.17\linewidth]{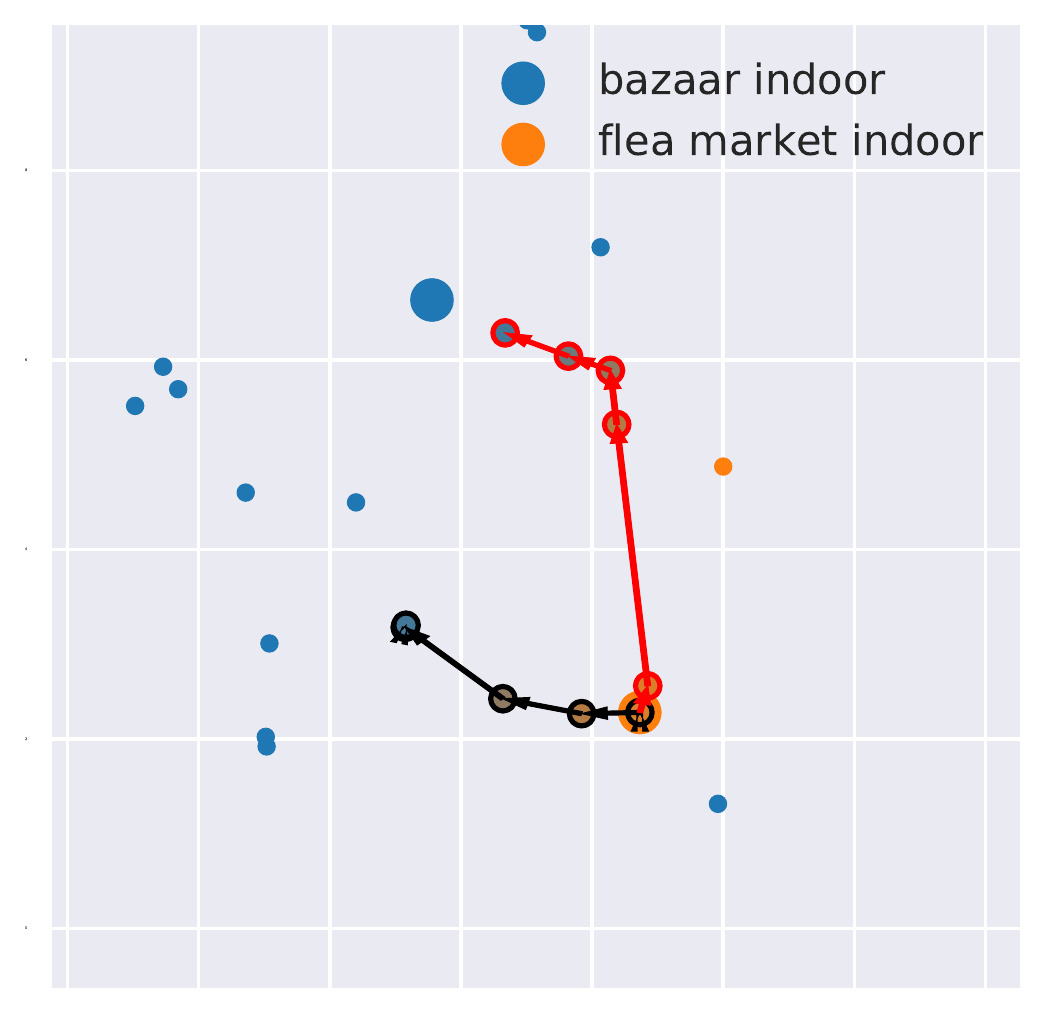}
\includegraphics[width=0.17\linewidth]{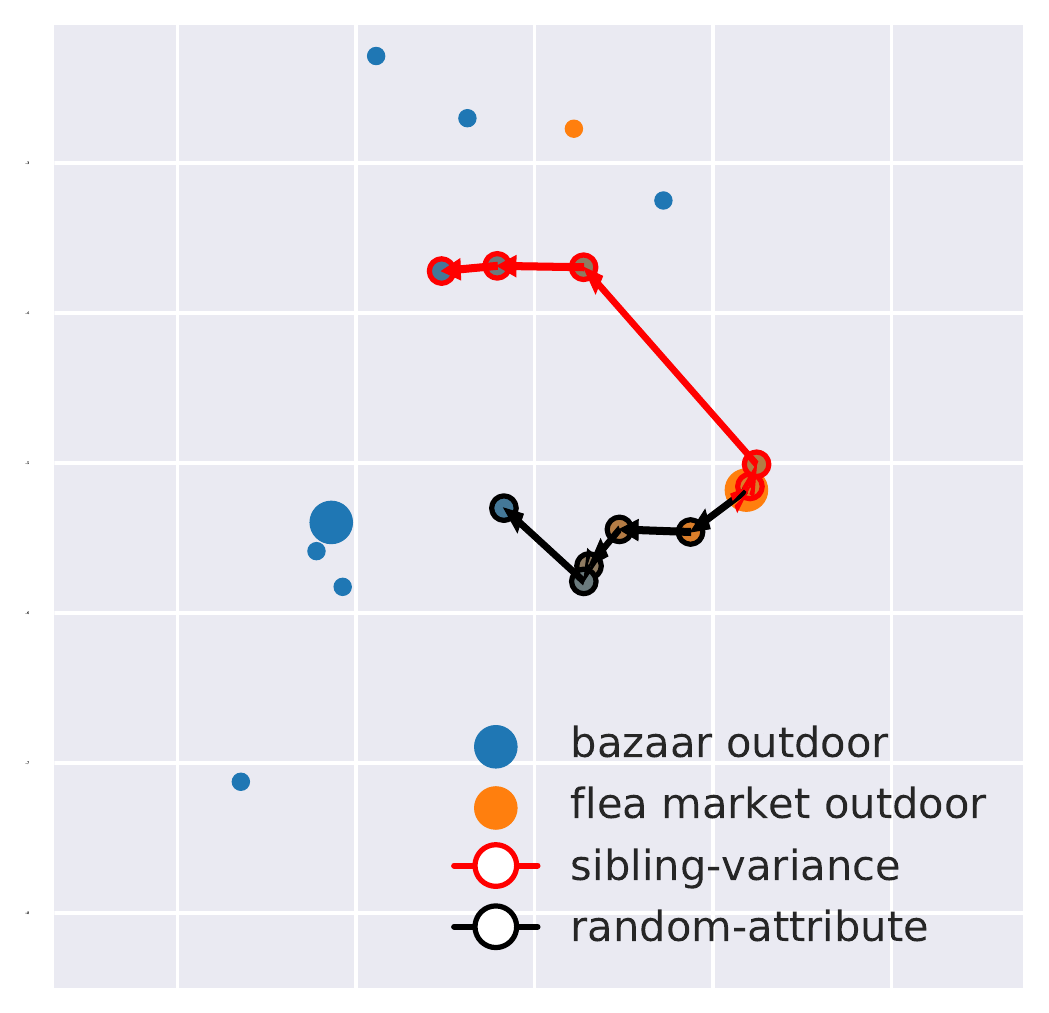}
\includegraphics[width=0.17\linewidth]{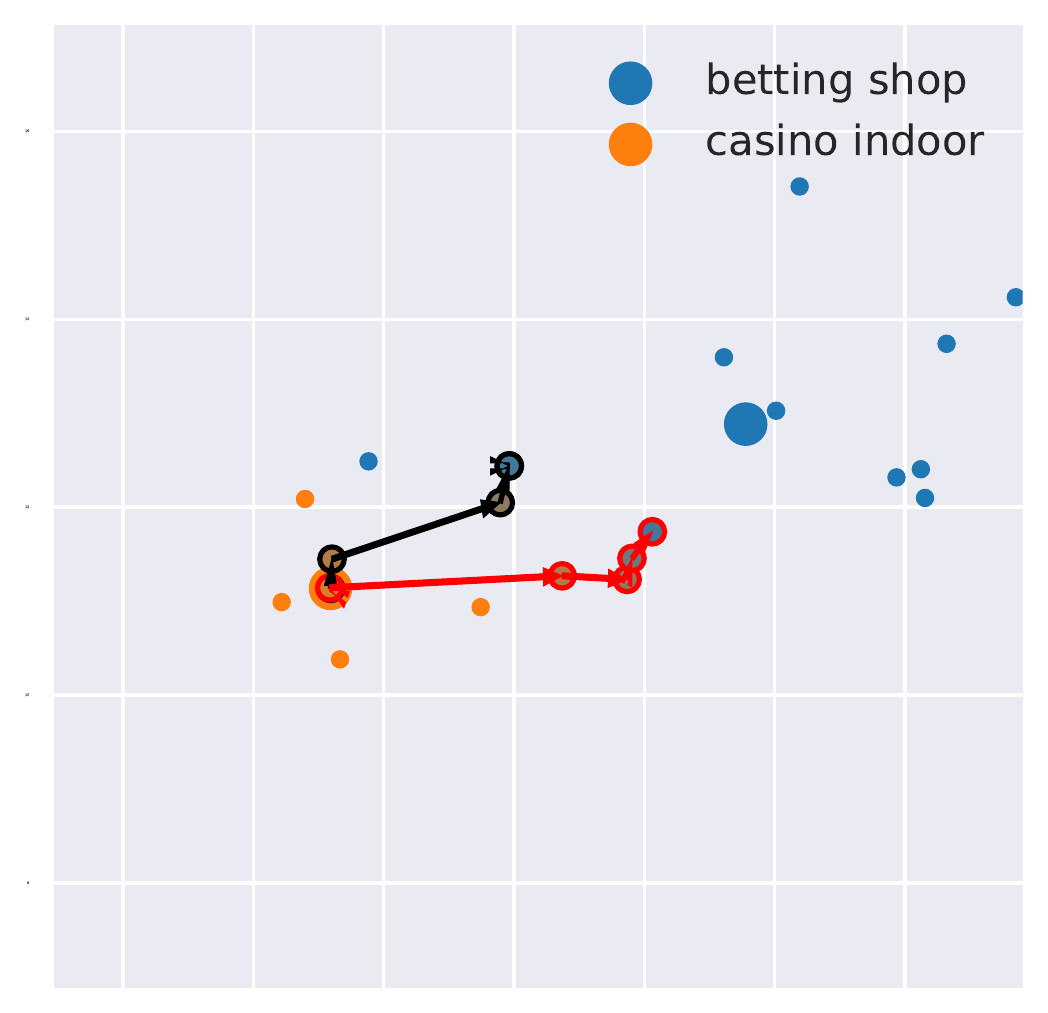}
\includegraphics[width=0.17\linewidth]{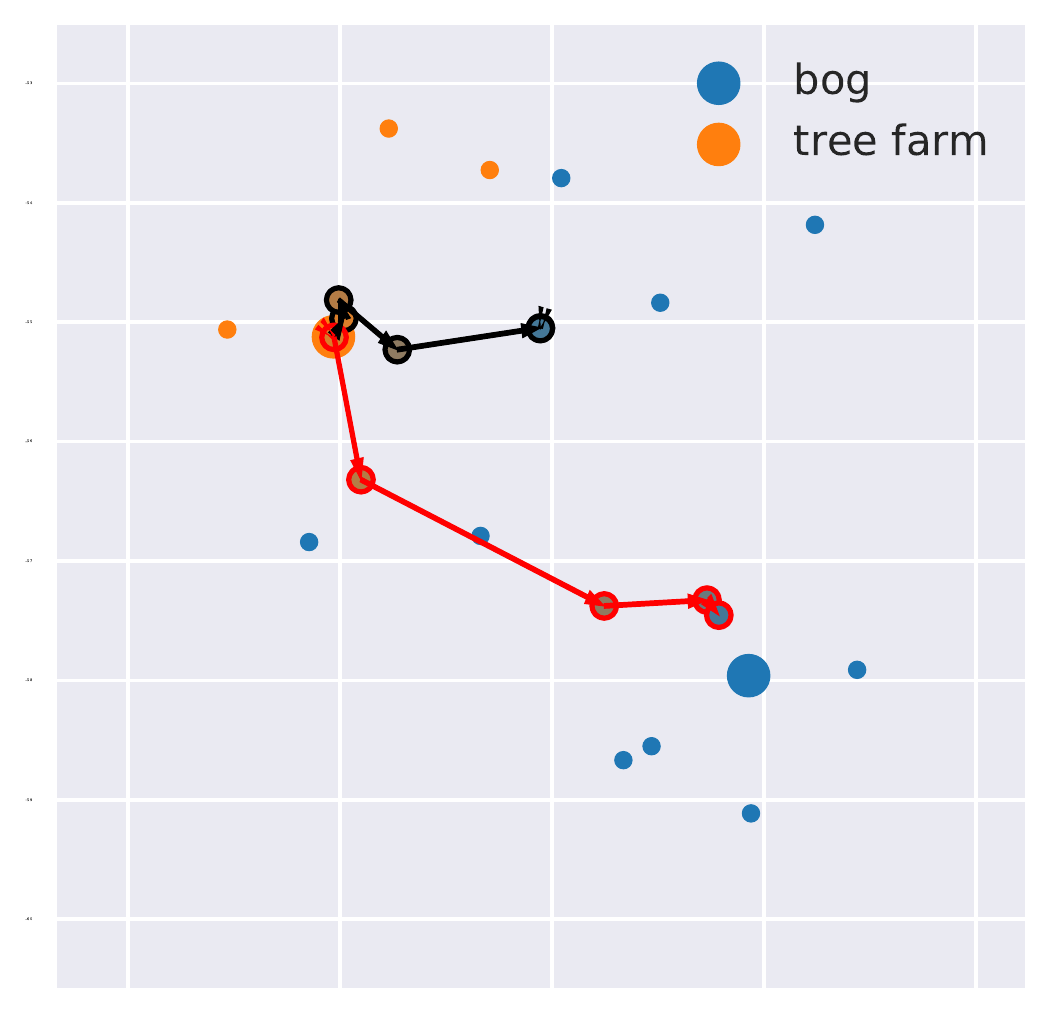}
\includegraphics[width=0.17\linewidth]{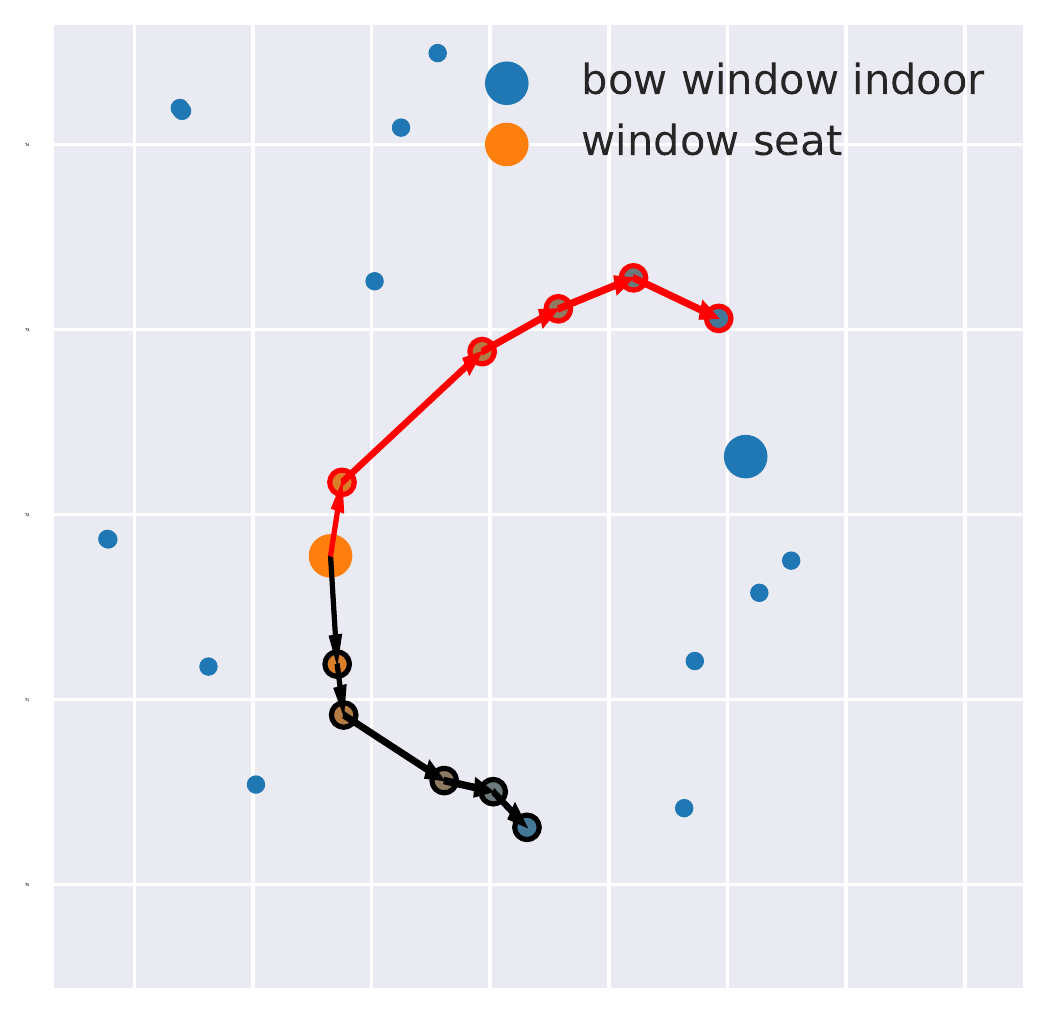}
\includegraphics[width=0.17\linewidth]{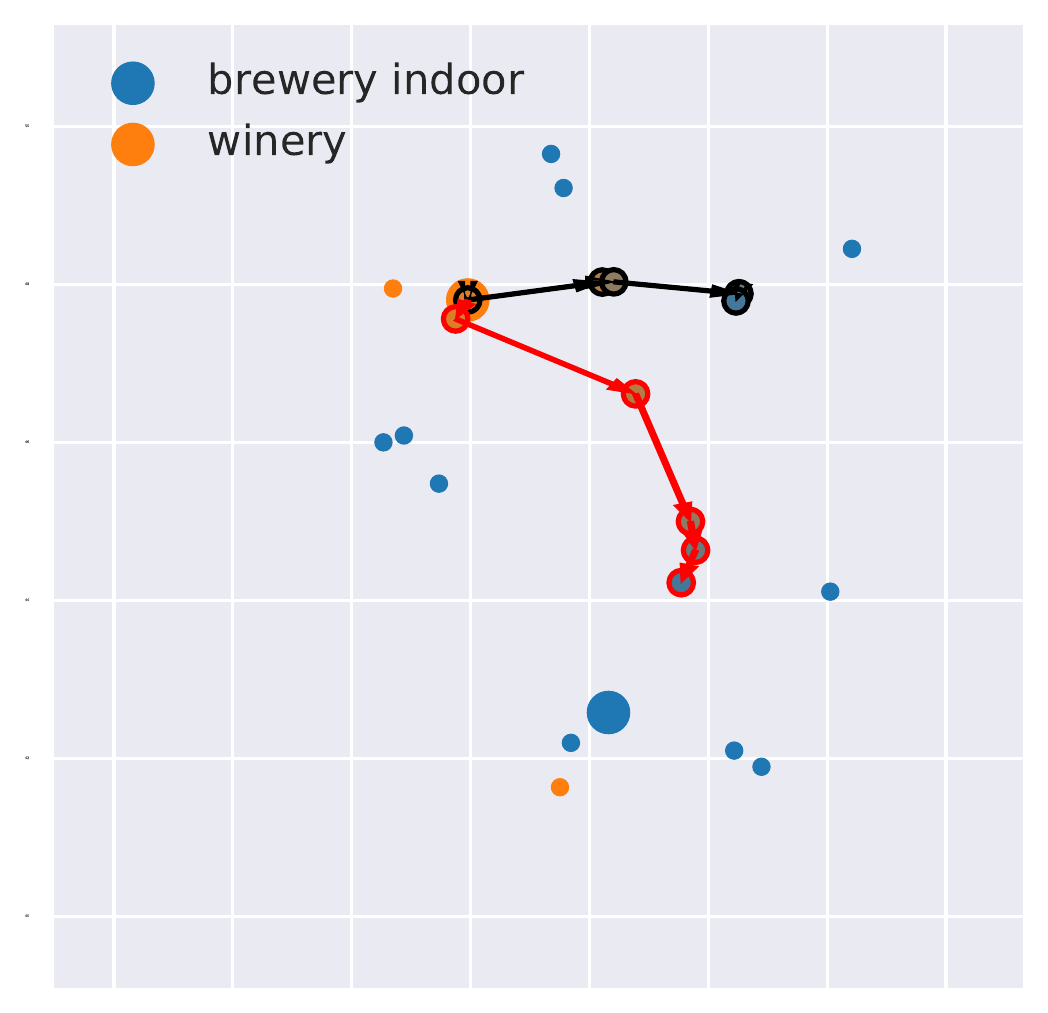}
\includegraphics[width=0.17\linewidth]{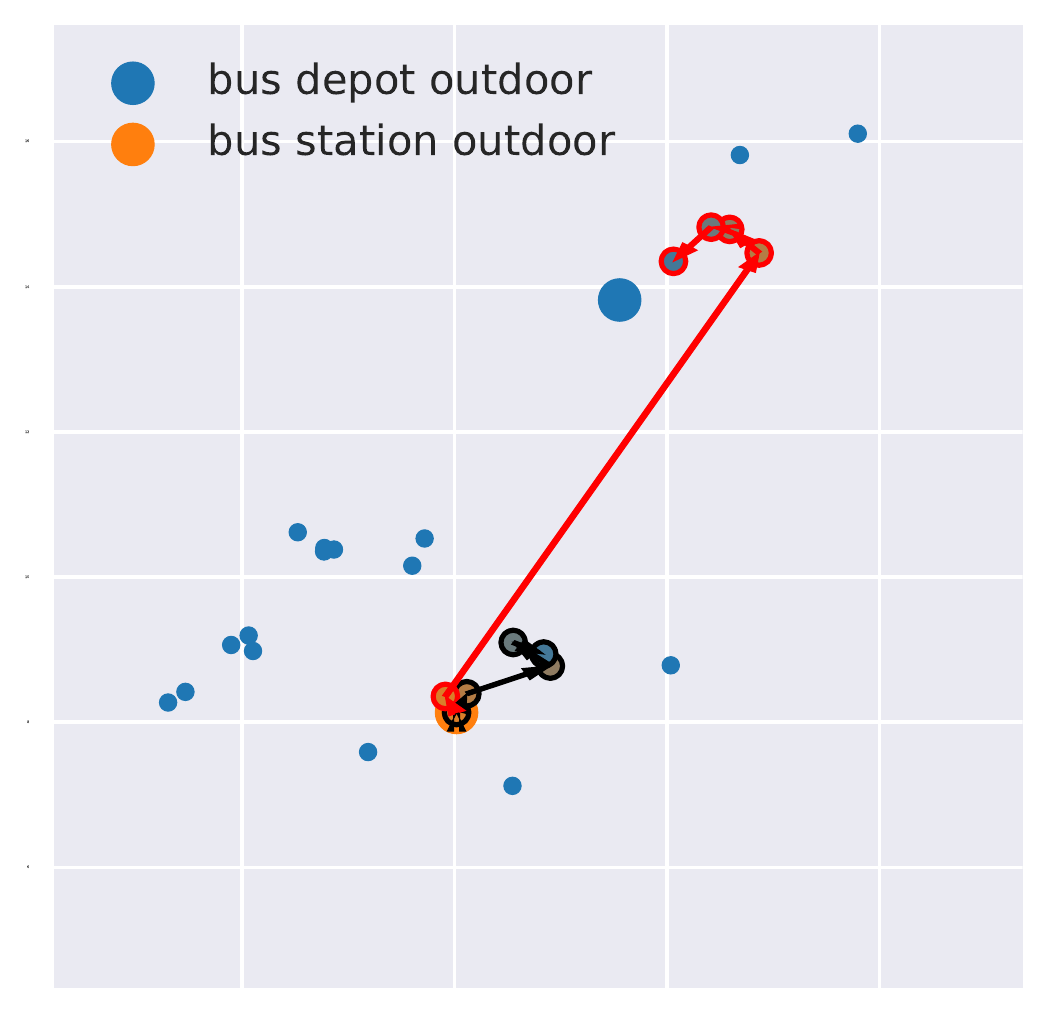}
\includegraphics[width=0.17\linewidth]{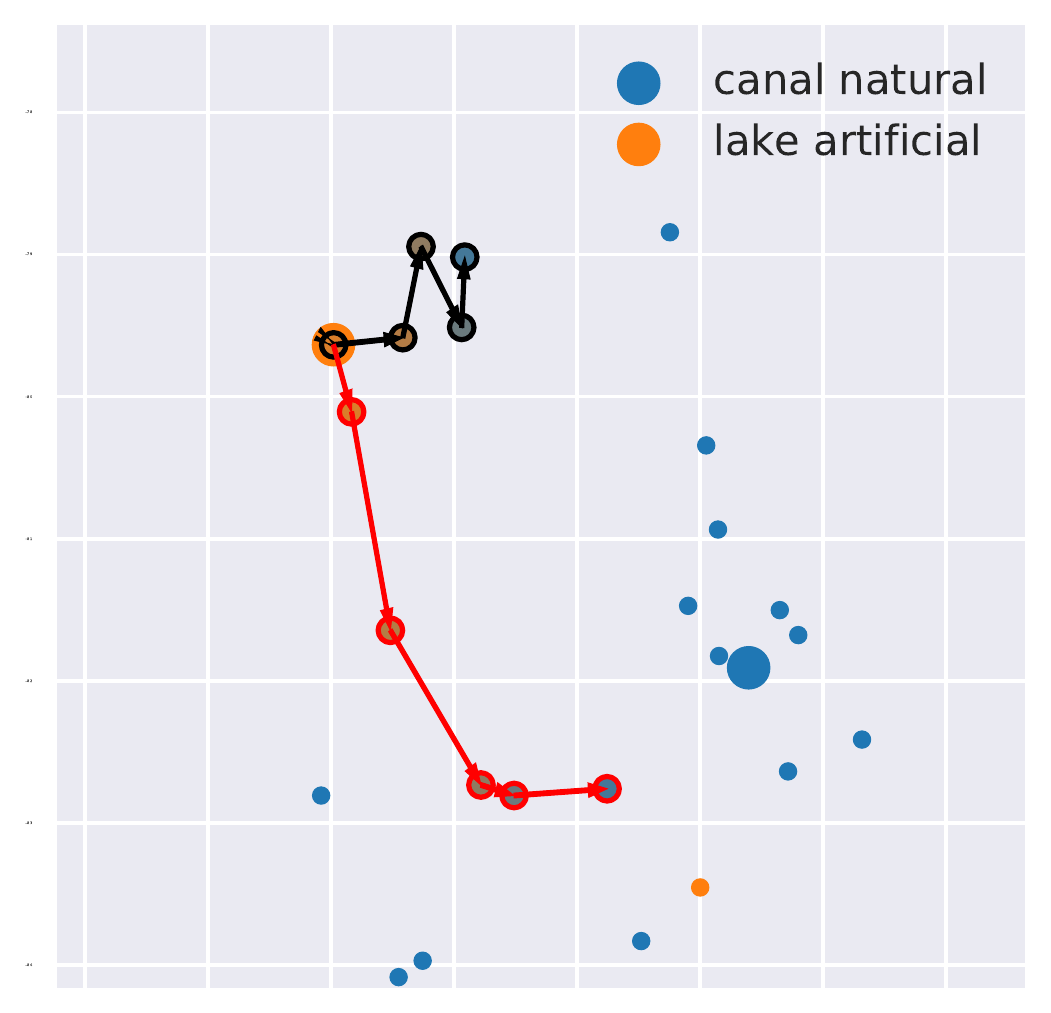}
\includegraphics[width=0.17\linewidth]{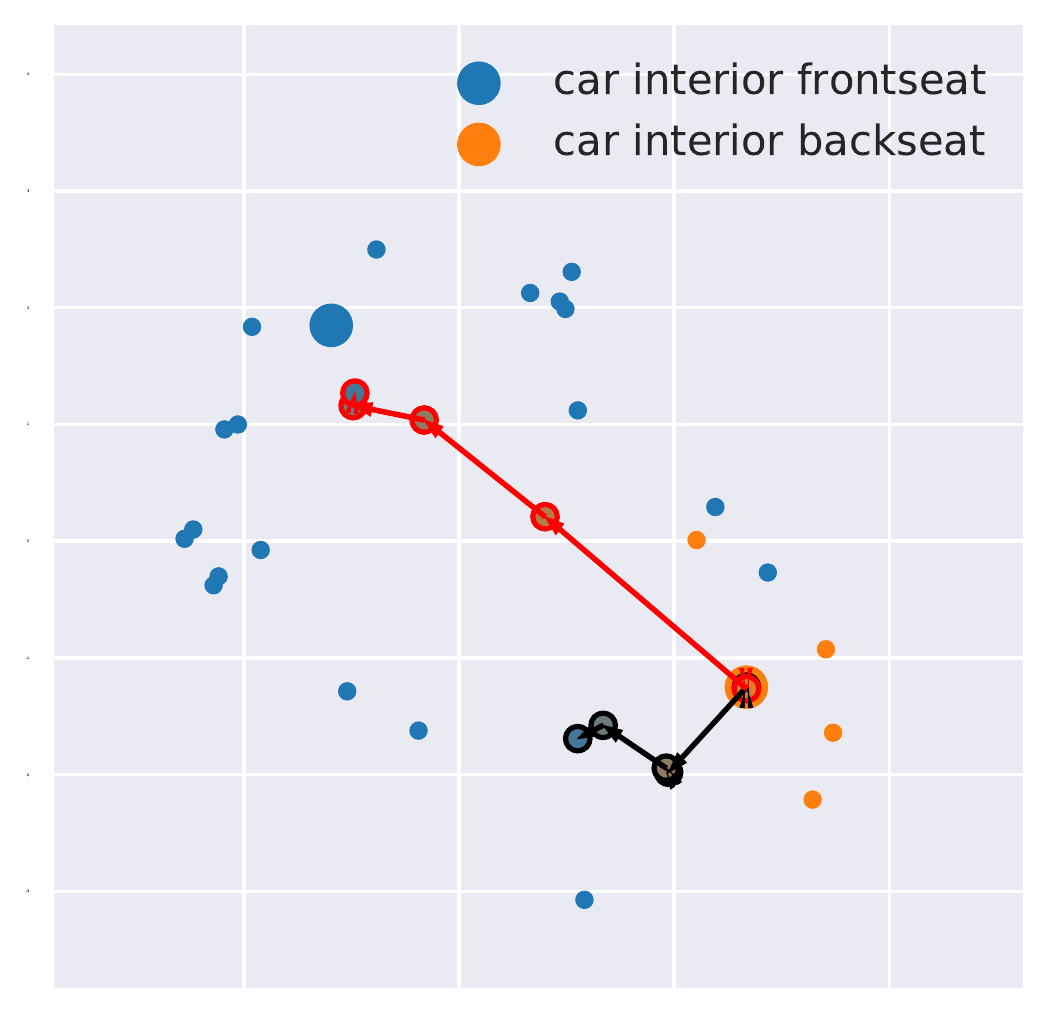}
\includegraphics[width=0.17\linewidth]{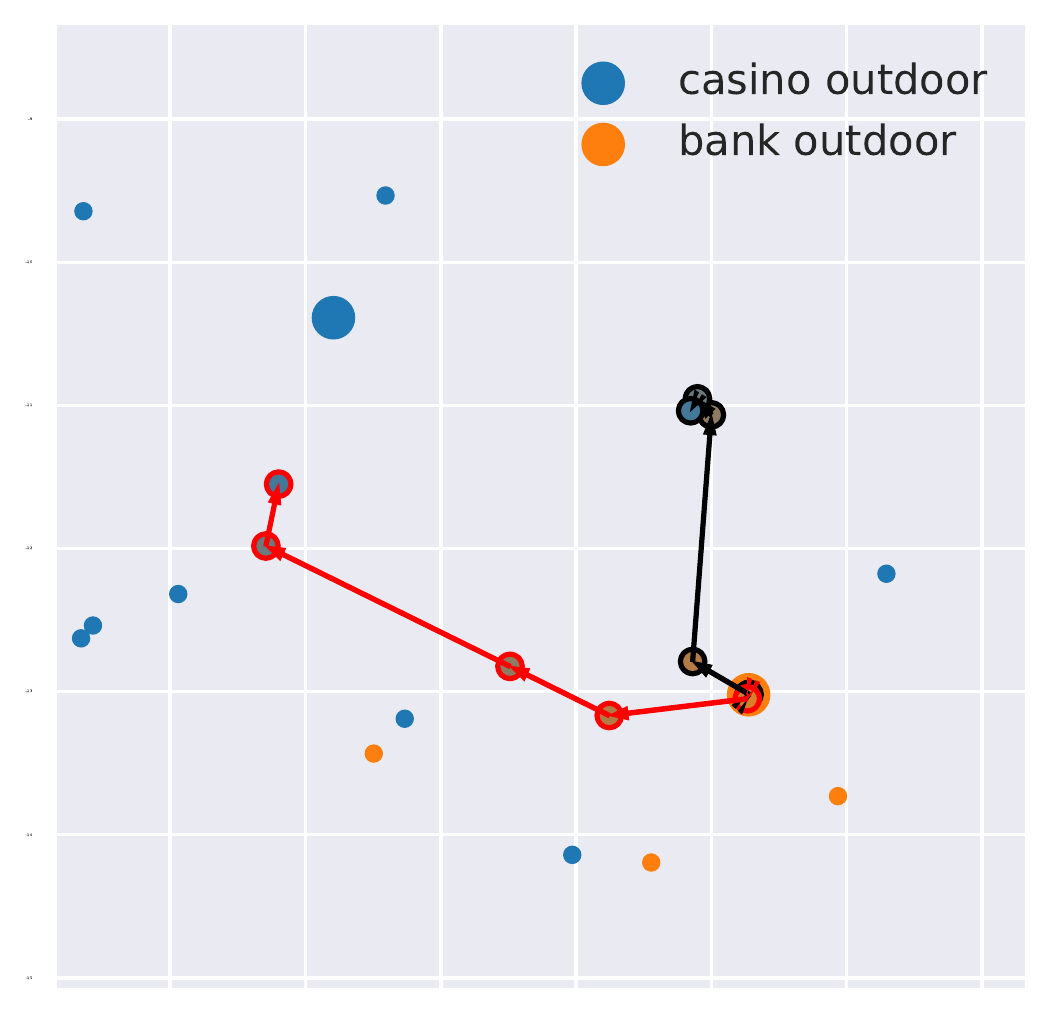}
\includegraphics[width=0.17\linewidth]{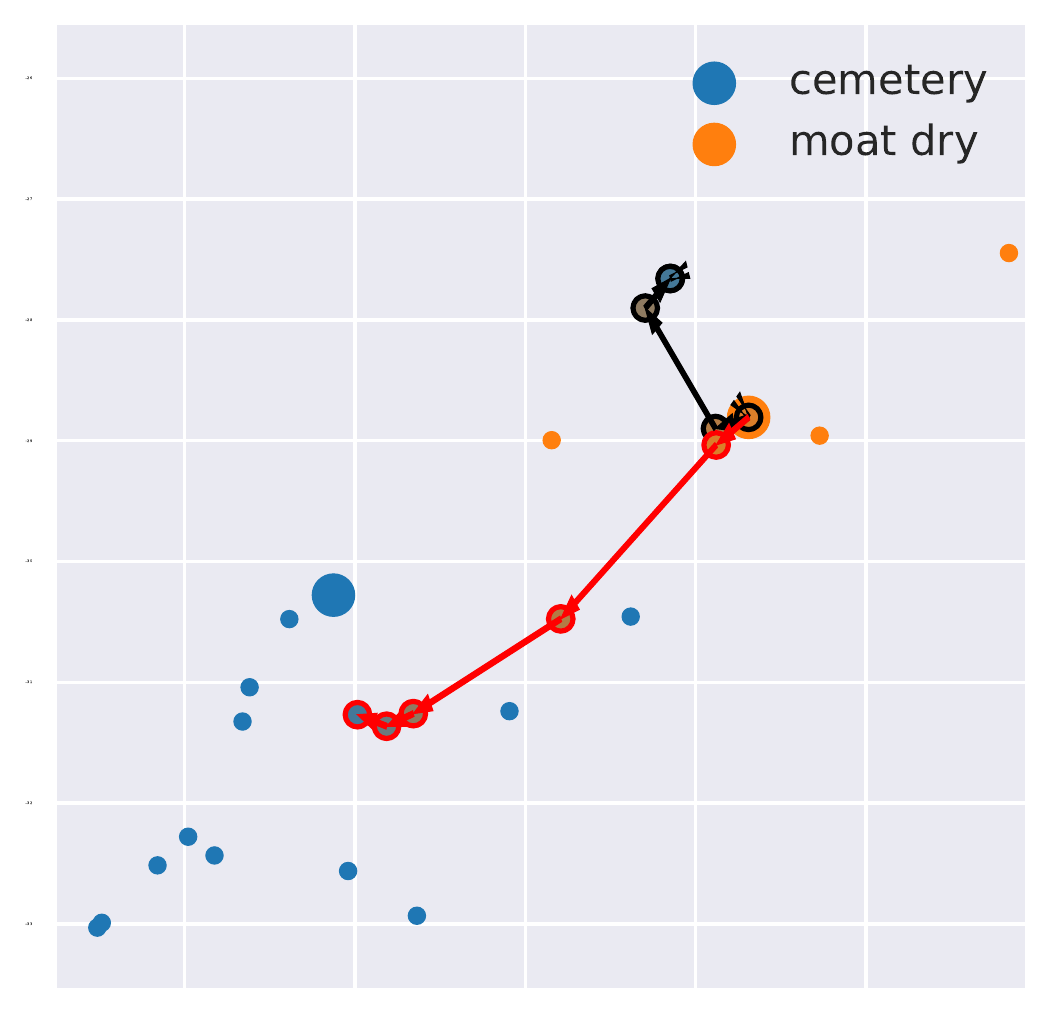}
\caption{t-SNE visualizations. For 20 SUN novel classes and corresponding similar base classes.
Smaller dots represent test images and larger dots represent class
attribute embeddings.
Red edges show the progression of novel class attributes as learners
interact using \emph{\sibvar}. Dots with black edges show the
progression with the \emph{random} function.
Both methods start at the base class attribute descriptor, and aim to
reach the novel class descriptor with as few interactions as possible. 
In most cases \emph{\sibvar} reaches closer to the novel class descriptor quicker in contrast to \emph{random}.
}
\label{fig:tsne_sun}
\end{figure*}

\begin{figure*}[h!]
\centering
\includegraphics[width=0.19\linewidth]{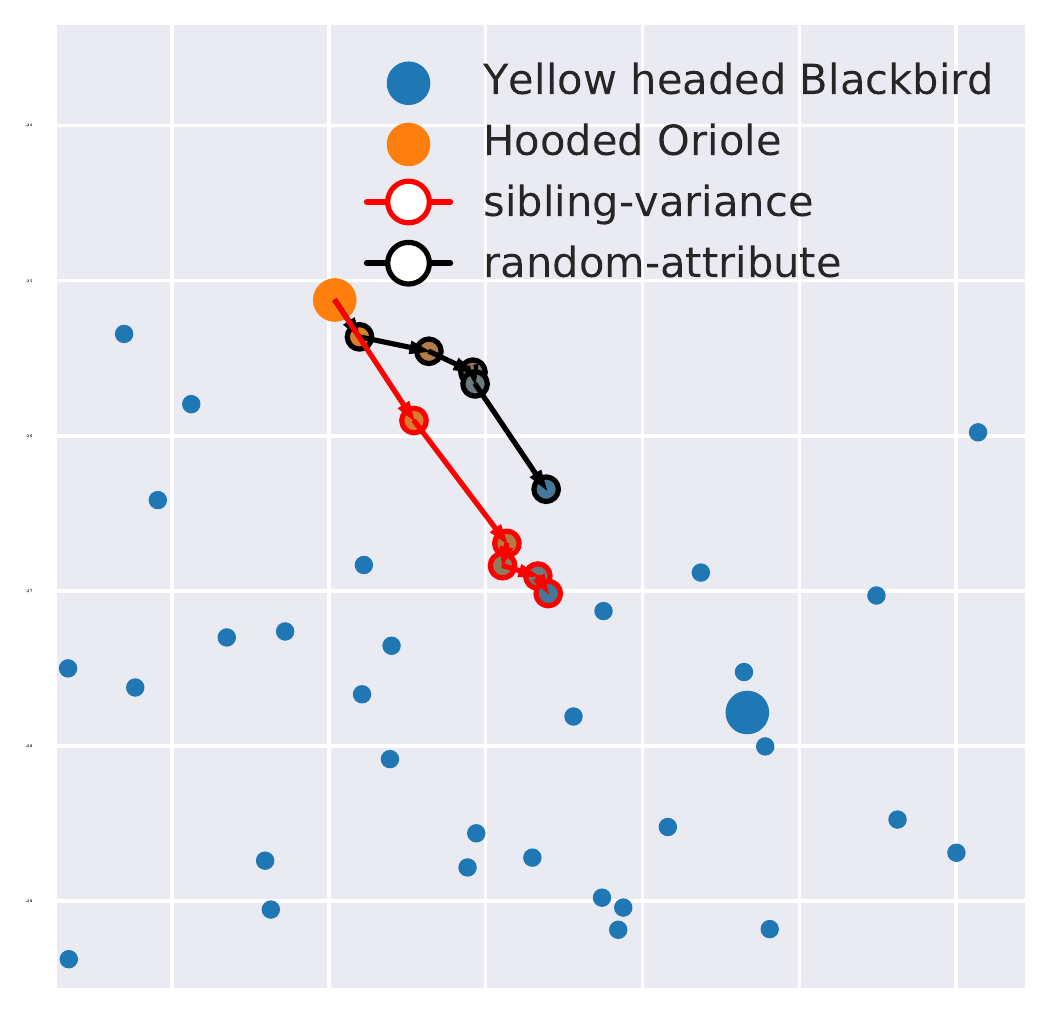}
\includegraphics[width=0.19\linewidth]{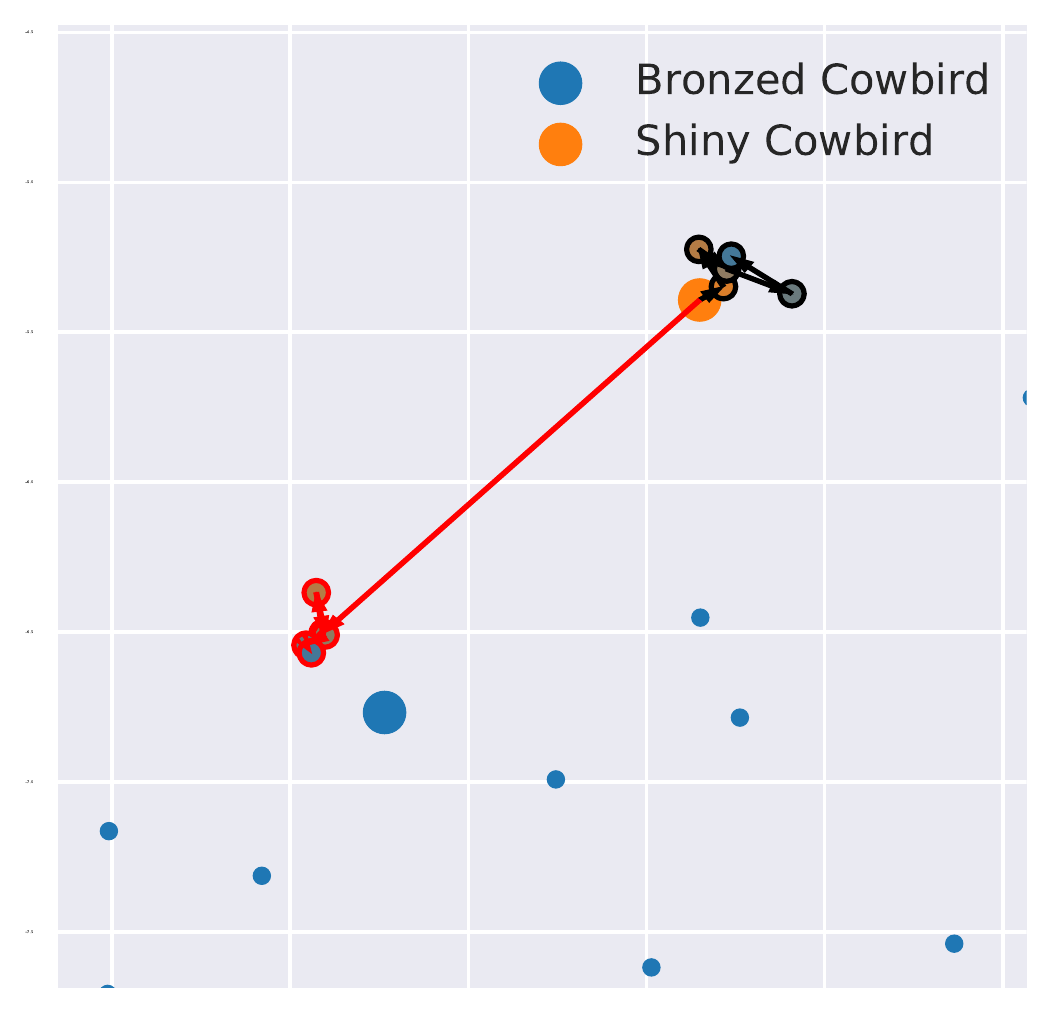}
\includegraphics[width=0.19\linewidth]{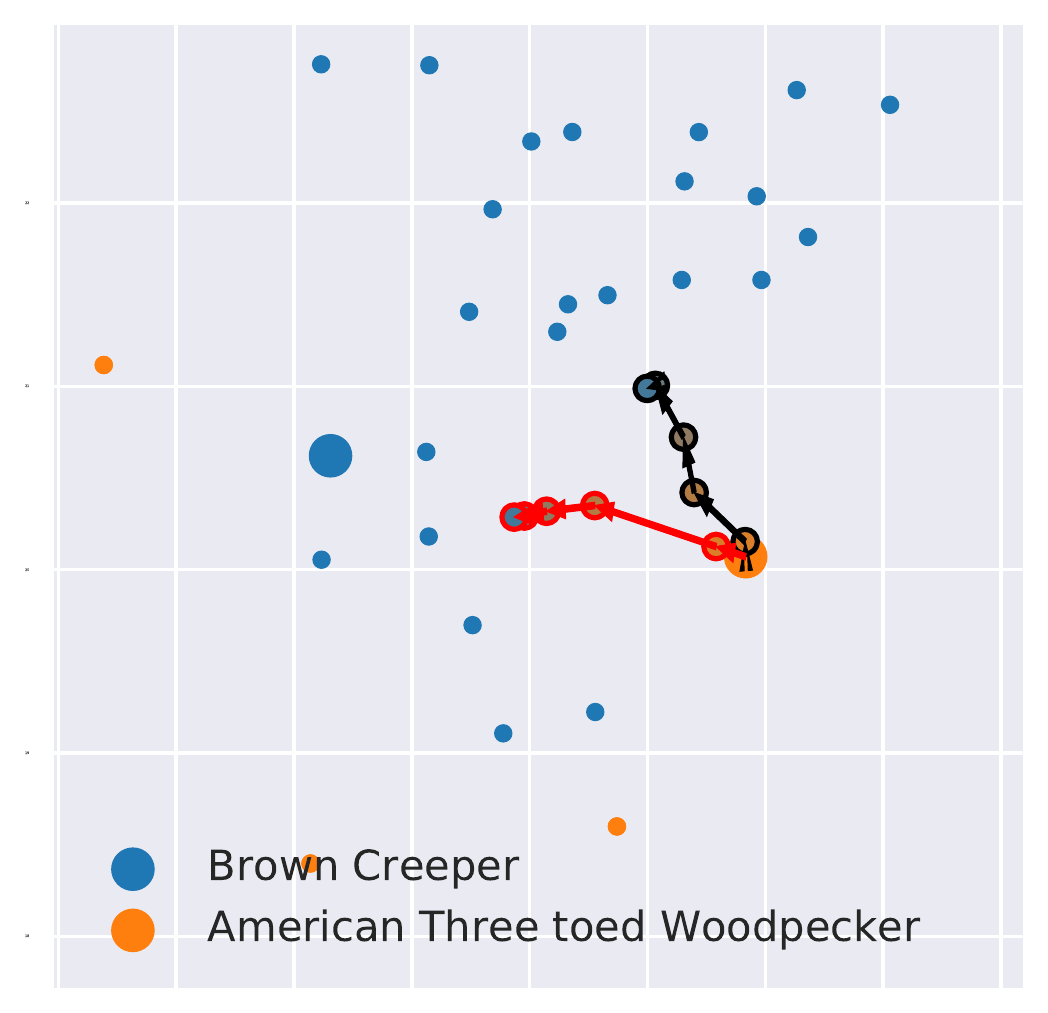}
\includegraphics[width=0.19\linewidth]{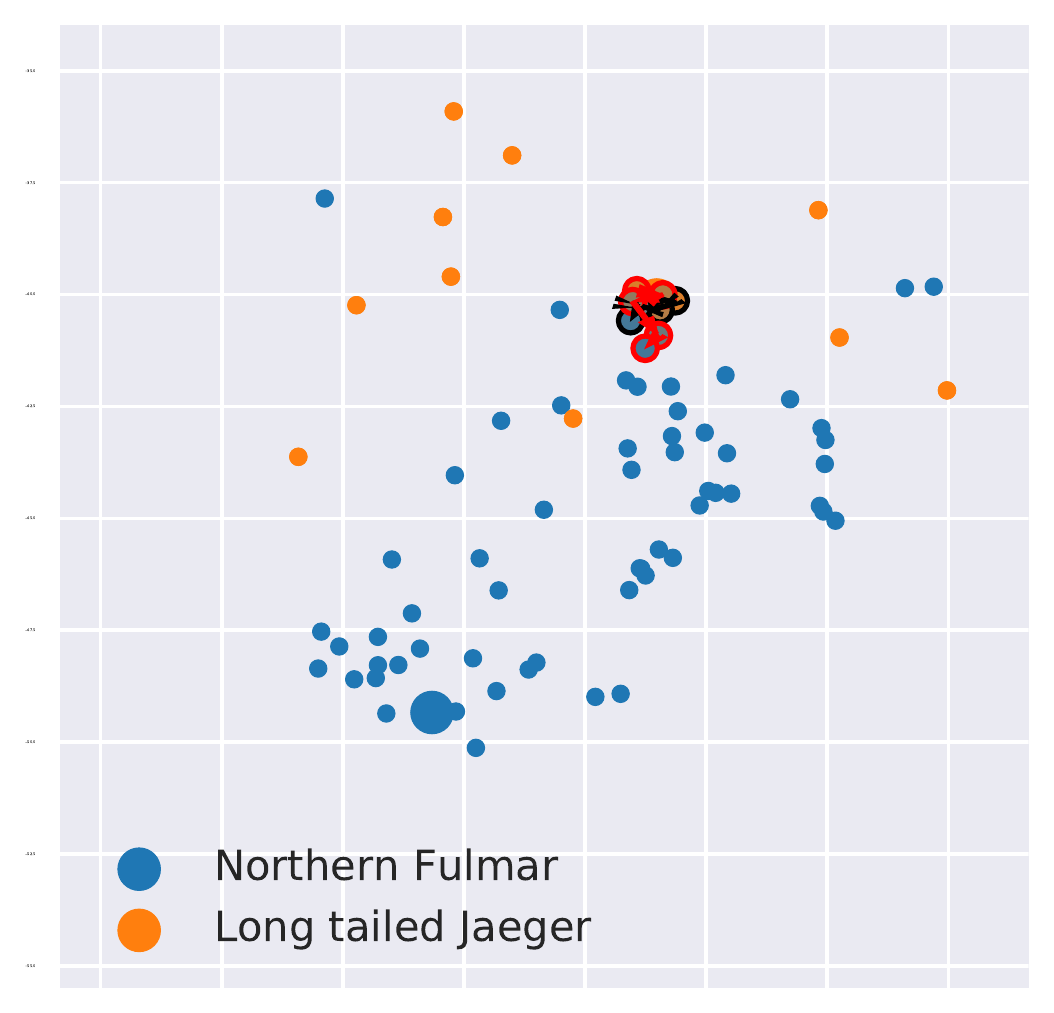}
\includegraphics[width=0.19\linewidth]{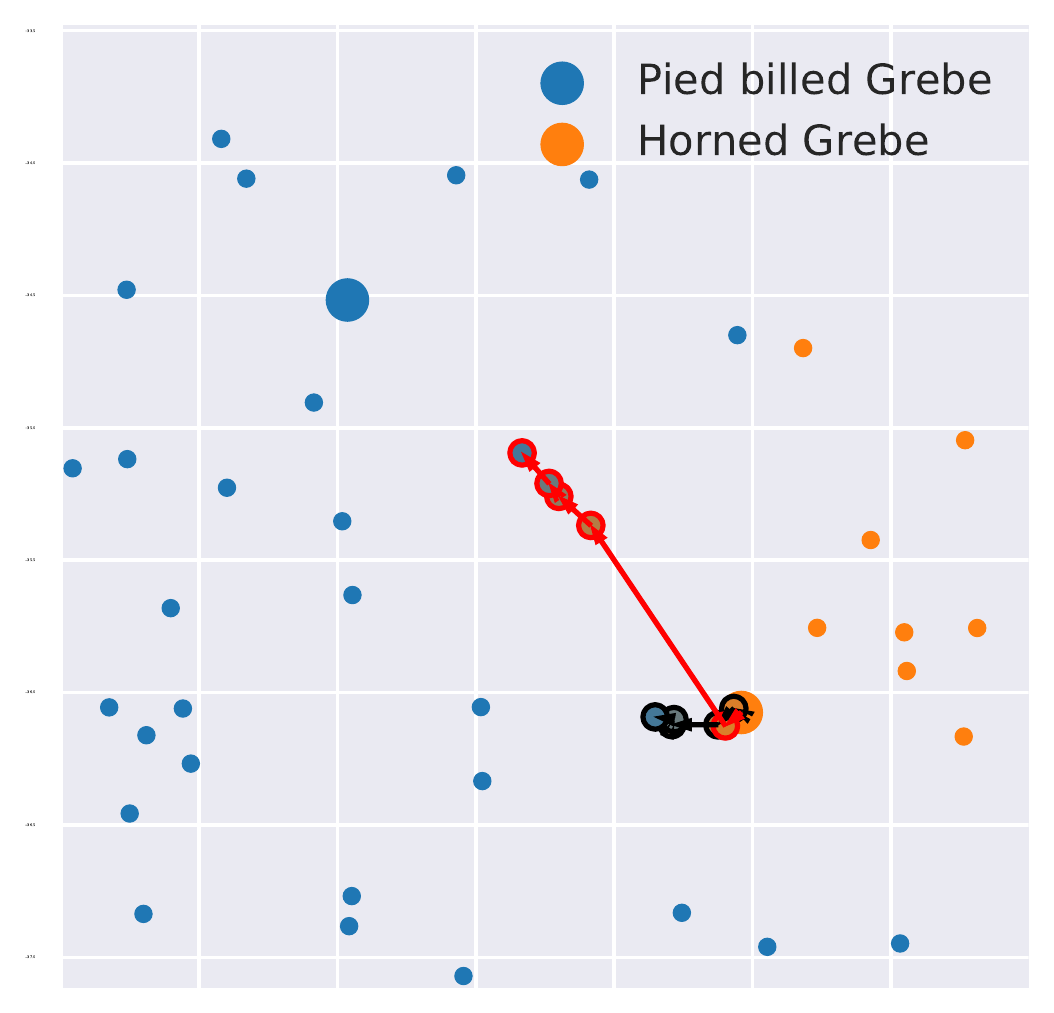}
\includegraphics[width=0.19\linewidth]{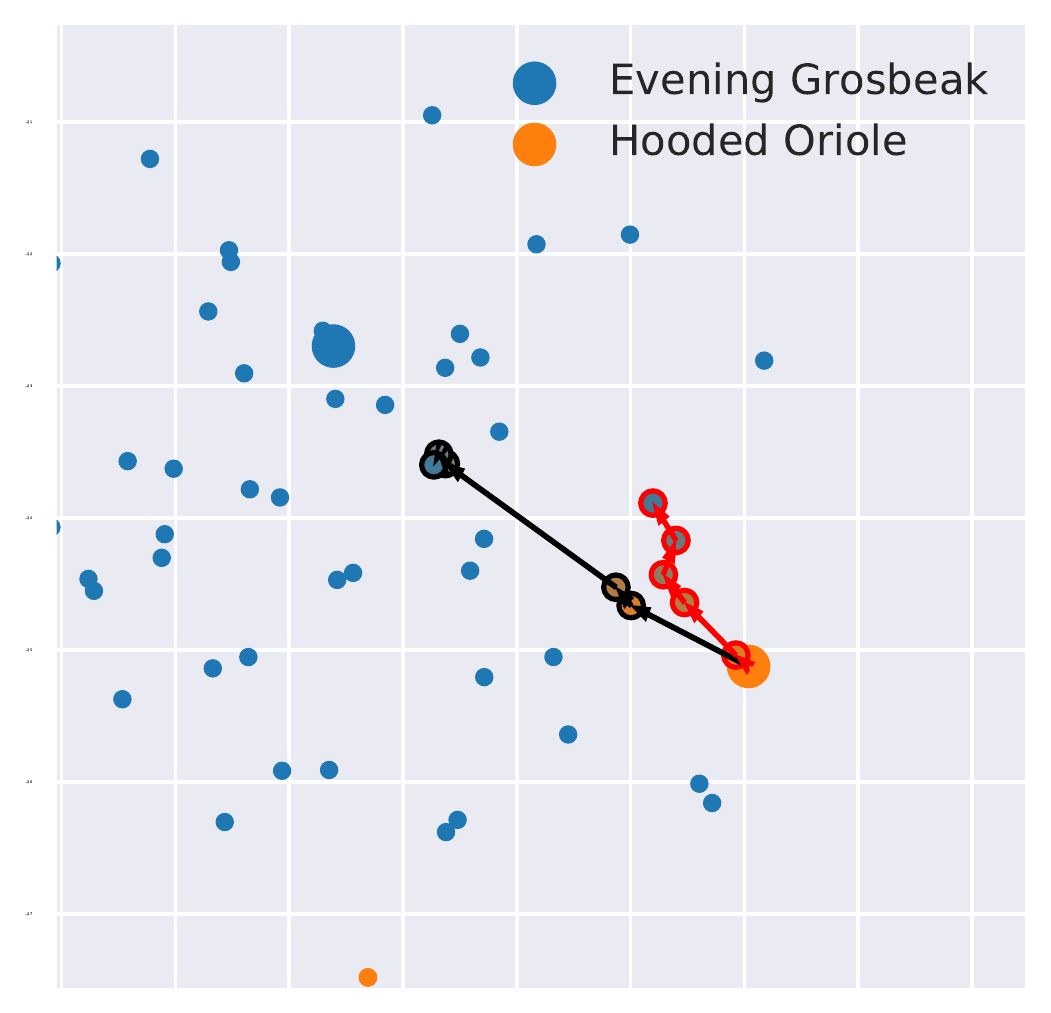}
\includegraphics[width=0.19\linewidth]{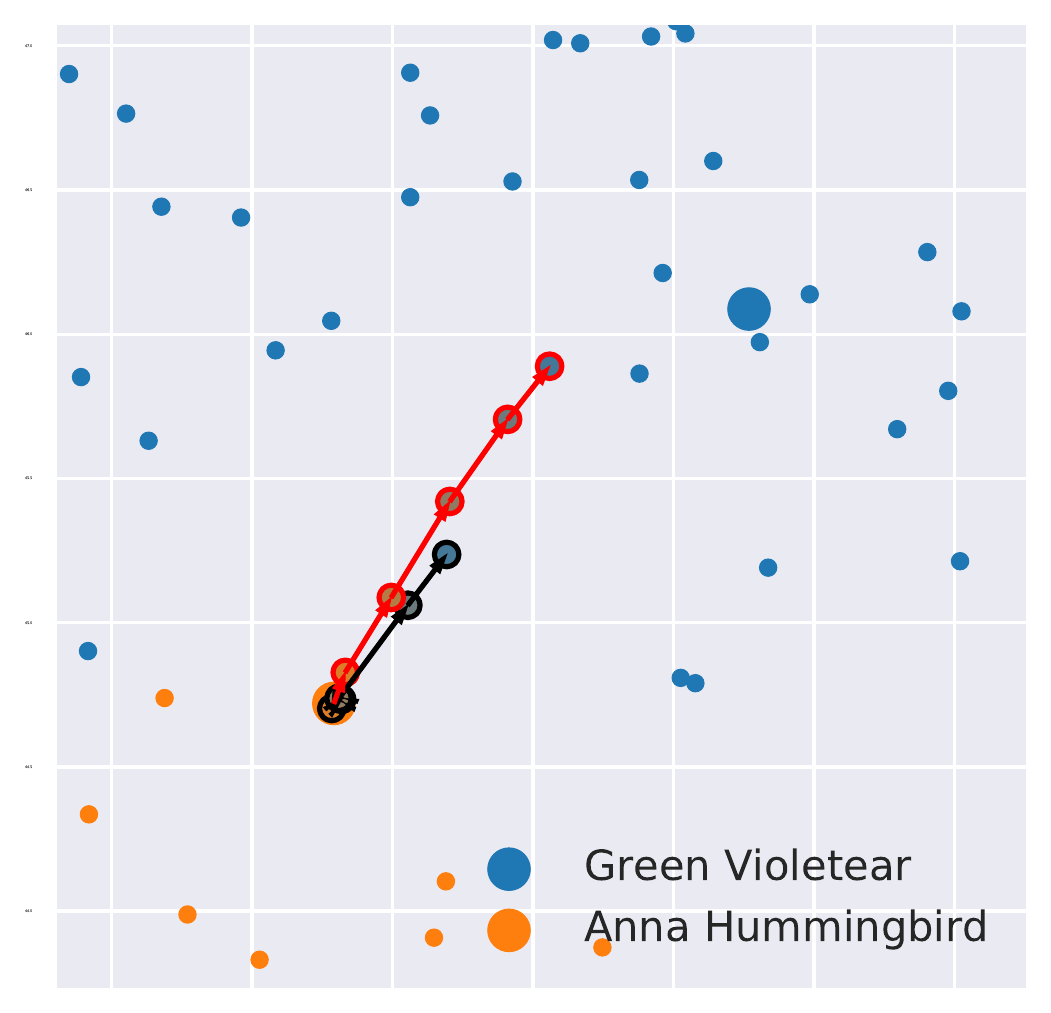}
\includegraphics[width=0.19\linewidth]{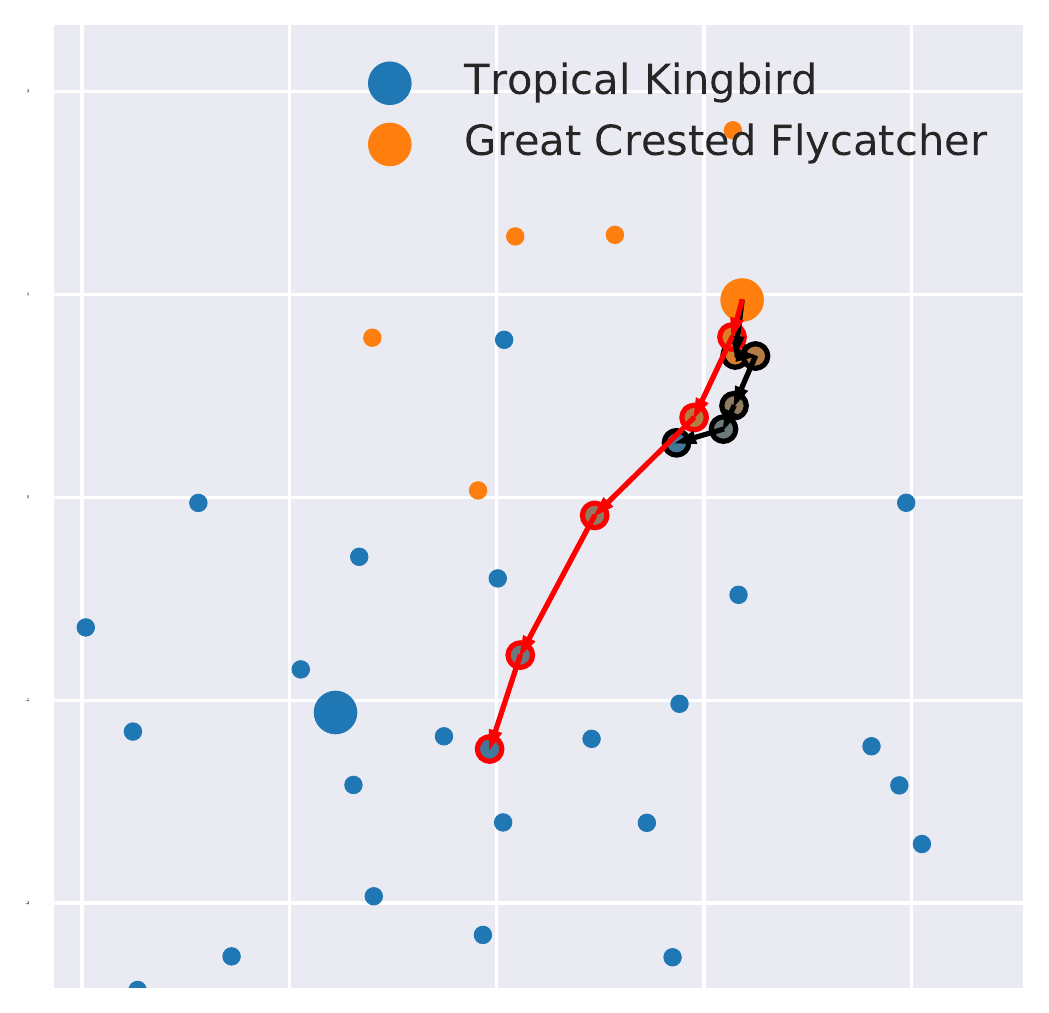}
\includegraphics[width=0.19\linewidth]{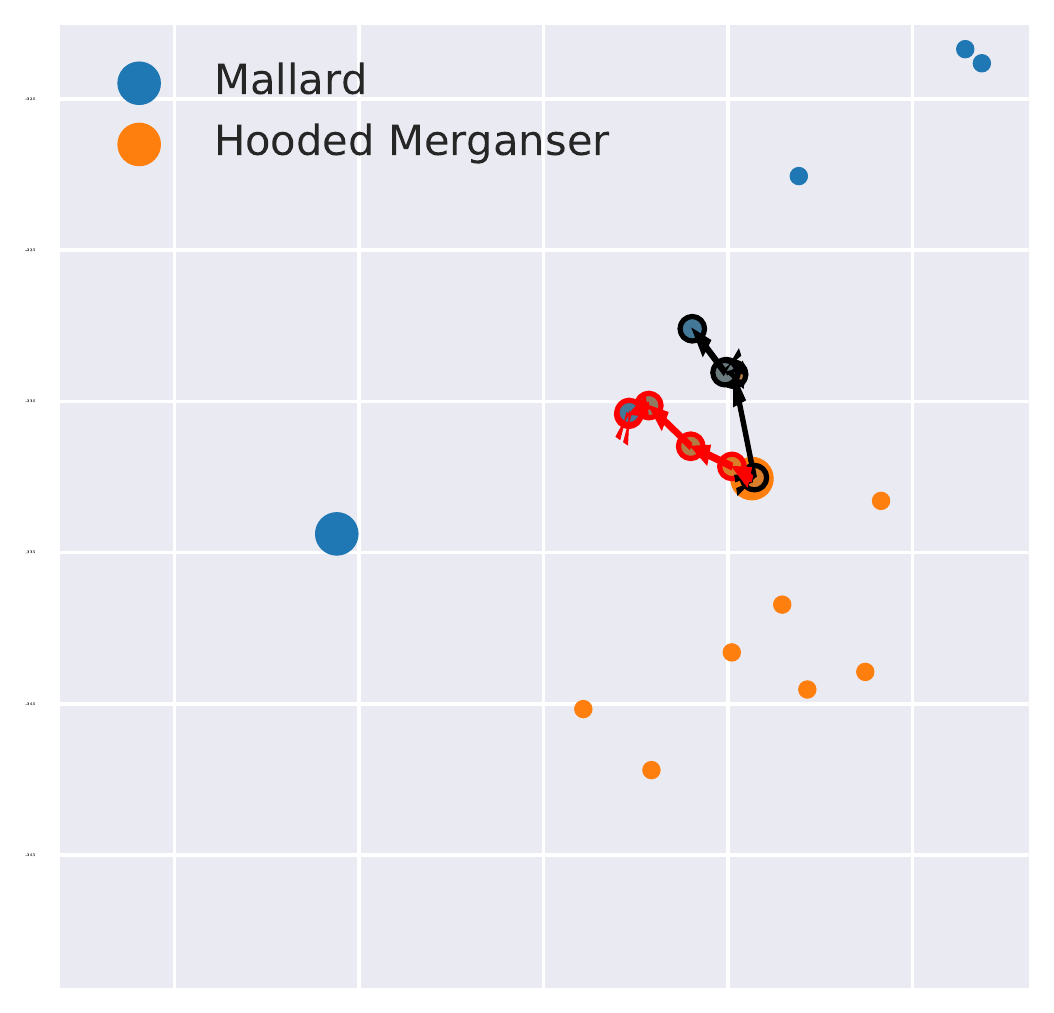}
\includegraphics[width=0.19\linewidth]{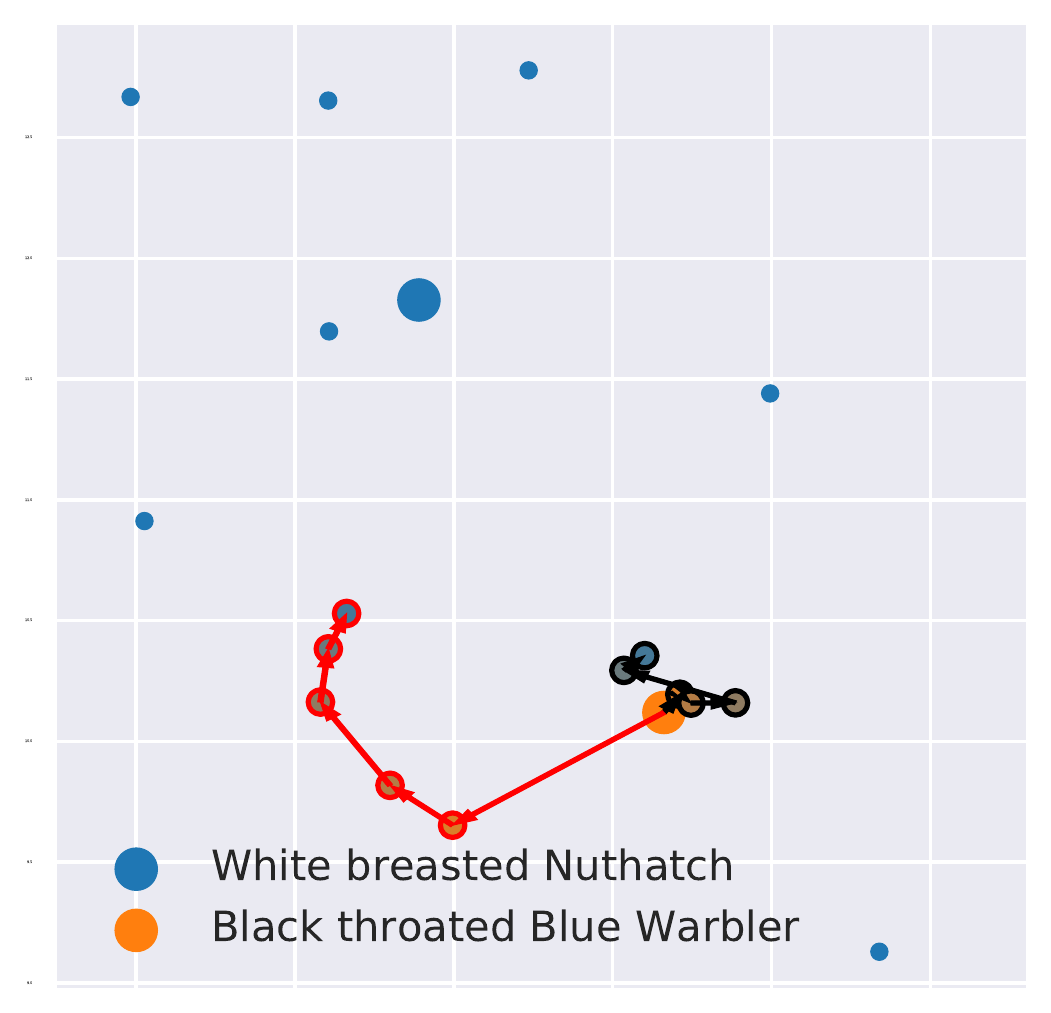}
\includegraphics[width=0.19\linewidth]{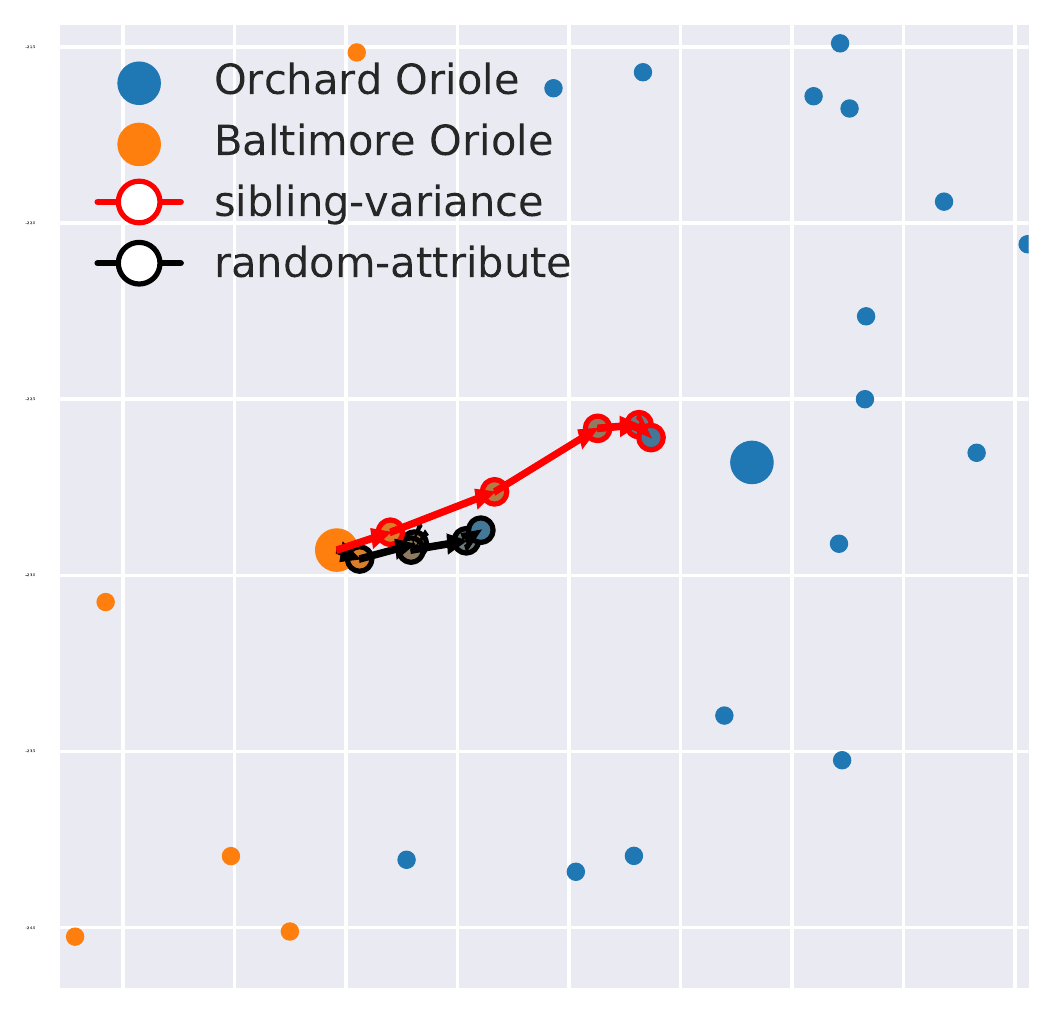}
\includegraphics[width=0.19\linewidth]{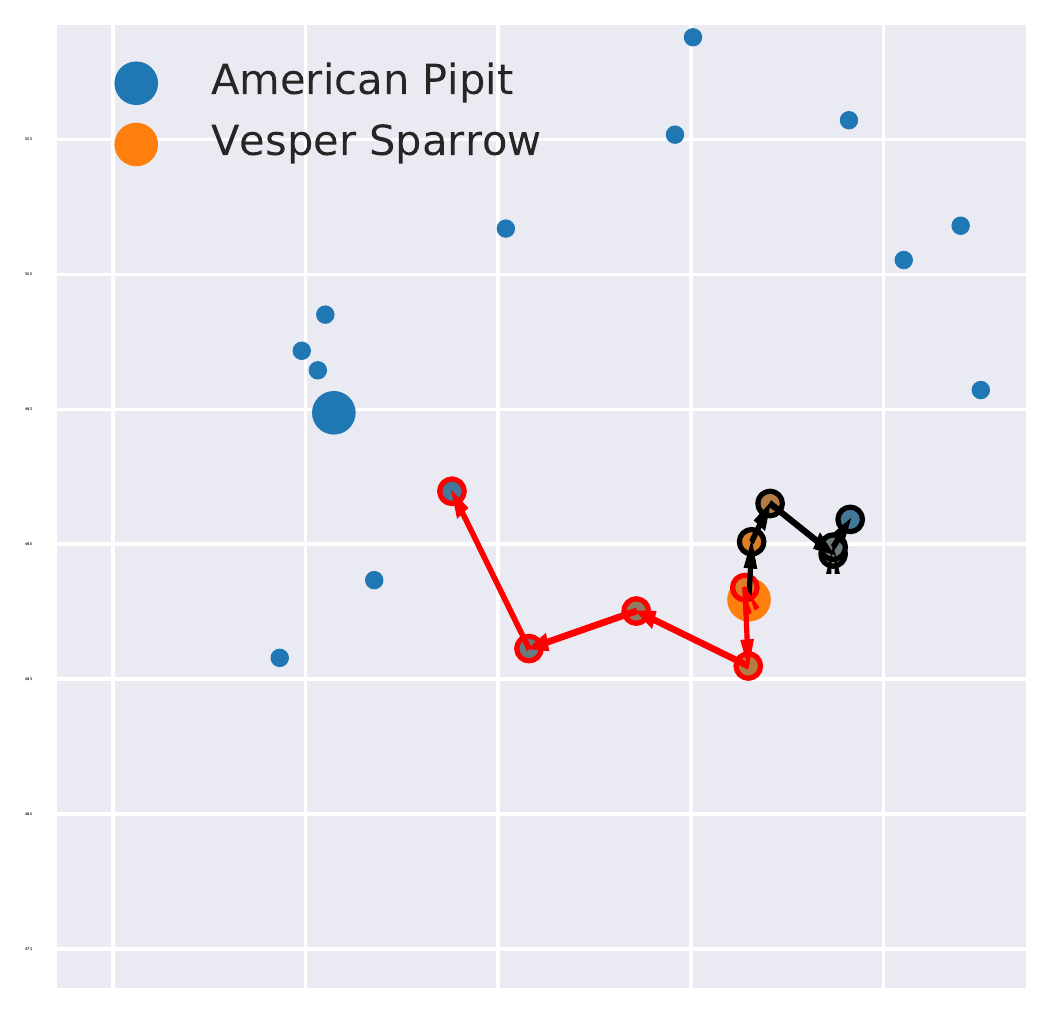}
\includegraphics[width=0.19\linewidth]{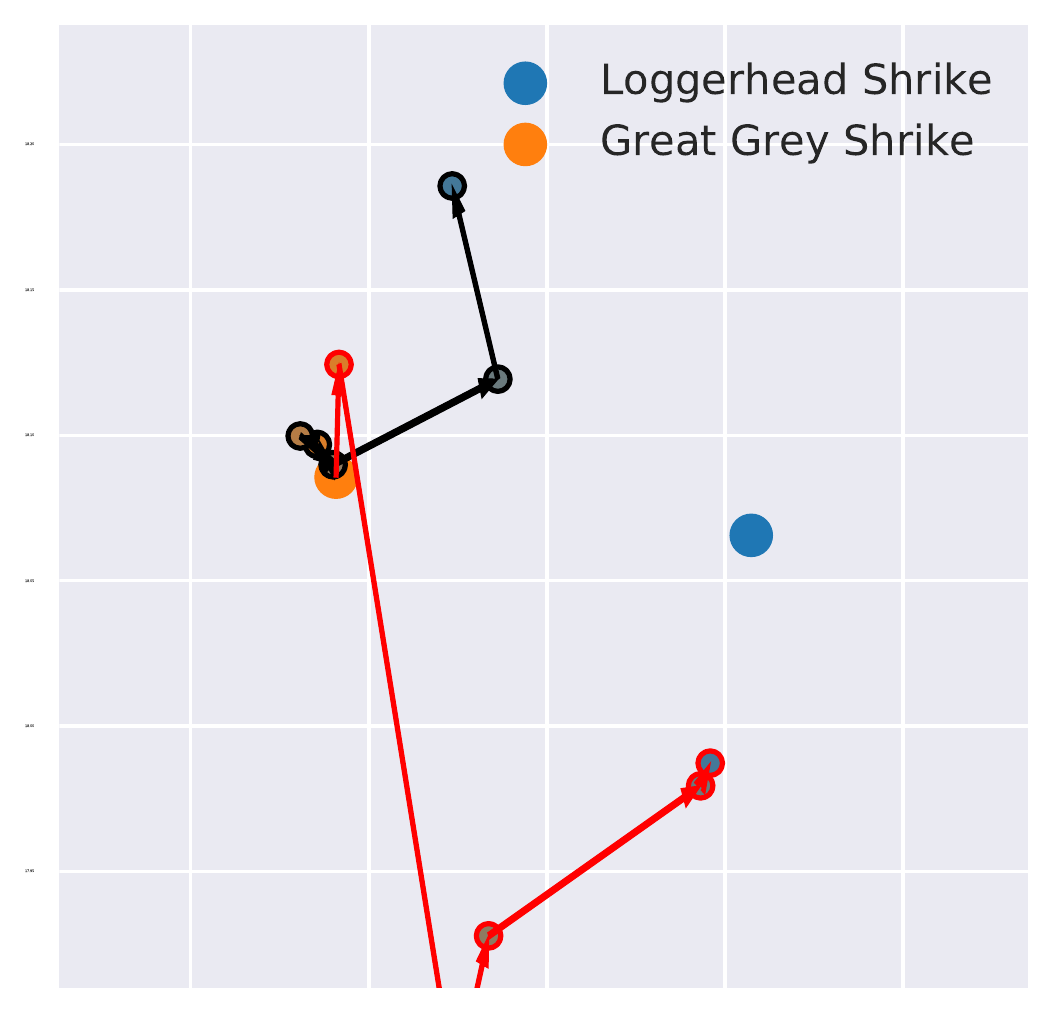}
\includegraphics[width=0.19\linewidth]{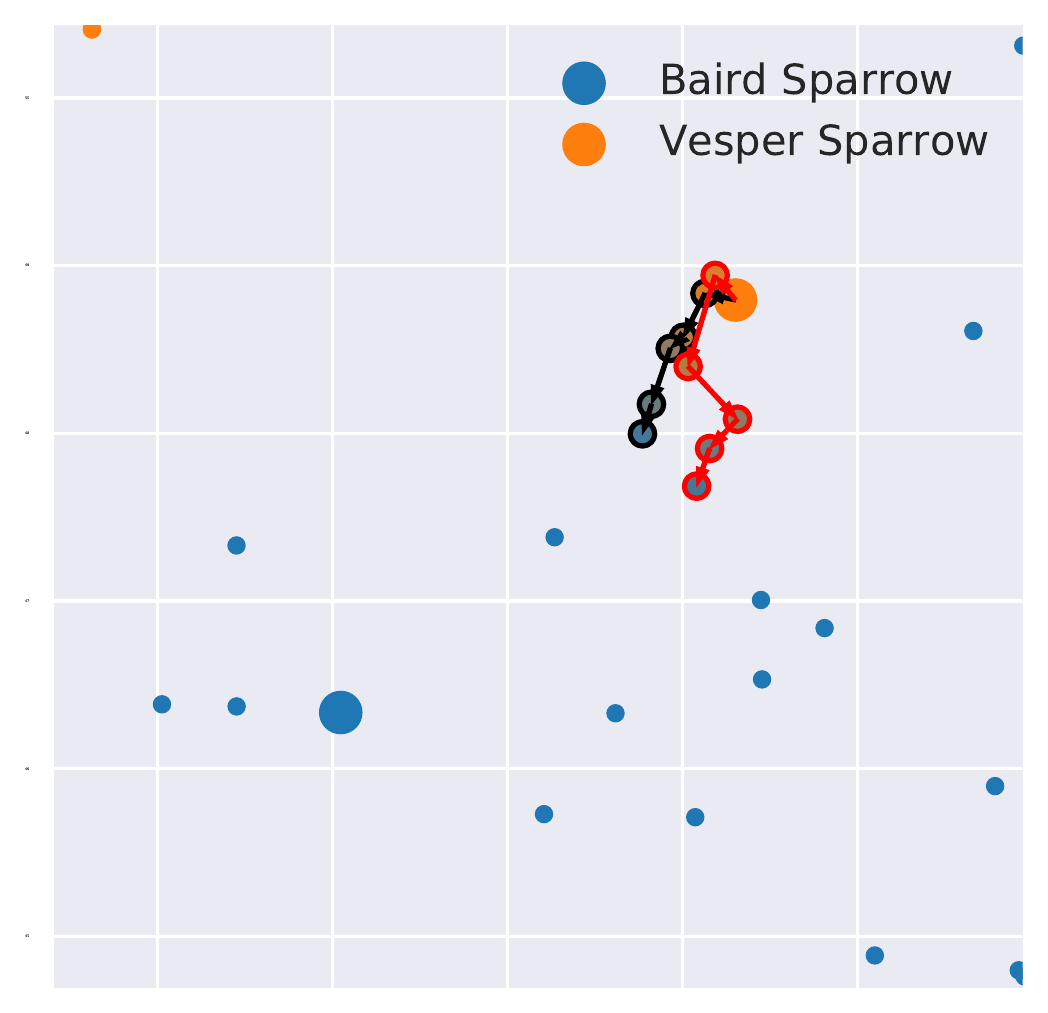}
\includegraphics[width=0.19\linewidth]{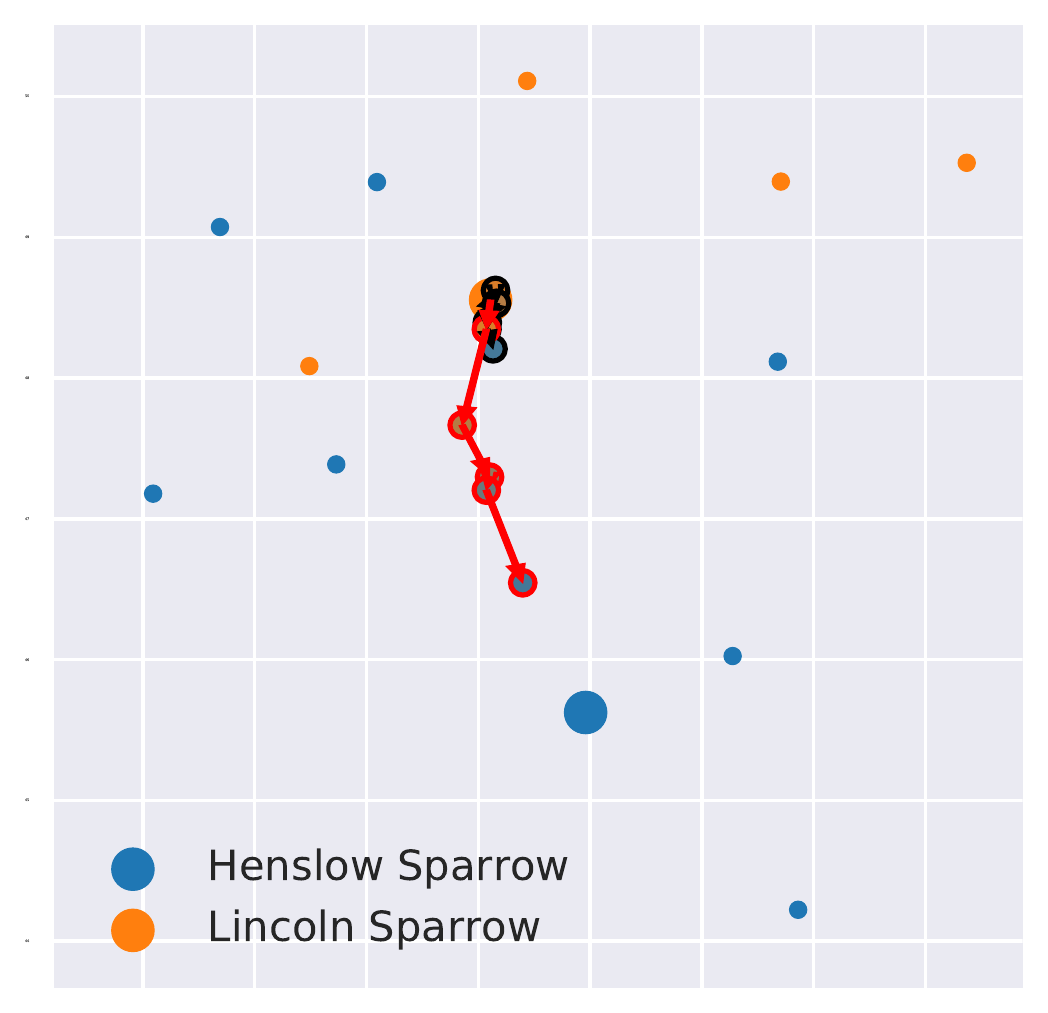}
\includegraphics[width=0.19\linewidth]{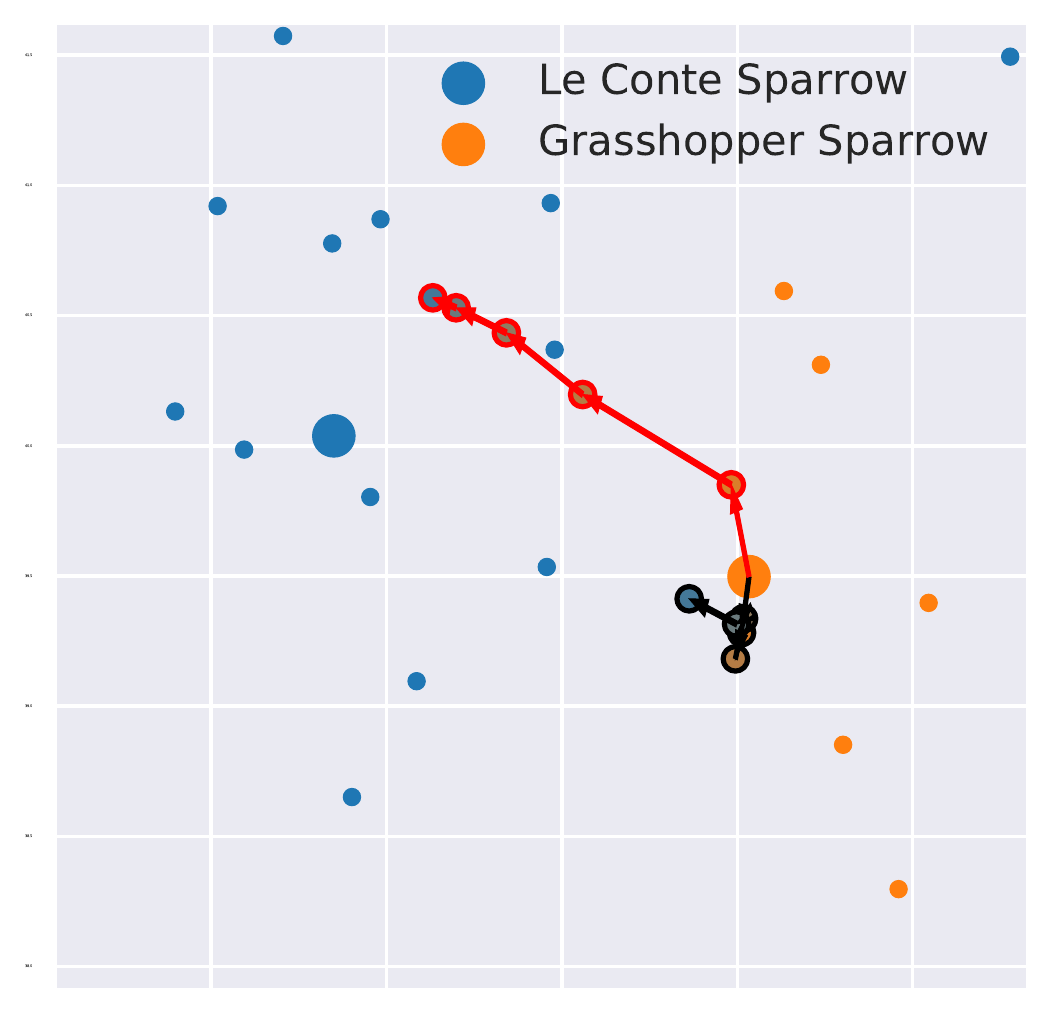}
\includegraphics[width=0.19\linewidth]{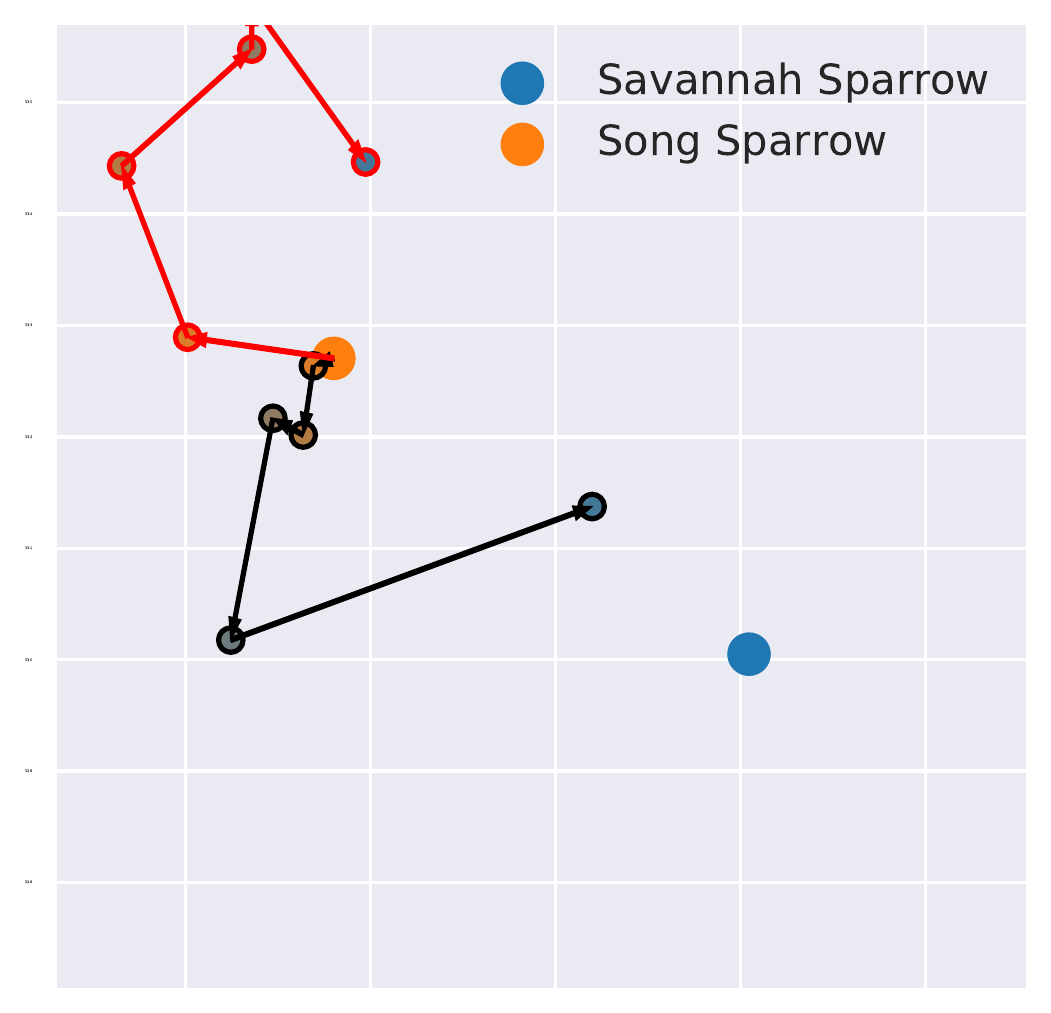}
\includegraphics[width=0.19\linewidth]{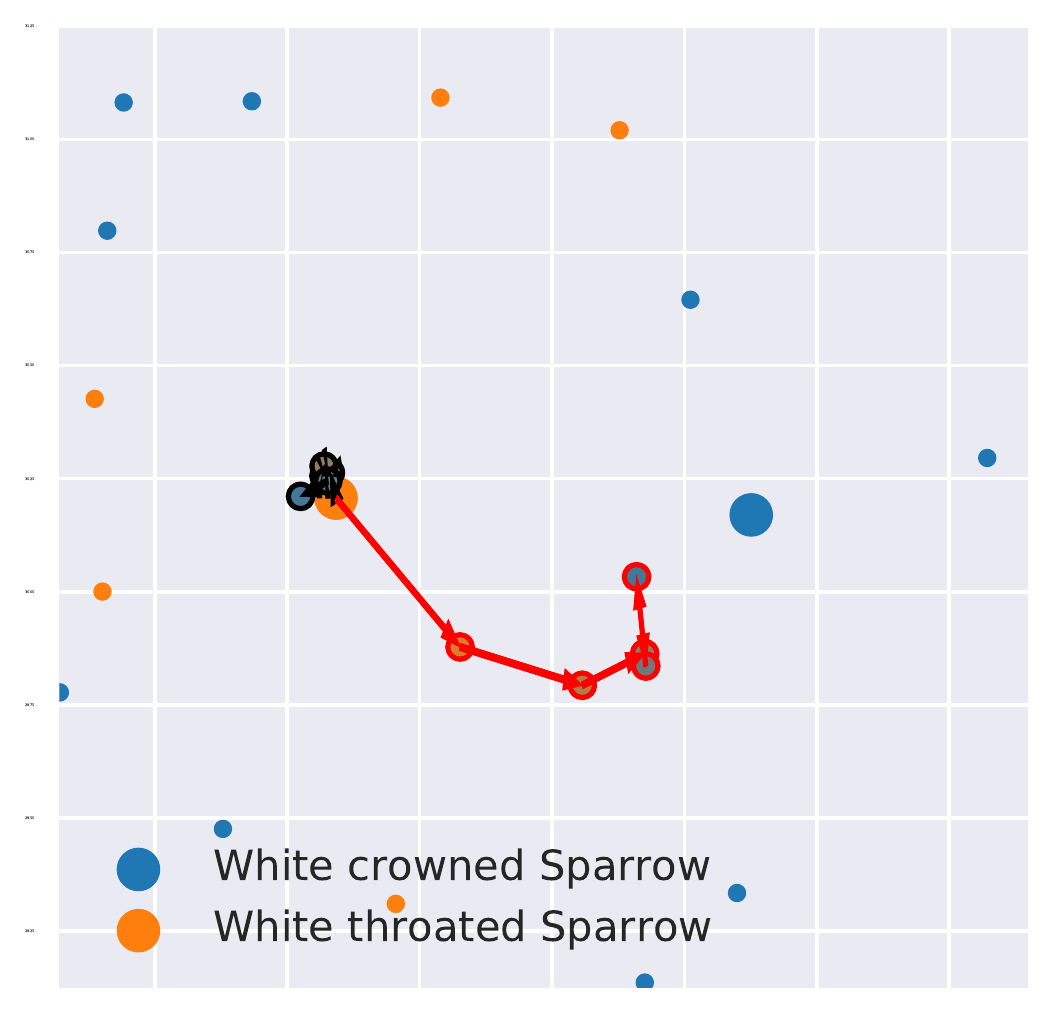}
\includegraphics[width=0.19\linewidth]{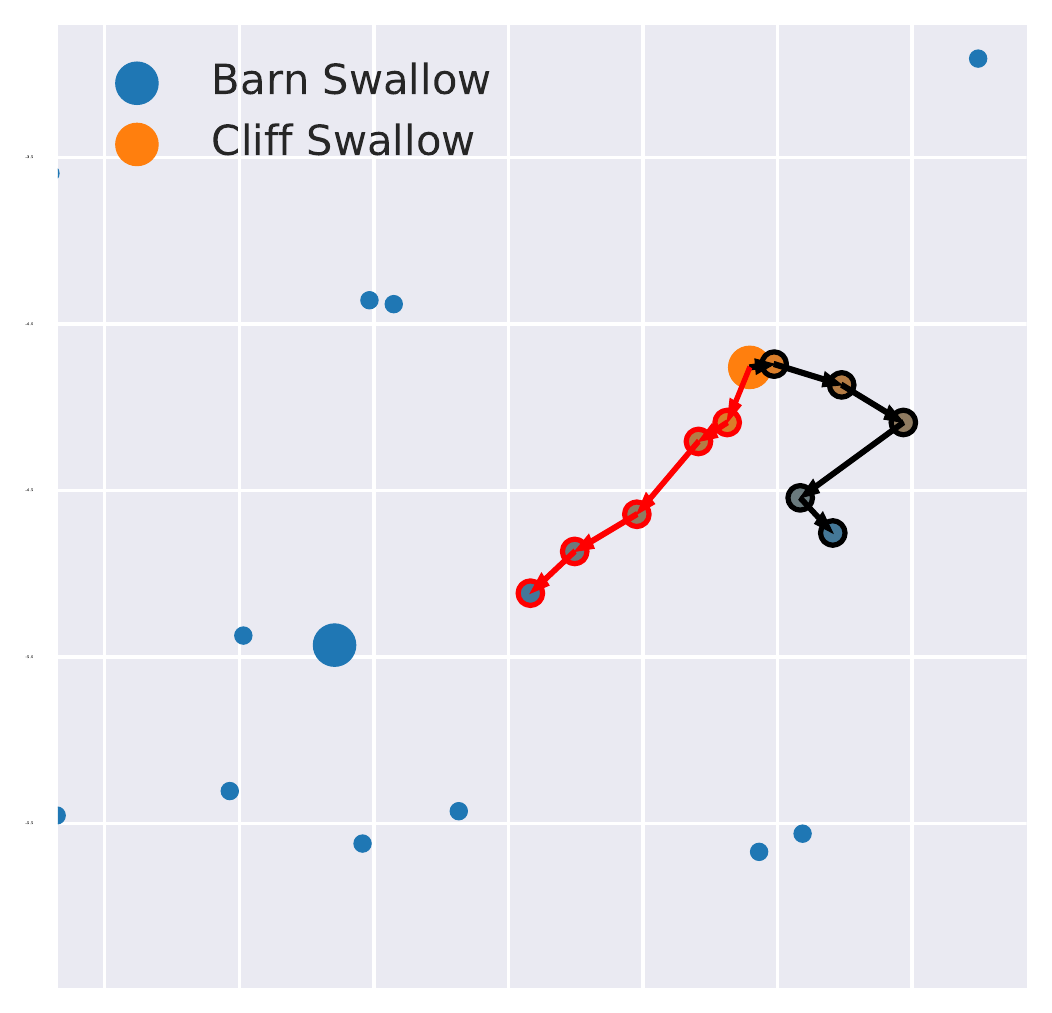}
\includegraphics[width=0.19\linewidth]{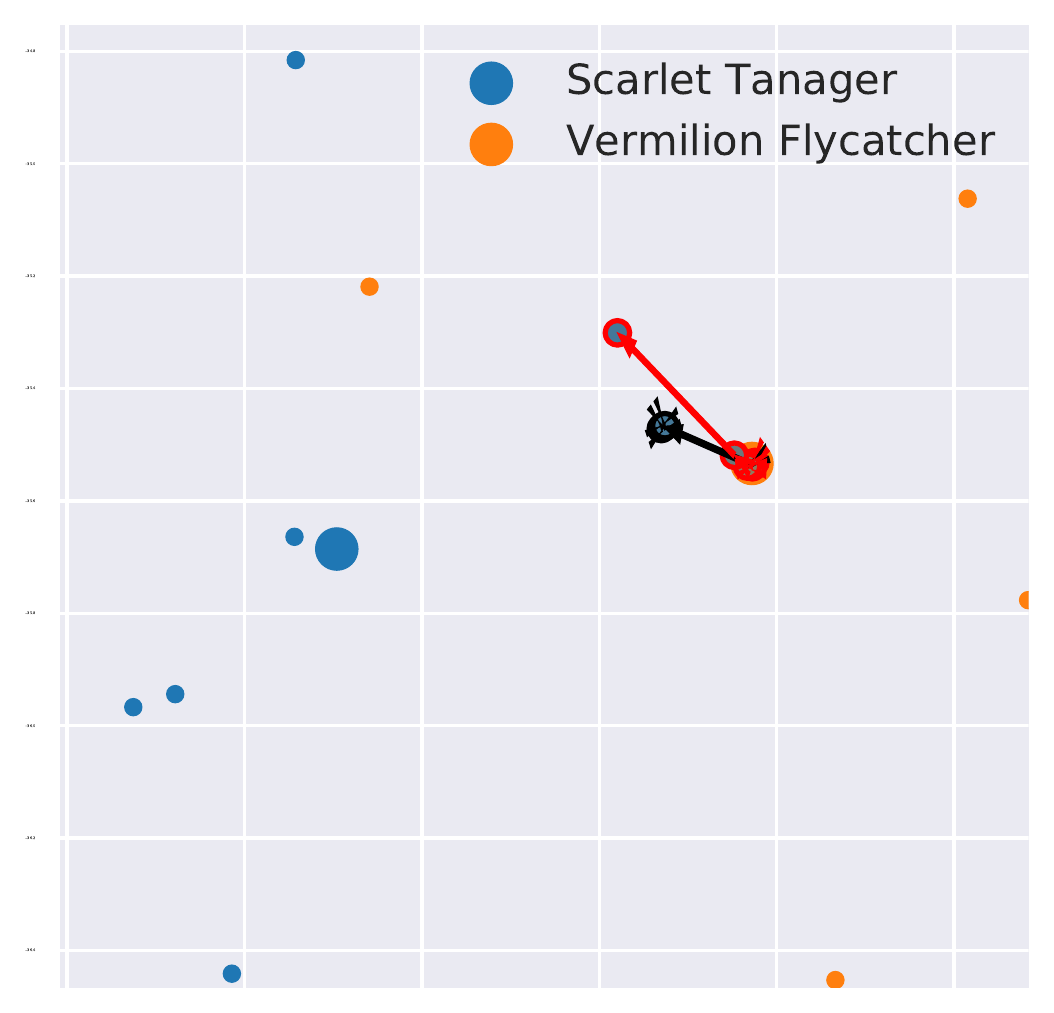}
\caption{t-SNE visualizations. For 20 CUB novel classes and corresponding similar base classes.
Smaller dots represent test images and larger dots represent class
attribute embeddings.
Red edges show the progression of novel class attributes as learners
interact using \emph{\sibvar}. Dots with black edges show the
progression with the \emph{random} function.
Both methods start at the base class attribute descriptor, and aim to reach the novel class descriptor with as fewer interactions as possible. 
In most cases \emph{\sibvar} reaches closer to the novel class descriptor quicker in contrast to \emph{random}.
}
\label{fig:tsne_cub}
\end{figure*}

Figure \ref{fig:tsne_sun} and \ref{fig:tsne_cub} show the progression of \emph{\sibvar} and random attributes for all 20 novel classes of CUB (out of 50) and  SUN (out of 70).   
Faster progression can be seen for \emph{\sibvar} over these classes as well.

\end{document}